\documentstyle[12pt]{article}  
\title{A Framework for Searching AND/OR Graphs with Cycles \thanks{Supported by
AICTE Research
Project 
on AI and Expert Systems, Work Order No. 1410/AICTE:AIES.}} 
\author{Ambuj Mahanti, Supriyo Ghose and Samir K. Sadhukhan \\
Indian Institute of Management Calcutta \\
D. H. Road, Joka, Kolkata 700 104, India.\\                                         
email: {am@iimcal.ac.in, supriyo@iimcal.ac.in, samir@iimcal.ac.in}}

\newtheorem{th}{Theorem}[section]

\newtheorem{lem}{Lemma}[section]
\newtheorem{defi}{Definition}[section]
\newtheorem{rem}{Remark}[section]

\newcounter{S1ct1}
\newcounter{S1ct2}

\newcounter{S2ctr1}
\newcounter{S2ctr2}
\newcounter{S2ctr3}
\newcounter{S2ctr4}

\newcounter{defctr2}

\newcounter{defctr4}
\newcounter{defctr6}
\newcounter{defctr7}

\newcounter{defctr9}

\newcommand{\qed}{\hfill $\Box$ }

\begin{document}          
\pagenumbering{arabic}

\maketitle

\begin{abstract}

Search in cyclic AND/OR graphs was traditionally known to be an unsolved problem. In the recent past several important studies have been reported in this domain. In this paper, 
we have taken a fresh look at the
problem. First, a new and comprehensive theoretical framework for cyclic AND/OR graphs has been
presented, which was found missing in the recent literature. Based on
this framework, two best-first search algorithms, S1 and S2, have been developed.
S1 does uninformed search and is a simple modification of the Bottom-up
algorithm by Martelli and Montanari. S2 performs a heuristically guided search and replicates the modification in Bottom-up's
successors, namely HS and $\mbox{AO}^{*}$. Both S1 and S2 solve
the problem of searching AND/OR graphs in presence of cycles. We then present a detailed analysis for the correctness and complexity results of S1 and S2, using the proposed framework. We have observed through experiments that S1 and S2 output correct results in all cases.

\end{abstract}

\section {Introduction}

AND/OR graphs [Chang and Slagle 1971; Martelli and Montanari 1973, 1978; Levi and Sirovich 1976; Nilsson 1980; Bagchi and Mahanti 1983; Pearl 1984; Mahanti and Bagchi 1985; Kumar 1991] are generalizations of directed graphs used in the problem-decomposition approach in artificial intelligence. In an AND/OR graph, a node represents a problem to be solved which can be decomposed into several smaller subproblems, which, in turn, may be broken down into even smaller subproblems and so on. The basic objective in searching an AND/OR graph is to find a {\em solution graph} of {\em least cost} following a cost criterion defined suitably.

Till date, the most famous among AND/OR graph algorithms is $\mbox{AO}^{*}$ [Nilsson 1980], which follows the principle of {\em best-first} search and uses an {\em admissible} or lower bound heuristic. There has also been work  on other issues, such as searching in the presence of inadmissible heuristics [Mahanti and Bagchi 1985; Chakrabarti, Ghosh and DeSarkar 1988] and searching in limited memory [Chakrabarti, Ghosh, Acharya and DeSarkar 1990]. However, a critical assumption in all of these work is that, {\em the underlying AND/OR graph must be acyclic.} Without this assumption - i.e. in the presence of cycles - the AND/OR graph search becomes much more complicated. There is an initial difficulty in defining a solution graph in a cyclic AND/OR graph. There are other problems in using an algorithm like $\mbox{AO}^{*}$ which is based on the technique of arc-marking, such as it may end up marking a cycle. These difficulties had led to the assumption of acyclicity, which permeated the whole of AND/OR graph literature and was re-emphasized in [Nilsson 1980].

It can be easily observed that the assumption of acyclicity is not always acceptable in real life problems. For instance, in logic programming every logical equivalence represents a cycle in the corresponding AND/OR graph. There are several other applications where cyclic AND/OR graph formulations are useful, such as in assembly/disassembly sequences [DeMello and Sanderson 1991; Jim$\acute{e}$nez and Torras 2000] and robotic task plans [Cao and Sanderson 1998]. This problem of searching AND/OR graphs in the presence of cycles has attracted the attention of researchers for a long time, and in recent times several algorithms have been reported [Chakrabarti 1994; Hvalica 1996; Jim$\acute{e}$nez and Torras 2000] that solve AND/OR graphs containing cycles. However a common issue with all of these studies is that, they do not provide a theoretical framework for cyclic AND/OR graphs. They are based mostly on the existing framework for acyclic AND/OR graphs, which makes it difficult to actually prove any of the properties of the algorithms. 

The major contributions of this paper are as follows: \footnote {Our study involves the search for solution graphs that {\em do not} contain cycles. This is the conventional direction of AND/OR graph work, as orthogonal to a recent study [Hansen and Zilberstein 1998] in which the authors work on a variation of the conventional model that allows cycles in the solution graphs.}

\begin{enumerate}

\item We provide a new theoretical framework for cyclic AND/OR graphs, which subsumes the existing framework for acyclic graphs. This framework redefines all the basic concepts such as solution graph, cost of solution and admissible heuristics taking into consideration the presence of cycles.

\item We present two algorithms, S1 and S2, for searching AND/OR graphs with cycles. Algorithm S1 searches the implicit graph without constructing an explicit graph, while S2 is an $\mbox{AO}^{*}$-style algorithm that, at each stage, calls an explicit-graph-growing outer loop and a cost-computing inner loop. S2 uses heuristic estimates of nodes to guide the search intelligently, which S1 does not. S1 and S2 are easy to understand and are designed using the well-known principle of best-first search. 

\item We discuss in detail the theoretical properties of S1 and S2. These theoretical properties are proved using the proposed AND/OR graph framework mentioned above.

\item Finally, the experimental results presented in this paper shed some light on the performance of the various algorithms both in acyclic and cyclic cases, and provide further insights into the development of algorithms for AND/OR graphs. 

\end{enumerate}

\section {Definitions and Previous Work}

In this section, we make a brief survey of the existing literature on AND/OR graphs. \footnote{Any term used here but not explicitly defined will follow the meaning as given in [Nilsson 1980].}

\subsection{Definitions} 

An AND/OR graph $G$ is a directed graph where a node represents a problem to be
solved, and its immediate successor nodes represent the subproblems into which
the parent problem can be transformed or decomposed. $G$ contains a special node, the {\em start node}, 
that represents the initial (root) problem to be solved. $G$ also contains a set of {\em leaf
nodes}, which are of two types: terminal and nonterminal. While the {\em terminal leaf nodes} represent
subproblems with known solutions, the {\em nonterminal leaf nodes} represent subproblems
which are not solvable. Each node has finitely many children. Any node can be either an AND node or an OR node. An OR node can be solved by solving any of its children, while an AND node can be solved by solving all of its children. Without any loss of generality we assume that all leaf nodes are OR nodes. (It is
important to note that these definitions of AND and OR nodes are in line with
[Pearl 1984] and [Martelli and Montanari 1973, 1978], but not same as in [Nilsson 1980], which uses the concept of {\em k-connectors}.)

We represent the start node by $s$, the terminal leaves by $t$, $t_1$, $t_2$, ...
and all other nodes by $m$, $n$, $p$, $q$, $r$,.... For the sake of completeness,
we allow the start node $s$ to be either a terminal leaf, or a nonterminal leaf, or any internal node
of $G$. The set of terminal leaves is denoted by $T$, and the set of nonterminal leaves by $NT$. Thus $T \bigcup NT$ is the set of all leaf nodes in $G$.

Each arc in $G$ represents the application of a production rule. Generally a cost is associated per rule application - thus each arc $(m,n)$ of $G$ has a cost $c(m,n) \geq \delta > 0$ associated with it, where $\delta$ is a small positive number.

Let $G$ be an acyclic AND/OR graph and $m$ be a node in $G$. A {\em solution graph} $D(m)$ rooted at or below $m$ is a finite subgraph of $G$ that represents a complete solution to $m$. It is defined as follows:

\begin{enumerate}

\item $m \in D(m)$;
\item if $n$ is an OR node in $G$ and $n$ is in $D(m)$, then exactly one of its immediate successors in $G$ is in $D(m)$;
\item if $n$ is an AND node in $G$ and $n$ is in $D(m)$, then all of its immediate successors in $G$ are in $D(m)$;
\item every maximal directed path in $D(m)$ ends in a terminal leaf node.

\end{enumerate}

Note that there may be multiple solution graphs rooted at a node, and each solution graph satisfies the definition of an AND/OR graph. A function $h(n,D(m))$ assigns a cost value to each of the nodes $n$ in $D(m)$ as follows:

\begin{enumerate}
\item $h(n,D(m)) = 0$ if $n$ is a terminal leaf node;
\item $h(n,D(m)) = c(n,n^{\prime}) + h(n^{\prime},D(m))$ if $n$ is an OR node and $n^{\prime}$ is its immediate successor in $D(m)$;
\item $h(n,D(m)) = \sum_{i=1}^{k} [c(n,n_i) + h(n_i,D(m))]$ if $n$ is an AND node with immediate successors $n_1,..,n_k$ in $D(m)$.
\end{enumerate}

Thus, $h(m,D(m))$ is the cost of a solution graph $D(m)$ below $m$. If $m$ has one or more solution graphs below it, then the cost of a minimal-cost solution graph is denoted by $h^{*}(m)$. If $m$ has no solution graph below it, $h^{*}(m)$ is taken to be $\infty$.
Thus $h^{*}(s)$ is the cost of a minimal cost solution graph below $s$. A search algorithm is required to output a solution graph below $s$ with minimal cost. The definition of cost used here represents the {\em sum-cost criterion}; there is another alternative, called the {\em max-cost criterion} which differs in the way that the cost of an AND node $n$ is evaluated as the maximum of $c(n,n_i) + h(n_i,D(m))$, evaluated over all its children $n_i$. 

As described above, an AND/OR graph is implicitly defined by the root node $s$,
set of production rules and their costs, a heuristic function, and a set of terminal
and nonterminal leaf nodes. This is called the {\em implicit graph}, $G$, and a search
algorithm usually works by constructing a subgraph of the implicit graph,
called the {\em explicit graph} $G^{\prime}$. Initially only the root node $s$ belongs to the explicit graph $G^{\prime}$. Once $s$ is expanded, its children and all their connecting arcs are added to $G^{\prime}$. $G^{\prime}$ grows as more and more nodes are expanded and new nodes and arcs are added to it. At any instant the nodes of $G^{\prime}$ which have no children are called {tip nodes}.

In the study of acyclic AND/OR graphs, the notion of a {\em potential solution graph (psg)} of an explicit graph is very similar to the notion of a solution graph of an implicit AND/OR graph. A psg $D^{\prime}(m)$ below a node $m$ in $G^{\prime}$ is a finite subgraph of $G^{\prime}$ with the following properties:

\begin{enumerate}

\item $m \in D^{\prime}(m)$;
\item For every node $n \in D^{\prime}(m)$ which is not a tip node of $G^{\prime}$:

\begin{enumerate}
\item exactly one of the immediate successors of $n$ in $G^{\prime}$ is in $D^{\prime}(m)$ when $n$ is an OR node;
 
\item all of its immediate successors in $G^{\prime}$ are in $D^{\prime}(m)$ when $n$ is an AND node;

\end{enumerate}

\item every maximal directed path in $D^{\prime}(m)$ ends in a tip node of $G^{\prime}$.

\end{enumerate}

Occasionally a non-negative heuristic function $\hat{h}(n)$ is defined for each node $n$ in $G$. The value of this heuristic function is an estimate of the cost of solving that node, and is used in search algorithms to guide the search process towards least-costly solution graphs. A heuristic function is called {\em admissible} if $\hat{h}(n) \le h^{*}(n) \forall n \in G.$ The heuristic estimate is $0$ for a terminal leaf node and $\infty$ for a nonterminal leaf node.

The cost of a node $n$ in a psg $D^{\prime}(m)$, denoted as $h(n,D^{\prime}(m))$, is defined in an identical manner as the cost of $n$ in a solution graph $D(m)$, with the first condition being replaced by $h(n,D^{\prime}(m)) = \hat{h}(n)$ if $n$ is a tip node in $G^{\prime}$. The cost of a minimal-cost psg below node $n$ in $G^{\prime}$ is denoted as $h^{\prime}(n)$.

\subsection{Algorithms for Acyclic AND/OR Graphs}

One of the early algorithms on AND/OR graphs was the Bottom-up algorithm [Martelli and Montanari 1973]. This algorithm, which is an extension of the shortest-path algorithm [Dijkstra 1959], operates on the (entire) implicit AND/OR graph
and evaluates the nodes according to a dynamic schedule determined by the
cost-dominance of nodes. It was followed by an improved algorithm, HS [Martelli and Montanari 1978], that takes heuristic information into account and works by creating an explicit graph. HS is a top-down iterative method that first constructs an explicit graph $G^{\prime}$ with only node $s$. In each iteration, HS chooses a tip node of $G^{\prime}$ (initially the start node $s$) for expansion, and adds its children with their connecting arcs to $G^{\prime}$. Then a bottom-up cost revision process is performed, whereby at each OR node one of its least-costly children is chosen and the corresponding arc is marked; at each AND node all the children are chosen and their arcs are marked. At the end of this process, below every node $n$ in $G^{\prime}$, a complete psg $D^{\prime}(n)$ is marked. This marked psg $D^{\prime}(n)$ is also a least costly psg below $n$. In this manner the iterations continue, until it is found that the marked psg below $s$ is a solution graph which is then outputted by HS, or the marked psg contains a nonterminal leaf node when failure termination is reported. Algorithm HS was modified by introducing the concept of $k$-connectors and renamed as algorithm $\mbox{AO}^{*}$ [Nilsson 1980]. $\mbox{AO}^{*}$ has been traditionally used to find minimal-cost solutions to AND/OR graphs.

It was proved by [Martelli and Montanari 1978] that HS outputs minimal-cost solutions
if the heuristic function satisfies the {\em monotone restriction}. [Bagchi and Mahanti 1983] generalized this result and showed that HS output
minimal-cost solutions even if the heuristic function merely satisfies the weaker condition of {\em admissibility}. This generalization easily carries over to $\mbox{AO}^{*}$ also.

However, as has been widely documented, $\mbox{AO}^{*}$ can fail if the AND/OR graph
contains cycles, which we illustrate in Figures \ref{fig10}(a) and \ref{fig10}(b). In both figures, $t$ is a terminal leaf node and marked arcs are
crossed by a line.

\unitlength=0.42mm
\special{em:linewidth 0.4pt}
\linethickness{0.4pt}
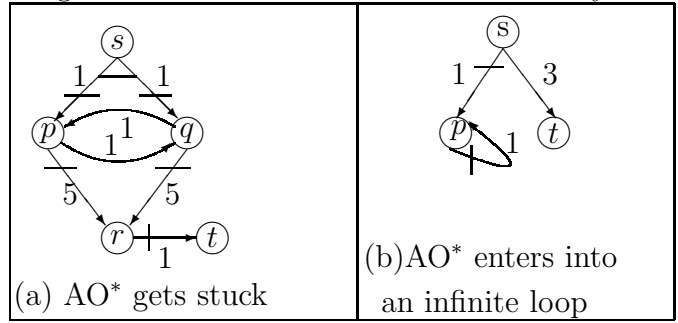
\begin{figure}
\centering
\caption{Problems with $\mbox{AO}^{*}$ in case of cycles} 
\label{fig10}
\begin{picture}(210.00,100.00)
\put(56.00,59.00){\circle{10.20}}
\put(12.00,59.00){\makebox(0,0)[cc]{$p$}}
\put(34.00,26.00){\makebox(0,0)[cc]{$r$}}
\put(34.00,88.00){\makebox(0,0)[cc]{$s$}}
\put(56.00,59.00){\makebox(0,0)[cc]{$q$}}
\put(19.00,40.00){\makebox(0,0)[cc]{5}}
\put(49.00,40.00){\makebox(0,0)[lc]{5}}
\put(156.00,86.00){\vector(-2,-3){14.67}}
\put(156.00,86.00){\vector(3,-4){16.67}}
\put(142.00,60.00){\makebox(0,0)[cc]{$p$}}
\put(171.00,78.00){\makebox(0,0)[cc]{3}}
\put(142.00,78.00){\makebox(0,0)[cc]{1}}
\put(159.00,56.00){\makebox(0,0)[cc]{1}}
\put(156.00,92.00){\makebox(0,0)[cc]{s}}
\put(34.00,26.00){\circle{10.00}}
\put(12.00,59.00){\circle{10.00}}
\put(34.00,88.00){\circle{10.00}}
\put(156.00,91.00){\circle{10.00}}
\put(141.00,59.00){\circle{10.00}}
\bezier{296}(139.00,54.00)(174.00,40.00)(145.00,62.00)
\put(151.00,57.00){\vector(-1,1){6.00}}
\bezier{164}(17.00,61.00)(34.00,72.00)(52.00,61.00)
\bezier{172}(16.00,56.00)(34.00,44.00)(52.00,56.00)
\put(48.00,53.00){\vector(1,1){3.00}}
\put(39.00,26.00){\vector(1,0){20.00}}
\put(49.00,23.00){\makebox(0,0)[ct]{1}}
\put(32.00,52.00){\makebox(0,0)[cb]{1}}
\put(34.00,83.00){\vector(-1,-1){20.00}}
\put(34.00,83.00){\vector(1,-1){19.00}}
\put(56.00,54.00){\vector(-3,-4){18.00}}
\put(12.00,54.00){\vector(3,-4){18.00}}
\put(28.00,77.00){\line(1,0){12.00}}
\put(22.00,74.00){\makebox(0,0)[cb]{1}}
\put(46.00,74.00){\makebox(0,0)[lb]{1}}
\put(37.00,64.00){\makebox(0,0)[ct]{1}}
\put(21.00,63.00){\vector(-2,-1){4.00}}
\put(0.00,0.00){\framebox(110.00,100.00)[cc]{}}
\put(11.00,48.00){\line(1,0){10.00}}
\put(46.00,48.00){\line(1,0){11.00}}
\put(44.00,30.00){\line(0,-1){8.00}}
\put(17.00,71.00){\line(1,0){11.00}}
\put(41.00,71.00){\line(1,0){10.00}}
\put(1.00,7.00){\makebox(0,0)[lc]{(a) $\mbox{AO}^{*}$ gets stuck}}
\put(110.00,0.00){\framebox(100.00,100.00)[cc]{}}
\put(147.00,80.00){\line(1,0){9.00}}
\put(146.00,46.00){\line(0,1){9.00}}
\put(152.00,19.00){\makebox(0,0)[cc]{(b)$\mbox{AO}^{*}$ enters into}}
\put(150.00,5.00){\makebox(0,0)[cc]{an infinite loop}}
\put(64.00,26.00){\circle{10.00}}
\put(64.00,26.00){\makebox(0,0)[cc]{$t$}}
\put(172.00,59.00){\circle{10.20}}
\put(172.00,59.00){\makebox(0,0)[cc]{$t$}}
\end{picture}
\end{figure}

In the explicit graph shown in Figure \ref{fig10}(a), $\mbox{AO}^{*}$ expands nodes $s$, $p$ and $q$ in the first three iterations and then expands $r$. Now during the bottom-up computation, it gets stuck with nodes $p$ and $q$. The bottom-up computation uses a list $Z$ [Nilsson 1980, pp 104, Step 10] to ensure that the nodes being evaluated are selected in a topological order. However in this example, after expansion of $r$ both $p$ and $q$ will be present in $Z$ and none can be selected for evaluation as it is the predecessor of the other. Hence $\mbox{AO}^{*}$ gets stuck,
even though there is a solution graph of cost 14 below $s$.

In the explicit graph shown in Figure \ref{fig10}(b), on the other hand, $\mbox{AO}^{*}$ first expands the root node
$s$ and marks the arc ($s$,$p$) since $p$ is the least-costly child below $s$. Then, after expanding the node $p$, it marks the arc ($p$,$p$). Now, during the bottom-up cost revision process it tries to go upward following all marked arcs above $p$, thereby entering into an infinite
self-loop. However, there exists a solution graph from $s$ to the terminal leaf $t$ with a cost of 3. 

These examples clearly show that in the presence of cycles,
$\mbox{AO}^{*}$ may never even
terminate, let alone find a minimal-cost solution.

\subsection{Algorithms for Cyclic AND/OR Graphs}

The first attempt in searching cyclic AND/OR graphs was made in 1994 when two algorithms, Iterative\_revise and $\mbox{REV}^{*}$ were presented [Chakrabarti 1994]. While Iterative\_revise is a top-down recursive algorithm for searching AND/OR graphs, $\mbox{REV}^{*}$ is a strictly bottom-up algorithm that uses parent-pointers and has a better performance than Iterative-Revise. Its operation is briefly described below.

$\mbox{REV}^{*}$ starts searching the AND/OR graph by putting all its leaf nodes in a list called OPEN and assigning heuristic values to them. It then does the following work iteratively: takes out a least-costly node from OPEN, assigns cost values to its {\em parent} nodes by using parent-pointers and goes up the graph as long as the siblings of the current level node have all been evaluated. When it gets stuck in this upward phase, it again selects a node from OPEN and starts another upward phase to evaluate nodes. In this way $\mbox{REV}^{*}$ continues until the start node has been evaluated. Cycles get eliminated due to the cost dominance rule, and $\mbox{REV}^{*}$ outputs a correct solution cost whenever there is a solution graph. 

$\mbox{REV}^{*}$ is a simple algorithm that solves the long-standing problem of searching explicit AND/OR graphs in the presence of cycles. Recently, it has been shown [Jim$\acute{e}$nez and Torras 2000] that it is possible to improve the efficiency of $\mbox{REV}^{*}$ significantly by making some modifications. These modifications derive partly from an earlier work in the {\em acyclic} domain to improve the efficiency of $\mbox{AO}^{*}$ namely, algorithm CF [Mahanti and Bagchi 1985]. CF's control structure for node expansion has been utilized in [Jim$\acute{e}$nez and Torras 2000] in their algorithm INT. As the authors observe, INT's "top-down search strategy is based on Mahanti and Bagchi's CF, whereas its bottom-up cost revision process is inspired in Chakrabarti's $\mbox{REV}^{*}$." The bottom-up cost-revision process of INT is primarily based on $\mbox{REV}^{*}$; but in the process, it also employs the superior cost-updation strategy of CF. This makes INT a feasible alternative for searching cyclic AND/OR graphs. However, it still
has the inefficiency that nodes are considered for cost-revision even when they are not likely to be affected as a result of the current node-expansion. As it is well-known in $\mbox{AO}^{*}$ and CF-like algorithms, it is sufficient and economical to visit only nodes whose costs, arc markings or solved status change as a result of the expansion of a new node. This observation has been implemented in their next algorithm $CFC_{REV^*}$ by creating the OPEN list with only a subset of the leaf nodes. As the OPEN list ultimately decides which nodes are going to be visited in the cost-revision phase of the algorithm, this strategy significantly cuts down the number of nodes evaluated, particularly when the percentage of AND nodes is high. $CFC_{REV^*}$ has been implemented and found to be very efficient compared to its predecessor $\mbox{REV}^{*}$. However, the $CFC_{REV^*}$ algorithm is unwieldy and non-intuitive. 
 
Along with the work on $\mbox{REV}^{*}$ and $CFC_{REV^*}$, one other attempt [Hvalica 1996] has been made to solve cyclic AND/OR graphs.  In Hvalica's algorithm, a special technique is used for loop avoidance. When a node is expanded a dummy node $x_f$ is attached to it with a high heuristic value. If expansion of the current node creates a cycle, the dummy node $x_f$ offers an alternative route to come out of it. The exit from the cycle occurs when the cost of the expanded node, computed through its children, exceeds the high cost of the dummy child $x_f$. This method, although quite interesting, may become inefficient in practice.

\unitlength=0.40mm
\special{em:linewidth 0.4pt}
\linethickness{0.4pt}
\begin{figure}
\centering
\caption{Does $p$ have a solution?} 
\label{fig20}
\begin{picture}(88.00,80.00)
\put(43.00,59.00){\vector(-3,-4){17.33}}
\put(43.00,59.00){\vector(3,-4){17.33}}
\put(37.00,61.00){\vector(1,0){2.00}}
\put(32.00,49.00){\makebox(0,0)[rc]{1}}
\put(53.00,50.00){\makebox(0,0)[lc]{1}}
\put(14.00,55.00){\makebox(0,0)[rc]{1}}
\put(43.00,66.00){\makebox(0,0)[cc]{$s$}}
\put(26.00,30.00){\makebox(0,0)[cc]{$p$}}
\put(60.00,30.00){\makebox(0,0)[cc]{$t$}}
\put(5.00,10.00){\makebox(0,0)[lc]{Implicit graph $G$}}
\put(0.00,0.00){\framebox(90.00,80.00)[cc]{}}
\put(26.00,31.00){\circle{12.00}}
\put(43.00,66.00){\circle{12.00}}
\bezier{232}(19.00,32.00)(9.00,60.00)(37.00,61.00)
\put(60.00,31.00){\circle{12.00}}
\end{picture}
\end{figure}
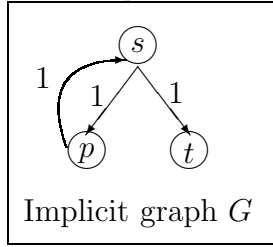

\subsection{Motivation for current work}

Although the recent algorithms on cyclic AND/OR graphs as discussed in the preceding subsection are claimed to be correct,
their correctness proofs stand on a weak theoretical base. This is due to the reason that the theoretical constructs such as solution graph, potential solution graph, cost of solution etc. which are fundamental to the AND/OR graph algorithms and their properties, have never been properly re-established in the context of cycles. For instance, if we look at Figure \ref{fig20}, we find this interesting question: does node $p$ have a solution graph below it? Looking from the top (i.e. from start node $s$) it would appear not (as that would create a cycle with predecessor $s$); however looking from the node $p$ itself there is a solution graph below it passing through node $s$. So how do we construct a solution graph in the presence of cycles? How do we define the cost function appropriately? These questions have never been adequately addressed in the literature. The recent papers on cyclic AND/OR graphs [Chakrabarti 1994; Hvalica 1996; Jim$\acute{e}$nez and Torras 2000] seem to have completely ignored this issue and worked with cyclical constructs. \footnote{Examples of such cyclical definitions are presented in the Appendix for the purpose of review.} 

\unitlength=0.42mm
\special{em:linewidth 0.4pt}
\linethickness{0.4pt}
\begin{figure}
\centering
\caption{Combining solution graphs may lead to cycles} 
\label{fig25}
\begin{picture}(240.00,120.00)
\put(56.00,79.00){\circle{10.20}}
\put(12.00,79.00){\makebox(0,0)[cc]{$p$}}
\put(34.00,46.00){\makebox(0,0)[cc]{$r$}}
\put(56.00,79.00){\makebox(0,0)[cc]{$q$}}
\put(49.00,60.00){\makebox(0,0)[lc]{5}}
\put(34.00,46.00){\circle{10.00}}
\put(12.00,79.00){\circle{10.00}}
\bezier{172}(16.00,76.00)(34.00,64.00)(52.00,76.00)
\put(48.00,73.00){\vector(1,1){3.00}}
\put(39.00,46.00){\vector(1,0){20.00}}
\put(49.00,43.00){\makebox(0,0)[ct]{1}}
\put(32.00,72.00){\makebox(0,0)[cb]{1}}
\put(56.00,74.00){\vector(-3,-4){18.00}}
\put(1.00,27.00){\makebox(0,0)[lc]{(a)Solution graph}}
\put(64.00,46.00){\circle{10.00}}
\put(64.00,46.00){\makebox(0,0)[cc]{$t$}}
\put(136.00,79.00){\circle{10.20}}
\put(92.00,79.00){\makebox(0,0)[cc]{$p$}}
\put(114.00,46.00){\makebox(0,0)[cc]{$r$}}
\put(136.00,79.00){\makebox(0,0)[cc]{$q$}}
\put(99.00,60.00){\makebox(0,0)[cc]{5}}
\put(114.00,46.00){\circle{10.00}}
\put(92.00,79.00){\circle{10.00}}
\bezier{164}(97.00,81.00)(114.00,92.00)(132.00,81.00)
\put(119.00,46.00){\vector(1,0){20.00}}
\put(129.00,43.00){\makebox(0,0)[ct]{1}}
\put(92.00,74.00){\vector(3,-4){18.00}}
\put(117.00,84.00){\makebox(0,0)[ct]{1}}
\put(101.00,83.00){\vector(-2,-1){4.00}}
\put(82.00,27.00){\makebox(0,0)[lc]{(b)Solution graph}}
\put(144.00,46.00){\circle{10.00}}
\put(144.00,46.00){\makebox(0,0)[cc]{$t$}}
\put(216.00,79.00){\circle{10.20}}
\put(172.00,79.00){\makebox(0,0)[cc]{$p$}}
\put(194.00,46.00){\makebox(0,0)[cc]{$r$}}
\put(194.00,108.00){\makebox(0,0)[cc]{$s$}}
\put(216.00,79.00){\makebox(0,0)[cc]{$q$}}
\put(179.00,60.00){\makebox(0,0)[cc]{5}}
\put(209.00,60.00){\makebox(0,0)[lc]{5}}
\put(194.00,46.00){\circle{10.00}}
\put(172.00,79.00){\circle{10.00}}
\put(194.00,108.00){\circle{10.00}}
\bezier{164}(177.00,81.00)(194.00,92.00)(212.00,81.00)
\bezier{172}(176.00,76.00)(194.00,64.00)(212.00,76.00)
\put(208.00,73.00){\vector(1,1){3.00}}
\put(199.00,46.00){\vector(1,0){20.00}}
\put(209.00,43.00){\makebox(0,0)[ct]{1}}
\put(192.00,72.00){\makebox(0,0)[cb]{1}}
\put(194.00,103.00){\vector(-1,-1){20.00}}
\put(194.00,103.00){\vector(1,-1){19.00}}
\put(216.00,74.00){\vector(-3,-4){18.00}}
\put(172.00,74.00){\vector(3,-4){18.00}}
\put(188.00,97.00){\line(1,0){12.00}}
\put(182.00,94.00){\makebox(0,0)[cb]{1}}
\put(206.00,94.00){\makebox(0,0)[lb]{1}}
\put(197.00,84.00){\makebox(0,0)[ct]{1}}
\put(181.00,83.00){\vector(-2,-1){4.00}}
\put(161.00,27.00){\makebox(0,0)[lc]{(c)Combination}}
\put(224.00,46.00){\circle{10.00}}
\put(224.00,46.00){\makebox(0,0)[cc]{$t$}}
\put(0.00,0.00){\framebox(80.00,120.00)[cc]{}}
\put(32.00,12.00){\makebox(0,0)[cc]{below $p$}}
\put(80.00,0.00){\framebox(80.00,120.00)[cc]{}}
\put(115.00,12.00){\makebox(0,0)[cc]{below $q$}}
\put(190.00,12.00){\makebox(0,0)[cc]{below $s$}}
\put(160.00,0.00){\framebox(80.00,120.00)[cc]{}}
\end{picture}
\end{figure}
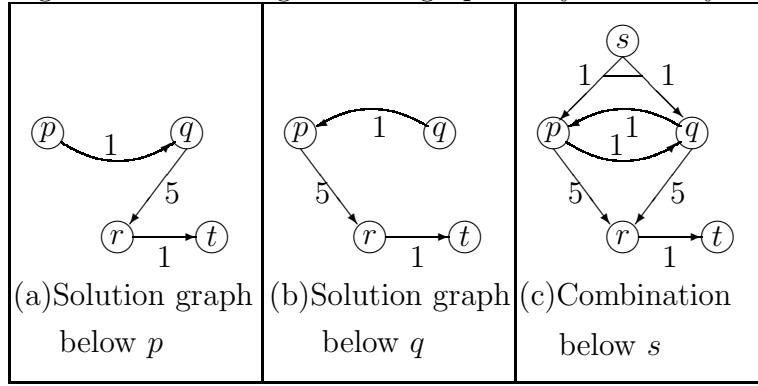

The graphs in Figure \ref{fig25} give another example of the necessity of theoretical support for cyclic AND/OR graph algorithms. The graphs in Figure \ref{fig25}(a) and \ref{fig25}(b) are solution graphs below $p$ and $q$ respectively, in the implicit graph of Figure \ref{fig10}(a); however if they are combined below $s$ as in Figure \ref{fig25}(c), they clearly do not form a solution graph below $s$. What is more fatal is that they create a cycle. The question then arises: can such solution graphs be at all combined (in an appropriate way) to form a solution graph below the parent? We delve deep into these issues in this paper.

We first provide a theoretical framework for cyclic AND/OR graphs, which forms the conceptual basis for our algorithms. We then present two algorithms, S1 and S2, which are based on the time-tested principle of best-first search (the previous algorithms do not seem to have adhered to this principle). The analysis of S1 and S2 have been done in detail using our theoretical framework. We conclude the paper with some comments on the results of our detailed experiments.

\section{Framework for Cyclic AND/OR Graphs}

In this section we generalize the existing AND/OR graph search framework  
for graphs containing cycles. In the proposed new framework, the concept of Maximal Extendable Subgraph (MES) plays a pivotal role.

An MES below a node is constructed in a top-down manner as explained below. During the construction, at an OR node $x$ we take one child and include it if it does not form a cycle with the part of the MES constructed so far. If $x$ has no child, or if the selected child forms a cycle, the construction ceases to proceed below $x$. When $x$ is an AND
node, construction continues below $x$ only if no child of $x$ forms a cycle with the part of the MES constructed so far. 

\subsection{MESs in an Implicit AND/OR Graph}

\begin{defi} \label{def00} For any AND/OR graph $\mathcal{G}$ (implicit, explicit or other) we define the following:

\begin{list}{\roman{defctr2})}{\usecounter{defctr2}}
   \item  For any node $p$ in $\mathcal{G}$, 
    the child\_set is $\Gamma(p, \mathcal{G})$ = \{$q \mid q$ is a child of $p$ in $\mathcal{G}$\}.
   \item  The set of all nodes of $\mathcal{G}$ which have no children is $Z_{\mathcal{G}} = \{x \mid x \in \mathcal{G} \bigwedge$$ \Gamma(x,\mathcal{G}) = \phi \}.$   
 \end{list}
\end{defi} 

For example, a leaf node, say $x$, of an implicit graph $G$ is one
          for which $\Gamma(x,G) = \phi$.
Again, in $G$, $Z_{G} = T \bigcup NT$ = set of all terminal and nonterminal leaf nodes. 

\begin{defi}  Let $G$ be an implicit AND/OR graph and $n$ be a node in $G$. A Maximal Extendable Subgraph (MES) $M(n,G)$, rooted at or below $n$, is defined as follows:

\begin{list}{\roman{defctr2})}{\usecounter{defctr2}} 
   \item $n \in M(n,G)$
   \item For every node $x \in M(n,G)$ which is an internal node in $G$:
\newline 
Let $y_1$, $y_2$,...,$y_k$ be the children of $x$ in $G$. Now,
    \begin{list}{\alph{defctr7})}{\usecounter{defctr7}}
    \item If $x$ is an OR node, then select any one child $y_i$, $1 \leq i \leq k$. Include $y_i$ in $M(n,G)$ if it is not same as $x$ or any predecessor of $x$ in $M(n,G)$. Otherwise the construction ceases to proceed below $x$.
    \item If $x$ is an AND node, then include all $y_i$, $1 \leq i \leq k$, if none of them is same as $x$ or any predecessor of $x$ in $M(n,G)$. Otherwise the construction ceases to proceed below $x$.
\end{list}
\end{list}
\end{defi}

For any node $n \in G$, the MESs below it are enumerated as
$M_1(n,G), M_2(n,G), \ldots$. On the other hand, occasionally we may write $M(n,G)$ as $M(n)$ or just as $M$, when the parameters are clear from context.

\begin{rem} 
\end{rem}
\begin{description}
\item[({\em i})] If $G$ contains paths of infinite length, then it is clear from the
construction that there may be MESs with infinitely many nodes and arcs.
\item[({\em ii})] From the definition, it is clear that MESs cannot contain cycles.
\item[({\em iii})] There must exist at least one MES below every node in $G$.
\end{description}

\begin{figure}
\centering
\caption{An Implicit Graph and its MESs}
\label{fig26}
\special{em:linewidth 0.4pt}
\linethickness{0.4pt}
\unitlength=0.45mm
\begin{picture}(390.00,140.00)
\put(21.00,90.00){\circle{10.20}}
\put(21.00,90.00){\makebox(0,0)[cc]{$s$}}
\put(24.00,18.00){\makebox(0,0)[cc]{(a) Implicit}}
\put(24.00,8.00){\makebox(0,0)[cc]{Graph $G$}}
\put(21.00,60.00){\circle{10.00}}
\put(21.00,30.00){\circle{10.00}}
\put(21.00,85.00){\vector(0,-1){20.00}}
\put(21.00,55.00){\vector(0,-1){20.00}}
\put(52.00,60.00){\circle{10.20}}
\put(21.00,85.00){\vector(3,-2){29.00}}
\bezier{240}(21.00,55.00)(48.00,43.00)(25.00,63.00)
\put(29.00,60.00){\vector(-3,2){3.00}}
\put(14.00,87.00){\vector(1,1){2.00}}
\put(21.00,60.00){\makebox(0,0)[cc]{$n$}}
\put(52.00,60.00){\makebox(0,0)[cc]{$t$}}
\put(21.00,30.00){\makebox(0,0)[cc]{$r$}}
\bezier{272}(22.00,85.00)(56.00,77.00)(26.00,92.00)
\put(28.00,91.00){\vector(-2,1){2.00}}
\put(83.00,105.00){\circle{10.00}}
\put(83.00,105.00){\makebox(0,0)[cc]{$s$}}
\put(88.00,81.00){\makebox(0,0)[cc]{(b)$M_1(s,G)$}}
\put(138.00,125.00){\circle{10.00}}
\put(149.00,104.00){\circle{10.00}}
\put(138.00,120.00){\vector(1,-1){10.00}}
\put(139.00,81.00){\makebox(0,0)[cc]{(c)$M_2(s,G)$}}
\put(138.00,125.00){\makebox(0,0)[cc]{$s$}}
\put(149.00,104.00){\makebox(0,0)[cc]{$t$}}
\put(190.00,127.00){\circle{10.00}}
\put(190.00,106.00){\circle{10.00}}
\put(190.00,127.00){\makebox(0,0)[cc]{$s$}}
\put(190.00,106.00){\makebox(0,0)[cc]{$n$}}
\put(190.00,122.00){\vector(0,-1){11.00}}
\put(190.00,81.00){\makebox(0,0)[cc]{(d)$M_3(s,G)$}}
\put(295.00,123.00){\circle{10.00}}
\put(295.00,102.00){\circle{10.00}}
\put(295.00,123.00){\makebox(0,0)[cc]{$s$}}
\put(295.00,102.00){\makebox(0,0)[cc]{$n$}}
\put(295.00,118.00){\vector(0,-1){11.00}}
\put(297.00,81.00){\makebox(0,0)[cc]{(f)$M_5(s,G)$}}
\put(238.00,60.00){\circle{10.00}}
\put(238.00,38.00){\circle{10.00}}
\put(238.00,18.00){\circle{10.00}}
\put(238.00,55.00){\vector(0,-1){12.00}}
\put(238.00,33.00){\vector(0,-1){10.00}}
\put(238.00,18.00){\makebox(0,0)[cc]{$t$}}
\put(238.00,38.00){\makebox(0,0)[cc]{$s$}}
\put(238.00,60.00){\makebox(0,0)[cc]{$n$}}
\put(88.00,6.00){\makebox(0,0)[cc]{(h)$M_1(n,G)$}}
\put(81.00,42.00){\circle{10.00}}
\put(80.00,41.00){\makebox(0,0)[cc]{$n$}}
\put(242.00,6.00){\makebox(0,0)[cc]{(k)$M_4(n,G)$}}
\put(136.00,45.00){\circle{10.00}}
\put(136.00,24.00){\circle{10.00}}
\put(136.00,45.00){\makebox(0,0)[cc]{$n$}}
\put(136.00,24.00){\makebox(0,0)[cc]{$r$}}
\put(136.00,40.00){\vector(0,-1){11.00}}
\put(139.00,6.00){\makebox(0,0)[cc]{(i)$M_2(n,G)$}}
\put(191.00,47.00){\circle{10.00}}
\put(191.00,26.00){\circle{10.00}}
\put(191.00,47.00){\makebox(0,0)[cc]{$n$}}
\put(191.00,26.00){\makebox(0,0)[cc]{$s$}}
\put(191.00,42.00){\vector(0,-1){11.00}}
\put(190.00,6.00){\makebox(0,0)[cc]{(j)$M_3(n,G)$}}
\put(296.00,37.00){\circle{10.00}}
\put(296.00,19.00){\circle{10.00}}
\put(296.00,37.00){\makebox(0,0)[cc]{$n$}}
\put(296.00,19.00){\makebox(0,0)[cc]{$s$}}
\put(298.00,6.00){\makebox(0,0)[cc]{(l)$M_5(n,G)$}}
\put(238.00,130.00){\circle{10.00}}
\put(238.00,112.00){\circle{10.00}}
\put(238.00,130.00){\makebox(0,0)[cc]{$s$}}
\put(238.00,112.00){\makebox(0,0)[cc]{$n$}}
\put(238.00,125.00){\vector(0,-1){8.00}}
\put(238.00,92.00){\circle{10.00}}
\put(238.00,107.00){\vector(0,-1){10.00}}
\put(238.00,92.00){\makebox(0,0)[cc]{$r$}}
\put(242.00,81.00){\makebox(0,0)[cc]{(e)$M_4(s,G)$}}
\put(361.00,121.00){\circle{10.00}}
\put(361.00,121.00){\makebox(0,0)[cc]{$t$}}
\put(361.00,81.00){\makebox(0,0)[cc]{(g) $M_1(t,G)$}}
\put(360.00,37.00){\circle{10.00}}
\put(360.00,37.00){\makebox(0,0)[cc]{$r$}}
\put(383.00,6.00){\makebox(0,0)[rc]{(m) $M_1(r,G)$}}
\put(296.00,32.00){\vector(0,-1){8.00}}
\bezier{148}(16.00,59.00)(3.00,71.00)(14.00,87.00)
\put(115.00,140.00){\vector(0,0){0.00}}
\put(115.00,140.00){\vector(0,0){0.00}}
\put(115.00,140.00){\vector(0,0){0.00}}
\put(115.00,140.00){\vector(0,0){0.00}}
\put(0.00,140.00){\line(1,0){154.00}}
\put(114.00,0.00){\line(0,1){140.00}}
\put(0.00,140.00){\line(0,-1){139.00}}
\put(0.00,1.00){\line(0,-1){1.00}}
\put(0.00,0.00){\line(1,0){155.00}}
\put(152.00,140.00){\line(1,0){157.00}}
\put(155.00,0.00){\line(1,0){152.00}}
\put(309.00,140.00){\line(1,0){81.00}}
\put(390.00,140.00){\line(0,-1){140.00}}
\put(390.00,0.00){\line(-1,0){87.00}}
\put(62.00,0.00){\line(0,1){140.00}}
\put(62.00,70.00){\line(1,0){94.00}}
\put(155.00,70.00){\line(1,0){146.00}}
\put(301.00,70.00){\line(1,0){89.00}}
\put(165.00,140.00){\line(0,-1){140.00}}
\put(215.00,140.00){\line(0,-1){140.00}}
\put(329.00,140.00){\line(0,-1){140.00}}
\put(270.00,140.00){\line(0,-1){140.00}}
\end{picture}
\end{figure}
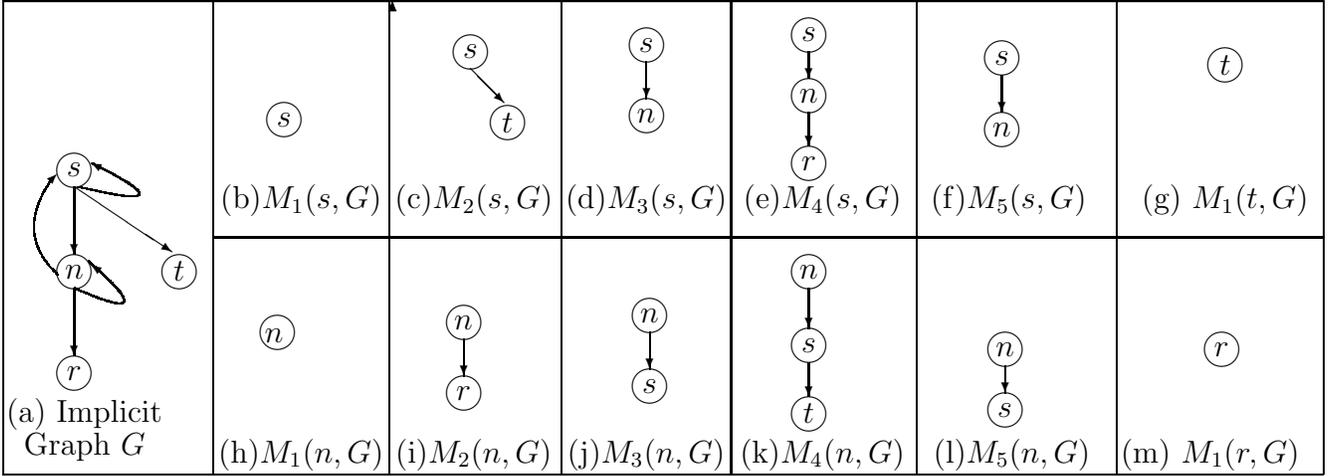

An AND/OR graph $G$ and all of its MESs are shown in Figure \ref{fig26}. It is interesting to note that some MESs may appear to be duplicate - for instance, the MESs $M_3(s,G)$ and $M_5(s,G)$ may appear to be the same even though they are actually different. This happens due to the presence of cycles, as is illustrated in Figure \ref{fig27}. In this figure, the dotted arrows represent the different children that these MESs selected at node $n$. But in either case, the selected child created a cycle and the MES was terminated at node $n$. Thus even though these two MESs appear to be same, they attempted to include different children below a node and hence are distinct. It may also be noted that an MES may appear to be a subgraph of another MES. Thus, $M_3(s,G)$ appears to be a subgraph of $M_4(s,G)$ in Figure \ref{fig26}. This is also explained by the presence of cycles. $M_3(s,G)$ and $M_4(s,G)$ have chosen different children, namely $n$ and $r$, below node $n$ - the former leading to a self-loop (and hence terminating the MES) and the latter continuing to node $r$. Thus the two MESs are distinct and none is a subgraph of the other.

\unitlength=0.45mm
\begin{figure}
\centering
\caption{Duplicate MESs below a node}
\label{fig27}
\linethickness{0.4pt}
\begin{picture}(133.00,82.00)
\put(24.00,70.00){\circle{10.20}}
\put(24.00,35.00){\circle{10.00}}
\put(24.00,35.00){\makebox(0,0)[cc]{$n$}}
\put(35.00,6.00){\makebox(0,0)[cc]{(a) $M_3(s,G)$}}
\put(0.00,0.00){\framebox(64.00,82.00)[cc]{}}
\put(24.00,70.00){\makebox(0,0)[cc]{$s$}}
\put(95.00,20.00){\line(-1,1){8.00}}
\put(86.00,30.00){\line(-2,5){3.67}}
\put(81.00,42.00){\line(0,1){7.00}}
\put(82.00,53.00){\line(3,5){5.00}}
\put(24.00,65.00){\vector(0,-1){25.00}}
\put(34.00,40.00){\line(1,0){5.00}}
\put(43.00,40.00){\line(3,-2){8.00}}
\put(51.00,32.00){\line(0,-1){6.00}}
\put(51.00,24.00){\line(-5,-3){8.00}}
\put(40.00,19.00){\line(-1,0){7.00}}
\put(64.00,0.00){\framebox(69.00,82.00)[cc]{}}
\put(90.00,7.00){\makebox(0,0)[cc]{(b) $M_5(s,G)$}}
\put(100.00,73.00){\circle{10.20}}
\put(100.00,20.00){\circle{10.00}}
\put(100.00,20.00){\makebox(0,0)[cc]{$n$}}
\put(100.00,73.00){\makebox(0,0)[cc]{$s$}}
\put(90.00,65.00){\vector(3,4){4.67}}
\put(100.00,68.00){\vector(0,-1){43.00}}
\put(34.00,39.00){\vector(-3,-2){5.00}}
\put(24.00,30.00){\line(4,-5){6.67}}
\end{picture}
\end{figure}
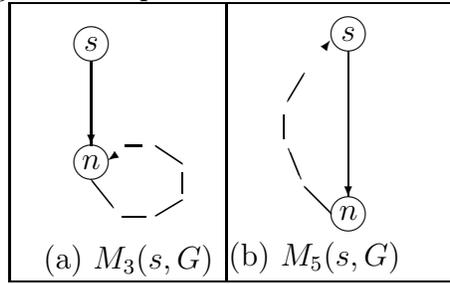

The question naturally arises: what can be the upper bound of MESs in a graph containing $N$ nodes? It may be easily verified that the number of MESs will be maximum if (a) the graph contains only OR nodes, thereby providing more choices at every node, and (b) the graph is structured as in Figure \ref{fig27a} (this graph is the particular instance for $N = 3$). Here, the number of MESs with $k$ nodes, $1 \leq k \leq N$, is $k \times \hspace{6pt} P^{\hspace{-0.55cm}N}_{k}$\hspace{0.1cm}, and the total number of MESs is $\sum_{k=1}^{N}k \times \hspace{6pt} P^{\hspace{-0.55cm}N}_{k}$. 

\unitlength=0.45mm
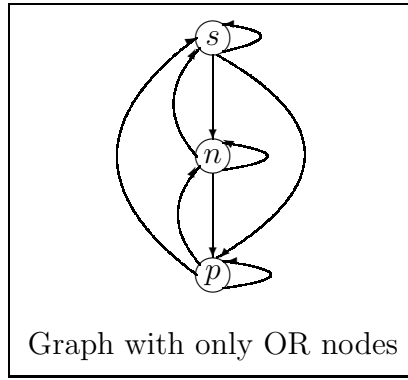
\begin{figure}
\centering
\caption{Maximum number of MESs}
\label{fig27a}
\linethickness{0.4pt}
\special{em:linewidth 0.4pt}
\linethickness{0.4pt}
\begin{picture}(113.00,110.00)
\put(60.00,100.00){\circle{10.20}}
\put(60.00,65.00){\circle{10.00}}
\put(60.00,30.00){\circle{10.00}}
\put(60.00,95.00){\vector(0,-1){25.00}}
\put(60.00,60.00){\vector(0,-1){25.00}}
\bezier{172}(55.00,65.00)(41.00,81.00)(56.00,97.00)
\bezier{156}(56.00,33.00)(43.00,48.00)(56.00,62.00)
\bezier{180}(63.00,96.00)(85.00,101.00)(63.00,104.00)
\bezier{212}(63.00,61.00)(89.00,65.00)(63.00,69.00)
\bezier{228}(63.00,26.00)(91.00,30.00)(63.00,34.00)
\bezier{480}(61.00,95.00)(113.00,66.00)(61.00,35.00)
\bezier{468}(55.00,30.00)(8.00,65.00)(55.00,100.00)
\put(54.00,95.00){\vector(1,2){1.00}}
\put(54.00,60.00){\vector(1,2){1.00}}
\put(66.00,68.50){\vector(-2,1){3.00}}
\put(65.00,34.00){\vector(-1,0){1.00}}
\put(63.00,36.00){\vector(-1,-1){1.00}}
\put(65.00,104.00){\vector(-1,0){2.00}}
\put(53.00,99.00){\vector(2,1){2.00}}
\put(0.00,0.00){\framebox(120.00,110.00)[cc]{}}
\put(60.00,9.00){\makebox(0,0)[cc]{Graph with only OR nodes}}
\put(60.00,100.00){\makebox(0,0)[cc]{$s$}}
\put(60.00,65.00){\makebox(0,0)[cc]{$n$}}
\put(60.00,30.00){\makebox(0,0)[cc]{$p$}}
\end{picture}
\end{figure}

\begin{defi} Let $M$ be an MES below a node $n$ in $G$. Then below every node $x \in M$ we define a sub-MES $\xi(x,M)$ of $M$ identically as $M$ is defined in $G$, by replacing $M$ with $\xi$ and $G$ with $M$ everywhere. 
\end{defi}

\begin{figure}
\centering
\caption{An Implicit Graph and its MESs and Sub-MESs}
\label{fig28}
\unitlength=0.45mm
\special{em:linewidth 0.4pt}
\linethickness{0.4pt}
\begin{picture}(310.00,225.00)
\put(135.00,80.00){\circle{10.00}}
\put(135.00,80.00){\makebox(0,0)[cc]{$n$}}
\put(155.00,50.00){\circle{10.00}}
\put(155.00,50.00){\makebox(0,0)[cc]{$q$}}
\put(135.00,20.00){\circle{10.00}}
\put(135.00,20.00){\makebox(0,0)[cc]{$t$}}
\put(135.00,75.00){\vector(1,-1){20.00}}
\put(155.00,45.00){\vector(-1,-1){20.00}}
\put(135.00,8.00){\makebox(0,0)[cc]{(e) MES $M_1(n,G)$ }}
\put(95.00,0.00){\framebox(95.00,95.00)[cc]{}}
\put(250.00,20.00){\makebox(0,0)[cc]{(f) Sub-MES $\xi(n,M_1(n,G))$,}}
\put(243.00,8.00){\makebox(0,0)[cc]{also an MES in $G$}}
\put(45.00,101.00){\makebox(0,0)[cc]{(a) Implicit Graph $G$}}
\put(26.00,120.00){\circle{10.00}}
\put(51.00,120.00){\circle{10.20}}
\put(76.00,120.00){\circle{10.00}}
\put(36.00,150.00){\circle{10.00}}
\put(64.00,150.00){\circle{10.00}}
\put(51.00,185.00){\circle{10.00}}
\put(51.00,215.00){\circle{10.00}}
\put(51.00,210.00){\vector(0,-1){20.00}}
\put(52.00,180.00){\vector(-2,-3){16.67}}
\put(52.00,180.00){\vector(1,-2){12.67}}
\put(36.00,145.00){\vector(-1,-2){10.00}}
\put(36.00,145.00){\vector(2,-3){13.33}}
\put(64.00,145.00){\vector(-1,-2){10.67}}
\put(64.00,145.00){\vector(1,-2){10.00}}
\put(33.00,139.00){\line(1,0){7.00}}
\bezier{224}(64.00,145.00)(39.00,134.00)(60.00,153.00)
\put(57.00,150.00){\vector(1,1){3.00}}
\bezier{480}(21.00,119.00)(-10.00,174.00)(46.00,183.00)
\put(43.00,182.00){\vector(3,1){3.00}}
\put(0.00,95.00){\framebox(95.00,130.00)[cc]{}}
\put(51.00,215.00){\makebox(0,0)[cc]{$s$}}
\put(51.00,185.00){\makebox(0,0)[cc]{$n$}}
\put(36.00,150.00){\makebox(0,0)[cc]{$p$}}
\put(64.00,150.00){\makebox(0,0)[cc]{$q$}}
\put(26.00,120.00){\makebox(0,0)[cc]{$x$}}
\put(51.00,120.00){\makebox(0,0)[cc]{$t$}}
\put(76.00,120.00){\makebox(0,0)[cc]{$r$}}
\put(140.00,101.00){\makebox(0,0)[cc]{(b) MES $M_1(s,G)$}}
\put(135.00,120.00){\circle{10.00}}
\put(160.00,120.00){\circle{10.20}}
\put(145.00,150.00){\circle{10.00}}
\put(160.00,185.00){\circle{10.00}}
\put(160.00,215.00){\circle{10.00}}
\put(160.00,210.00){\vector(0,-1){20.00}}
\put(161.00,180.00){\vector(-2,-3){16.67}}
\put(145.00,145.00){\vector(-1,-2){10.00}}
\put(145.00,145.00){\vector(2,-3){13.33}}
\put(142.00,139.00){\line(1,0){7.00}}
\put(160.00,215.00){\makebox(0,0)[cc]{$s$}}
\put(160.00,185.00){\makebox(0,0)[cc]{$n$}}
\put(145.00,150.00){\makebox(0,0)[cc]{$p$}}
\put(135.00,120.00){\makebox(0,0)[cc]{$x$}}
\put(160.00,120.00){\makebox(0,0)[cc]{$t$}}
\put(250.00,110.00){\makebox(0,0)[cc]{(c) Sub-MES $\xi(p,M_1(s,G))$,}}
\put(230.00,155.00){\circle{10.00}}
\put(255.00,155.00){\circle{10.20}}
\put(240.00,185.00){\circle{10.00}}
\put(240.00,180.00){\vector(-1,-2){10.00}}
\put(240.00,180.00){\vector(2,-3){13.33}}
\put(237.00,174.00){\line(1,0){7.00}}
\put(240.00,185.00){\makebox(0,0)[cc]{$p$}}
\put(230.00,155.00){\makebox(0,0)[cc]{$x$}}
\put(255.00,155.00){\makebox(0,0)[cc]{$t$}}
\put(240.00,100.00){\makebox(0,0)[cc]{not an MES in $G$}}
\put(225.00,34.00){\circle{10.00}}
\put(225.00,34.00){\makebox(0,0)[cc]{$t$}}
\put(26.00,20.00){\circle{10.00}}
\put(51.00,20.00){\circle{10.20}}
\put(36.00,50.00){\circle{10.00}}
\put(51.00,85.00){\circle{10.00}}
\put(36.00,45.00){\vector(-1,-2){10.00}}
\put(36.00,45.00){\vector(2,-3){13.33}}
\put(33.00,39.00){\line(1,0){7.00}}
\bezier{480}(21.00,19.00)(-10.00,74.00)(46.00,83.00)
\put(43.00,82.00){\vector(3,1){3.00}}
\put(51.00,85.00){\makebox(0,0)[cc]{$n$}}
\put(36.00,50.00){\makebox(0,0)[cc]{$p$}}
\put(26.00,20.00){\makebox(0,0)[cc]{$x$}}
\put(51.00,20.00){\makebox(0,0)[cc]{$t$}}
\put(45.00,6.00){\makebox(0,0)[cc]{(d) MES $M_1(p,G)$}}
\put(95.00,95.00){\framebox(95.00,130.00)[cc]{}}
\put(190.00,95.00){\framebox(120.00,130.00)[cc]{}}
\put(190.00,0.00){\framebox(120.00,95.00)[cc]{}}
\put(0.00,0.00){\framebox(95.00,95.00)[cc]{}}
\put(225.00,85.00){\circle{10.00}}
\put(225.00,85.00){\makebox(0,0)[cc]{$n$}}
\put(253.00,60.00){\circle{10.00}}
\put(253.00,60.00){\makebox(0,0)[cc]{$q$}}
\put(225.00,80.00){\vector(3,-2){24.00}}
\put(253.00,55.00){\vector(-3,-2){25.00}}
\end{picture}
\end{figure}
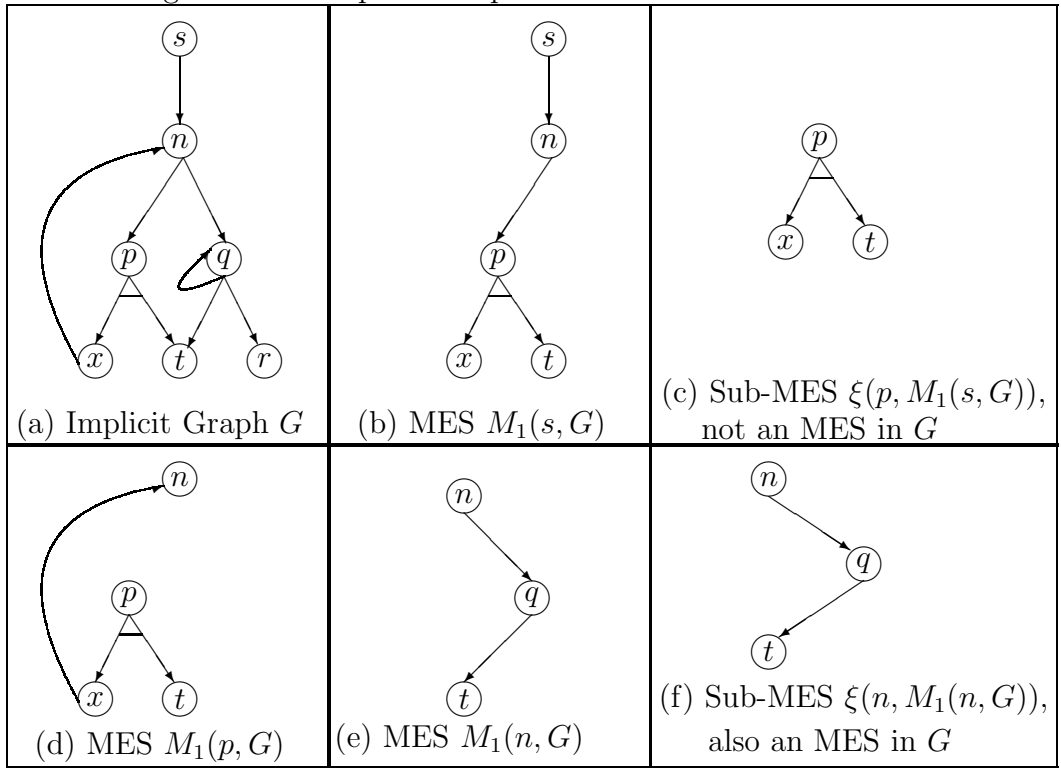

An AND/OR graph $G$ and some of its MESs and their sub-MESs are shown in Figure \ref{fig28}.

\begin{rem} \label{submes} 
\end{rem}
\begin{description}
\item[({\em i})] A sub-MES may or may not be an MES itself. This is shown in Figure \ref{fig28}. The sub-MES $\xi(p,M_1(s,G))$, coming from $M_1(s,G)$, is not an MES in $G$ (it would have been an MES if the arc ($x$,$n$) was present, as in $M_1(p,G)$). However, the sub-MES $\xi(n,M_1(n,G))$, coming from $M_1(n,G)$ is an MES in $G$. In general, if a sub-MES is rooted at the same node where the MES is rooted, the sub-MES will be an MES in $G$. Sub-MESs rooted at other nodes in the MES may or may not be MESs themselves in $G$.

\item[({\em ii})] There is exactly one sub-MES below every node in an MES.
\end{description}

\subsubsection{Classification of MESs in Implicit Graphs}

Depending upon the types of node (i.e. terminal leaf, nonterminal leaf, or other)
that the maximal paths of an MES terminate in, it is classified into
different types as follows. Note that, for any MES $M$, $Z_M$ represents the set of all nodes which have no children in $M$.

\begin{defi}
An MES $M(n,G)$ below a node $n \in G$ is said to be a

\begin{list}{\roman{defctr4})}{\usecounter{defctr4}}

\item {\bf type-I MES} (or, a {\bf solution graph}), if $Z_{M(n,G)} \subseteq T$; 
                           
\item {\bf type-II MES}, if 
                    \begin{list}{\alph{defctr9})}{\usecounter{defctr9}}
                    \item $Z_{M(n,G)} \subseteq Z_{G},$ and
                    \item $Z_{M(n,G)} \bigcap NT \not= \phi$;
                    \end{list}
\item {\bf type-III MES}, if $Z_{M(n,G)} \not \subseteq Z_{G}$. 

\end{list} 
\end{defi}

The different types of MESs capture the information
whether all the paths end in terminal leaf nodes, or some end in nonterminal leaf nodes, or some paths even get stuck in cycles on the way. The type numbers I, II and III
are a shorthand way of depicting the nature of an MES.

\begin{figure}
\centering
\caption{An implicit graph with some MESs} 
\label{fig30}
\unitlength=0.45mm
\special{em:linewidth 0.4pt}
\linethickness{0.4pt}
\begin{picture}(360.00,230.00)
\put(46.00,205.00){\circle{8.00}}
\put(26.00,170.00){\circle{8.25}}
\put(66.00,170.00){\circle{8.00}}
\put(26.00,135.00){\circle{8.00}}
\put(51.00,135.00){\circle{8.00}}
\put(81.00,135.00){\circle{8.25}}
\put(46.00,201.00){\vector(-2,-3){18.00}}
\put(46.00,201.00){\vector(2,-3){18.00}}
\put(26.00,166.00){\vector(0,-1){27.00}}
\put(26.00,166.00){\vector(3,-4){21.00}}
\put(66.00,166.00){\vector(-1,-2){13.67}}
\put(66.00,166.00){\vector(1,-2){13.67}}
\bezier{224}(66.00,166.00)(41.00,153.00)(62.00,172.00)
\bezier{484}(22.00,135.00)(-14.00,185.00)(42.00,205.00)
\put(39.00,204.00){\vector(3,1){3.00}}
\put(26.00,160.00){\line(1,0){4.00}}
\put(46.00,205.00){\makebox(0,0)[cc]{$n$}}
\put(66.00,170.00){\makebox(0,0)[cc]{$q$}}
\put(26.00,170.00){\makebox(0,0)[cc]{$p$}}
\put(26.00,135.00){\makebox(0,0)[cc]{$x$}}
\put(51.00,135.00){\makebox(0,0)[cc]{$t$}}
\put(81.00,135.00){\makebox(0,0)[cc]{$r$}}
\put(45.00,121.00){\makebox(0,0)[cc]{(a) Implicit Graph $G$}}
\put(0.00,115.00){\framebox(90.00,115.00)[cc]{}}
\put(135.00,107.00){\circle{8.00}}
\put(115.00,72.00){\circle{8.25}}
\put(155.00,72.00){\circle{8.00}}
\put(115.00,37.00){\circle{8.00}}
\put(140.00,37.00){\circle{8.00}}
\put(135.00,103.00){\vector(2,-3){18.00}}
\put(115.00,68.00){\vector(0,-1){27.00}}
\put(115.00,68.00){\vector(3,-4){21.00}}
\put(155.00,68.00){\vector(-1,-2){13.67}}
\bezier{484}(111.00,37.00)(75.00,87.00)(131.00,107.00)
\put(128.00,106.00){\vector(3,1){3.00}}
\put(115.00,62.00){\line(1,0){4.00}}
\put(135.00,107.00){\makebox(0,0)[cc]{$n$}}
\put(155.00,72.00){\makebox(0,0)[cc]{$q$}}
\put(115.00,72.00){\makebox(0,0)[cc]{$p$}}
\put(115.00,37.00){\makebox(0,0)[cc]{$x$}}
\put(140.00,37.00){\makebox(0,0)[cc]{$t$}}
\put(127.00,22.00){\makebox(0,0)[cc]{(f) type-I MES}}
\put(90.00,0.00){\framebox(90.00,115.00)[cc]{}}
\put(127.00,8.00){\makebox(0,0)[cc]{$M_1(p,G)$ below $p$}}
\put(225.00,222.00){\circle{8.00}}
\put(245.00,187.00){\circle{8.00}}
\put(260.00,152.00){\circle{8.25}}
\put(225.00,218.00){\vector(2,-3){18.00}}
\put(245.00,183.00){\vector(1,-2){13.67}}
\put(225.00,222.00){\makebox(0,0)[cc]{$n$}}
\put(245.00,187.00){\makebox(0,0)[cc]{$q$}}
\put(260.00,152.00){\makebox(0,0)[cc]{$r$}}
\put(217.00,137.00){\makebox(0,0)[cc]{(c) type-II MES}}
\put(180.00,115.00){\framebox(90.00,115.00)[cc]{}}
\put(217.00,123.00){\makebox(0,0)[cc]{$M_2(n,G)$ below $n$}}
\put(0.00,0.00){\framebox(90.00,115.00)[cc]{}}
\put(135.00,222.00){\circle{8.00}}
\put(155.00,187.00){\circle{8.00}}
\put(140.00,152.00){\circle{8.00}}
\put(135.00,218.00){\vector(2,-3){18.00}}
\put(155.00,183.00){\vector(-1,-2){13.67}}
\put(135.00,222.00){\makebox(0,0)[cc]{$n$}}
\put(155.00,187.00){\makebox(0,0)[cc]{$q$}}
\put(140.00,152.00){\makebox(0,0)[cc]{$t$}}
\put(127.00,137.00){\makebox(0,0)[cc]{(b) type-I MES}}
\put(90.00,115.00){\framebox(90.00,115.00)[cc]{}}
\put(127.00,122.00){\makebox(0,0)[cc]{$M_1(n,G)$ below $n$}}
\put(180.00,0.00){\framebox(90.00,115.00)[cc]{}}
\put(46.00,224.00){\circle{8.00}}
\put(46.00,220.00){\vector(0,-1){11.00}}
\put(46.00,224.00){\makebox(0,0)[cc]{$s$}}
\put(60.00,170.00){\vector(1,1){2.00}}
\put(305.00,222.00){\circle{8.00}}
\put(285.00,187.00){\circle{8.25}}
\put(285.00,152.00){\circle{8.00}}
\put(310.00,152.00){\circle{8.00}}
\put(305.00,218.00){\vector(-2,-3){18.00}}
\put(285.00,183.00){\vector(0,-1){27.00}}
\put(285.00,183.00){\vector(3,-4){21.00}}
\put(285.00,177.00){\line(1,0){4.00}}
\put(305.00,222.00){\makebox(0,0)[cc]{$n$}}
\put(285.00,187.00){\makebox(0,0)[cc]{$p$}}
\put(285.00,152.00){\makebox(0,0)[cc]{$x$}}
\put(310.00,152.00){\makebox(0,0)[cc]{$t$}}
\put(308.00,137.00){\makebox(0,0)[cc]{(d) type-III MES}}
\put(308.00,125.00){\makebox(0,0)[cc]{$M_3(n,G)$ below $n$}}
\put(270.00,115.00){\framebox(90.00,115.00)[cc]{}}
\put(224.00,72.00){\circle{8.00}}
\put(224.00,72.00){\makebox(0,0)[cc]{$q$}}
\put(224.00,22.00){\makebox(0,0)[cc]{(g) type-III MES}}
\put(224.00,11.00){\makebox(0,0)[cc]{$M_1(q,G)$ below $q$}}
\put(315.00,100.00){\circle{8.00}}
\put(288.00,60.00){\circle{8.00}}
\put(315.00,96.00){\vector(-3,-4){24.67}}
\put(315.00,23.00){\makebox(0,0)[cc]{(h) A subgraph which}}
\put(315.00,12.00){\makebox(0,0)[cc]{is not an MES}}
\put(315.00,100.00){\makebox(0,0)[cc]{$n$}}
\put(288.00,60.00){\makebox(0,0)[cc]{$p$}}
\put(270.00,0.00){\framebox(90.00,115.00)[cc]{}}
\put(46.00,100.00){\circle{8.00}}
\put(66.00,65.00){\circle{8.00}}
\put(46.00,96.00){\vector(2,-3){18.00}}
\put(46.00,100.00){\makebox(0,0)[cc]{$n$}}
\put(66.00,65.00){\makebox(0,0)[cc]{$q$}}
\put(45.00,22.00){\makebox(0,0)[cc]{(e) type-III MES }}
\put(45.00,10.00){\makebox(0,0)[cc]{$M_4(n,G)$ below $n$}}
\end{picture}
\end{figure}
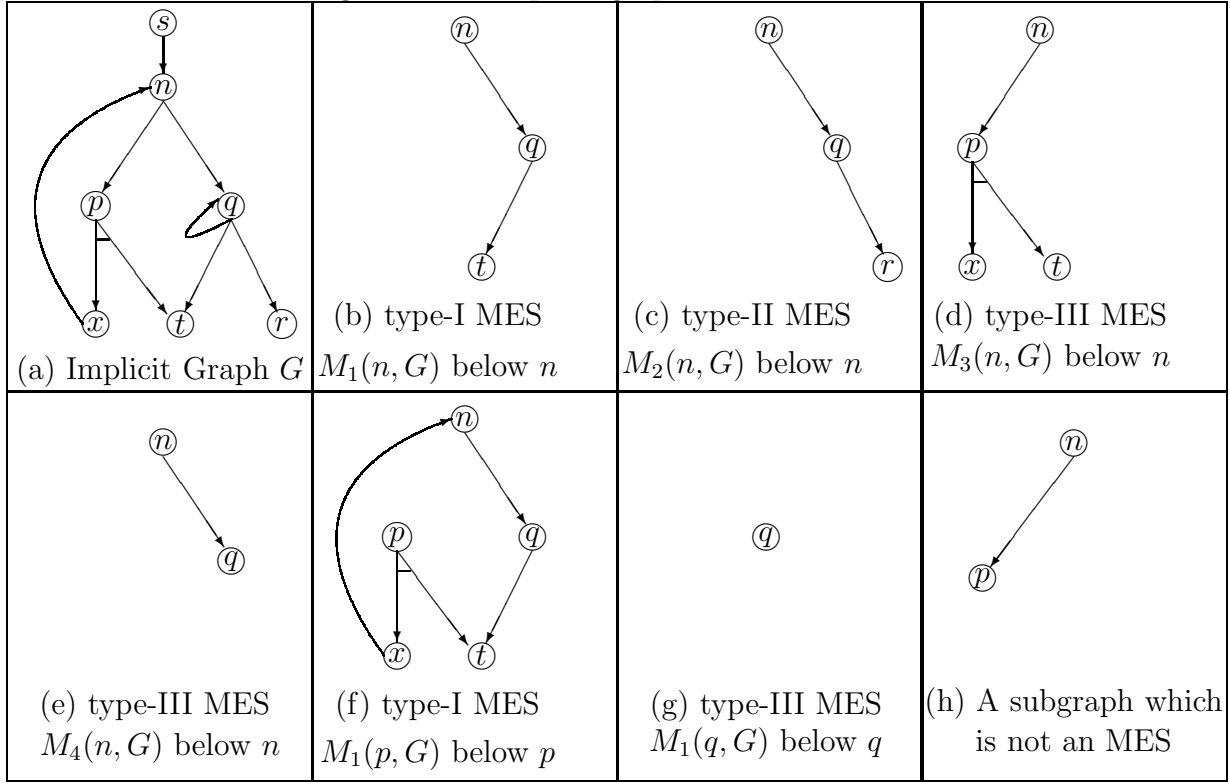

For example, for the implicit graph $G$ given in Figure \ref{fig30}(a), four MESs below $n$ are shown in Figure \ref{fig30}(b)-\ref{fig30}(e). These MESs are of types I, II, III and III respectively. In Figure \ref{fig30}(f), $M_1(p,G)$ is a type-I MES below $p$ that passes through $n$. It is interesting to note that $n$ is a predecessor of $p$ in $G$. In Figure \ref{fig30}(g), $M_1(q,G)$ is a type-III MES due to the self-loop at $q$. In Figure \ref{fig30}(h), the subgraph is not an MES, nor is it a sub-MES of $M_3(n,G)$.

\subsubsection{Classification of Nodes in Implicit Graphs}
              
A node in an implicit graph is classified into different types,
depending on the type of MESs below it.

\begin{defi}    \label{def60}
A node $n$ in $G$ is said to be of
\newcounter{rom_ctr0}
\begin{list}{\roman{rom_ctr0})}{\usecounter{rom_ctr0}}
  \item {\bf type-I}, if there is a type-I MES
  below $n$ in $G$.
  \item {\bf type-II}, if there is no MES of type-I, 
         but at least one MES of type-II below $n$ in $G$. 
  \item {\bf type-III}, if there is no MES of type-I or type-II below $n$ in $G$.
\end{list}

\end{defi}

\begin{rem}
\end{rem}
\begin{description}
\item[({\em i})] We classify the nodes as above to highlight the information
contained in them about their solvability, or the reasons for not being
solvable. A type-I node is a solvable one (i.e. it contains a solution graph
below it), a type-II node is not solvable as each of its MESs contains one or more
nonterminal leaves which are known to be unsolvable, and a type-III node is
not solvable as each of its MESs gets stuck at some internal node(s) of $G$ due to cycles. Thus there is a distinct difference between a type-II
node and a type-III node in $G$. 
\item[({\em ii})] If the graph $G$ is acyclic, there cannot be any type-III nodes in $G$. However, even if the graph is cyclic, there may not be any type-III nodes in it, as is evident from Figures \ref{fig50}(a) and \ref{fig50}(b).
\end{description}

For example, in Figures \ref{fig50}(a) to \ref{fig50}(c) we show three implicit graphs $G_1$, $G_2$ and $G_3$ and label each node with its type (I, II or III). Observe that, changes from $G_1$ to $G_2$ happen because $x$ has been made an AND node, and changes from $G_2$ to $G_3$ happen because $p$ has also been made an AND node.

\unitlength=0.45mm
\special{em:linewidth 0.4pt}
\linethickness{0.4pt}
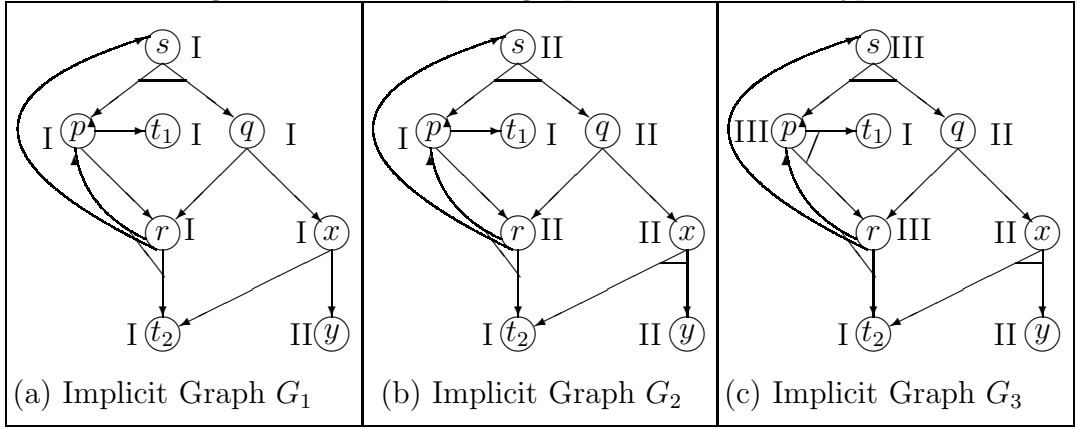
\begin{figure}
\centering
\caption{Three implicit graphs and their node types} 
\label{fig50}
\begin{picture}(315.00,125.00)
\put(46.00,112.00){\circle{10.00}}
\put(21.00,87.00){\circle{10.20}}
\put(71.00,87.00){\circle{10.00}}
\put(46.00,87.00){\circle{10.00}}
\put(46.00,57.00){\circle{10.00}}
\put(96.00,57.00){\circle{10.00}}
\put(46.00,27.00){\circle{10.00}}
\put(96.00,27.00){\circle{10.20}}
\put(46.00,107.00){\vector(-4,-3){21.00}}
\put(25.00,91.33){\vector(0,0){0.00}}
\put(46.00,107.00){\vector(4,-3){21.00}}
\put(26.00,87.00){\vector(1,0){15.00}}
\put(71.00,82.00){\vector(-1,-1){21.00}}
\put(71.00,82.00){\vector(1,-1){22.00}}
\put(46.00,52.00){\vector(0,-1){20.00}}
\put(96.00,52.00){\vector(0,-1){20.00}}
\put(96.00,52.00){\vector(-2,-1){45.00}}
\put(39.00,102.00){\line(1,0){14.00}}
\put(46.00,112.00){\makebox(0,0)[cc]{$s$}}
\put(56.00,112.00){\makebox(0,0)[cc]{I}}
\put(21.00,87.00){\makebox(0,0)[cc]{$p$}}
\put(46.00,87.00){\makebox(0,0)[cc]{$t_1$}}
\put(12.00,85.00){\makebox(0,0)[cc]{I}}
\put(56.00,86.00){\makebox(0,0)[cc]{I}}
\put(71.00,87.00){\makebox(0,0)[cc]{$q$}}
\put(84.00,86.00){\makebox(0,0)[cc]{I}}
\put(46.00,57.00){\makebox(0,0)[cc]{$r$}}
\put(54.00,58.00){\makebox(0,0)[cc]{I}}
\put(96.00,57.00){\makebox(0,0)[cc]{$x$}}
\put(46.00,27.00){\makebox(0,0)[cc]{$t_2$}}
\put(96.00,27.00){\makebox(0,0)[cc]{$y$}}
\put(87.00,57.00){\makebox(0,0)[cc]{I}}
\put(37.00,27.00){\makebox(0,0)[cc]{I}}
\put(87.00,27.00){\makebox(0,0)[cc]{II}}
\put(46.00,9.00){\makebox(0,0)[cc]{(a) Implicit Graph $G_1$}}
\bezier{688}(44.00,52.00)(-37.00,91.00)(42.00,115.00)
\bezier{156}(41.00,55.00)(22.00,64.00)(20.00,82.00)
\put(22.00,82.00){\vector(1,-1){21.00}}
\put(20.00,80.00){\vector(0,1){0.00}}
\put(39.00,114.00){\vector(2,1){2.00}}
\put(0.00,0.00){\framebox(105.00,125.00)[cc]{}}
\put(36.00,58.00){\line(3,-4){10.67}}
\put(151.00,112.00){\circle{10.00}}
\put(126.00,87.00){\circle{10.20}}
\put(176.00,87.00){\circle{10.00}}
\put(151.00,87.00){\circle{10.00}}
\put(151.00,57.00){\circle{10.00}}
\put(201.00,57.00){\circle{10.00}}
\put(151.00,27.00){\circle{10.00}}
\put(201.00,27.00){\circle{10.20}}
\put(151.00,107.00){\vector(-4,-3){21.00}}
\put(130.00,91.33){\vector(0,0){0.00}}
\put(151.00,107.00){\vector(4,-3){21.00}}
\put(131.00,87.00){\vector(1,0){15.00}}
\put(176.00,82.00){\vector(-1,-1){21.00}}
\put(176.00,82.00){\vector(1,-1){22.00}}
\put(151.00,52.00){\vector(0,-1){20.00}}
\put(201.00,52.00){\vector(0,-1){20.00}}
\put(201.00,52.00){\vector(-2,-1){45.00}}
\put(144.00,102.00){\line(1,0){14.00}}
\put(151.00,112.00){\makebox(0,0)[cc]{$s$}}
\put(161.00,112.00){\makebox(0,0)[cc]{II}}
\put(126.00,87.00){\makebox(0,0)[cc]{$p$}}
\put(151.00,87.00){\makebox(0,0)[cc]{$t_1$}}
\put(117.00,85.00){\makebox(0,0)[cc]{I}}
\put(161.00,86.00){\makebox(0,0)[cc]{I}}
\put(176.00,87.00){\makebox(0,0)[cc]{$q$}}
\put(189.00,86.00){\makebox(0,0)[cc]{II}}
\put(151.00,57.00){\makebox(0,0)[cc]{$r$}}
\put(161.00,58.00){\makebox(0,0)[cc]{II}}
\put(201.00,57.00){\makebox(0,0)[cc]{$x$}}
\put(151.00,27.00){\makebox(0,0)[cc]{$t_2$}}
\put(201.00,27.00){\makebox(0,0)[cc]{$y$}}
\put(190.00,57.00){\makebox(0,0)[cc]{II}}
\put(142.00,27.00){\makebox(0,0)[cc]{I}}
\put(190.00,27.00){\makebox(0,0)[cc]{II}}
\put(155.00,9.00){\makebox(0,0)[cc]{(b) Implicit Graph $G_2$}}
\bezier{688}(149.00,52.00)(68.00,91.00)(147.00,115.00)
\bezier{156}(146.00,55.00)(127.00,64.00)(125.00,82.00)
\put(127.00,82.00){\vector(1,-1){21.00}}
\put(125.00,80.00){\vector(0,1){0.00}}
\put(144.00,114.00){\vector(2,1){2.00}}
\put(105.00,0.00){\framebox(105.00,125.00)[cc]{}}
\put(141.00,58.00){\line(3,-4){10.67}}
\put(193.00,48.00){\line(1,0){8.00}}
\put(256.00,112.00){\circle{10.00}}
\put(231.00,87.00){\circle{10.20}}
\put(281.00,87.00){\circle{10.00}}
\put(256.00,87.00){\circle{10.00}}
\put(256.00,57.00){\circle{10.00}}
\put(306.00,57.00){\circle{10.00}}
\put(256.00,27.00){\circle{10.00}}
\put(306.00,27.00){\circle{10.20}}
\put(256.00,107.00){\vector(-4,-3){21.00}}
\put(235.00,91.33){\vector(0,0){0.00}}
\put(256.00,107.00){\vector(4,-3){21.00}}
\put(236.00,87.00){\vector(1,0){15.00}}
\put(281.00,82.00){\vector(-1,-1){21.00}}
\put(281.00,82.00){\vector(1,-1){22.00}}
\put(256.00,52.00){\vector(0,-1){20.00}}
\put(306.00,52.00){\vector(0,-1){20.00}}
\put(306.00,52.00){\vector(-2,-1){45.00}}
\put(249.00,102.00){\line(1,0){14.00}}
\put(256.00,112.00){\makebox(0,0)[cc]{$s$}}
\put(266.00,112.00){\makebox(0,0)[cc]{III}}
\put(231.00,87.00){\makebox(0,0)[cc]{$p$}}
\put(256.00,87.00){\makebox(0,0)[cc]{$t_1$}}
\put(220.00,87.00){\makebox(0,0)[cc]{III}}
\put(266.00,86.00){\makebox(0,0)[cc]{I}}
\put(281.00,87.00){\makebox(0,0)[cc]{$q$}}
\put(294.00,86.00){\makebox(0,0)[cc]{II}}
\put(256.00,57.00){\makebox(0,0)[cc]{$r$}}
\put(268.00,58.00){\makebox(0,0)[cc]{III}}
\put(306.00,57.00){\makebox(0,0)[cc]{$x$}}
\put(256.00,27.00){\makebox(0,0)[cc]{$t_2$}}
\put(306.00,27.00){\makebox(0,0)[cc]{$y$}}
\put(295.00,57.00){\makebox(0,0)[cc]{II}}
\put(247.00,27.00){\makebox(0,0)[cc]{I}}
\put(295.00,27.00){\makebox(0,0)[cc]{II}}
\put(256.00,9.00){\makebox(0,0)[cc]{(c) Implicit Graph $G_3$}}
\bezier{688}(254.00,52.00)(173.00,91.00)(252.00,115.00)
\bezier{156}(251.00,55.00)(232.00,64.00)(230.00,82.00)
\put(232.00,82.00){\vector(1,-1){21.00}}
\put(230.00,80.00){\vector(0,1){0.00}}
\put(249.00,114.00){\vector(2,1){2.00}}
\put(210.00,0.00){\framebox(105.00,125.00)[cc]{}}
\put(246.00,58.00){\line(3,-4){10.67}}
\put(298.00,48.00){\line(1,0){8.00}}
\put(240.00,87.00){\line(-2,-5){3.67}}
\end{picture}
\end{figure}

\subsubsection {Properties of MESs}

Now we discuss the inter-relationships between the
different types of nodes and different types of MESs. These 
results are the important building blocks in this proposed new framework for cyclic AND/OR graphs.

\begin{lem} \label{lem01}

For any MES $M(n,G)$:

\newcounter{rom_ctr30}
\begin{list}{\roman{rom_ctr30})}{\usecounter{rom_ctr30}}

\item  If $M(n,G)$ is of type-I, then for every $p \in M(n,G)$ the sub-MES $\xi(p,M(n,G))$ is also an MES, and it is of type-I.

\item  If $M(n,G)$ is of type-II, then for every $p \in M(n,G)$ the sub-MES $\xi(p,M(n,G))$ is also an MES, and it is of type-I  or type-II.
\end{list}
\end{lem}

{\bf Proof.} Clear. \qed

\begin{lem} \label{lem02}

Let $M$ be an MES below a node $n \in G$ and $p$ be any node in $M$.
Now,
\newcounter{rom_ctr31}
\begin{list}{\roman{rom_ctr31})}{\usecounter{rom_ctr31}}

\item If $M$ is of type-I then $p$ must be of type-I.
\item If $M$ is of type-II then $p$ must be of either type-I or type-II, and at least one such $p$ is of type-II. 
\item If $M$ is of type-III then $p$ may be of type-I or type-II or type-III.
\end{list}
\end{lem}

{\bf Proof.} Clear.  \qed

As an example of Lemma \ref{lem02}(\textit{iii}), we refer to Figure \ref{fig51}. Even though the MES $M_1(s,G_1)$ in Figure \ref{fig51}(b) is of type-III, all of its nodes $s$ and $n$ are of type-I in $G_1$ in Figure \ref{fig51}(a). Again, MES $M_1(s,G_2)$ in Figure \ref{fig51}(d) is of type-III, but all of its nodes $s$ and $n$ are of type-II in $G_2$ in Figure \ref{fig51}(c). It should be noted that the type of a node is defined globally based on all the MESs below it, but the type of an MES $M$ is based on the nodes of $Z_M$ - whether they are terminal, nonterminal of internal nodes of $G$.

\unitlength=0.45mm
\special{em:linewidth 0.4pt}
\linethickness{0.4pt}
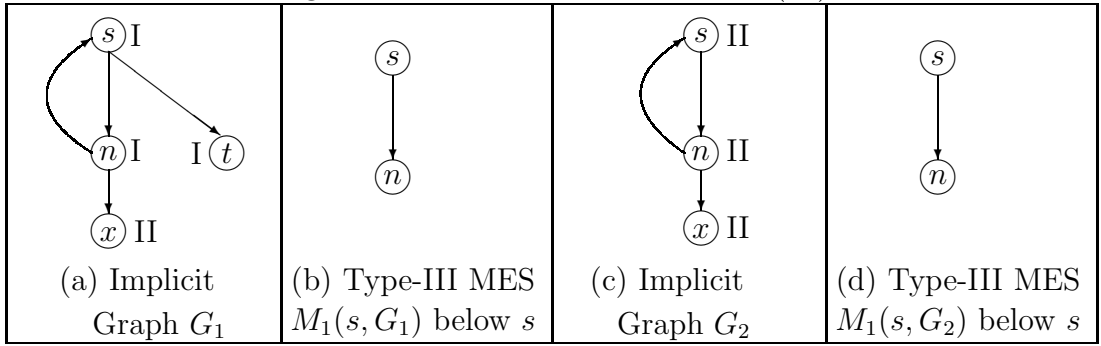
\begin{figure}
\centering
\caption{Illustration of Lemma \ref{lem02}(\textit{iii})} 
\label{fig51}
\begin{picture}(322.00,100.00)
\put(30.00,91.00){\circle{10.00}}
\put(30.00,56.00){\circle{10.20}}
\put(30.00,33.00){\circle{10.00}}
\put(65.00,56.00){\circle{10.00}}
\put(30.00,86.00){\vector(0,-1){25.00}}
\put(30.00,86.00){\vector(4,-3){33.00}}
\put(30.00,91.00){\makebox(0,0)[cc]{$s$}}
\put(30.00,56.00){\makebox(0,0)[cc]{$n$}}
\put(65.00,56.00){\makebox(0,0)[cc]{$t$}}
\put(30.00,32.00){\makebox(0,0)[cc]{$x$}}
\put(114.00,84.00){\circle{10.00}}
\put(114.00,49.00){\circle{10.20}}
\put(114.00,79.00){\vector(0,-1){25.00}}
\put(114.00,84.00){\makebox(0,0)[cc]{$s$}}
\put(114.00,49.00){\makebox(0,0)[cc]{$n$}}
\put(81.00,0.00){\framebox(80.00,100.00)[cc]{}}
\put(40.00,18.00){\makebox(0,0)[cc]{(a) Implicit }}
\put(45.00,5.00){\makebox(0,0)[cc]{Graph $G_1$}}
\put(120.00,18.00){\makebox(0,0)[cc]{(b) Type-III MES}}
\bezier{256}(25.00,56.00)(-2.00,73.00)(25.00,90.00)
\put(22.00,88.00){\vector(3,2){3.00}}
\put(120.00,6.00){\makebox(0,0)[cc]{$M_1(s,G_1)$ below $s$}}
\put(38.00,91.00){\makebox(0,0)[cc]{I}}
\put(38.00,57.00){\makebox(0,0)[cc]{I}}
\put(41.00,33.00){\makebox(0,0)[cc]{II}}
\put(56.00,56.00){\makebox(0,0)[cc]{I}}
\put(0.00,0.00){\framebox(81.00,100.00)[cc]{}}
\put(205.00,91.00){\circle{10.00}}
\put(205.00,56.00){\circle{10.20}}
\put(205.00,34.00){\circle{10.00}}
\put(205.00,86.00){\vector(0,-1){25.00}}
\put(205.00,91.00){\makebox(0,0)[cc]{$s$}}
\put(205.00,56.00){\makebox(0,0)[cc]{$n$}}
\put(205.00,33.00){\makebox(0,0)[cc]{$x$}}
\put(195.00,18.00){\makebox(0,0)[cc]{(c) Implicit}}
\put(200.00,5.00){\makebox(0,0)[cc]{Graph $G_2$}}
\put(197.00,88.00){\vector(3,2){3.00}}
\put(216.00,91.00){\makebox(0,0)[cc]{II}}
\put(216.00,57.00){\makebox(0,0)[cc]{II}}
\put(216.00,34.00){\makebox(0,0)[cc]{II}}
\put(161.00,0.00){\framebox(81.00,100.00)[cc]{}}
\bezier{248}(200.00,56.00)(174.00,74.00)(200.00,90.00)
\put(275.00,84.00){\circle{10.00}}
\put(275.00,49.00){\circle{10.20}}
\put(275.00,79.00){\vector(0,-1){25.00}}
\put(275.00,84.00){\makebox(0,0)[cc]{$s$}}
\put(275.00,49.00){\makebox(0,0)[cc]{$n$}}
\put(242.00,0.00){\framebox(80.00,100.00)[cc]{}}
\put(281.00,18.00){\makebox(0,0)[cc]{(d) Type-III MES}}
\put(281.00,6.00){\makebox(0,0)[cc]{$M_1(s,G_2)$ below $s$}}
\put(30.00,51.00){\vector(0,-1){13.00}}
\put(205.00,51.00){\vector(0,-1){12.00}}
\end{picture}
\end{figure}

\begin{th} \label{thm20} {\bf Sub-problem Composition Theorem for Implicit Graphs.} Let $G$ be an implicit 
AND/OR graph and n be any internal node in $G$. Now, \\
\newcounter{rom_ctr7}
\begin{list}{\roman{rom_ctr7})}{\usecounter{rom_ctr7}}
   \item If $n$ is an OR node, then:
   
    (a) $n$ is of type-I iff at least one child of $n$ is of type-I; 
   
   (b) $n$ is of type-II iff no child of $n$ is of type-I but at least one child is of type-II 
   
   (c) $n$ is of type-III iff all children of $n$ are of type-III.
   \item If $n$ is an AND node, then:
   
   (a) $n$ is of type-I iff every child of $n$ is of type-I; 
   
   (b) $n$ is of type-II iff  at least one child of $n$ is of type-II and no child is of type-III; 
   
   (c) $n$ is of type-III iff at least one child of $n$ is of type-III.
\end{list}                                    
\end{th}

Proof. Similar to the proof of Theorem \ref{thm25} on Explicit Graphs, presented later. \qed\\

\subsubsection{Costs of MESs in Implicit Graphs}

Finally we come to the notion of costs. For AND/OR graphs with cycles, we define the cost with respect to an MES.

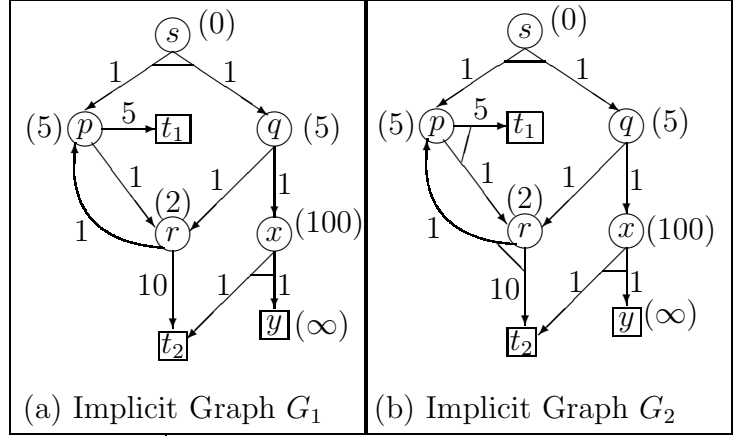
\begin{figure}
\centering
\caption{Two implicit graphs} 
\label{fig70}
\begin{picture}(213.00,128.00)
\put(46.00,-0.33){\line(0,1){0.33}}
\put(48.00,118.00){\circle{10.00}}
\put(22.00,90.00){\circle{10.00}}
\put(78.00,90.00){\circle{10.00}}
\put(48.00,59.00){\circle{10.00}}
\put(78.00,59.00){\circle{10.00}}
\put(78.00,54.00){\vector(0,-1){17.00}}
\put(78.00,54.00){\vector(-1,-1){26.00}}
\put(48.00,54.00){\vector(0,-1){23.00}}
\put(24.00,86.00){\vector(3,-4){18.67}}
\put(78.00,85.00){\vector(0,-1){21.00}}
\put(78.00,85.00){\vector(-1,-1){25.00}}
\put(27.00,90.00){\vector(1,0){16.00}}
\put(48.00,113.00){\vector(-3,-2){26.00}}
\put(48.00,113.00){\vector(3,-2){28.00}}
\bezier{236}(45.00,55.00)(15.00,57.00)(19.00,86.00)
\put(19.00,80.00){\vector(0,1){6.00}}
\put(81.00,44.00){\makebox(0,0)[cc]{1}}
\put(63.00,44.00){\makebox(0,0)[cc]{1}}
\put(42.00,44.00){\makebox(0,0)[cc]{10}}
\put(48.00,59.00){\makebox(0,0)[cc]{$r$}}
\put(78.00,59.00){\makebox(0,0)[cc]{$x$}}
\put(81.00,74.00){\makebox(0,0)[cc]{1}}
\put(94.00,62.00){\makebox(0,0)[cc]{(100)}}
\put(78.00,90.00){\makebox(0,0)[cc]{$q$}}
\put(61.00,75.00){\makebox(0,0)[cc]{1}}
\put(48.00,68.00){\makebox(0,0)[cc]{(2)}}
\put(37.00,76.00){\makebox(0,0)[cc]{1}}
\put(21.00,60.00){\makebox(0,0)[cc]{1}}
\put(22.00,90.00){\makebox(0,0)[cc]{$p$}}
\put(10.00,90.00){\makebox(0,0)[cc]{(5)}}
\put(35.00,95.00){\makebox(0,0)[cc]{5}}
\put(48.00,90.00){\makebox(0,0)[cc]{$t_{1}$}}
\put(31.00,107.00){\makebox(0,0)[cc]{1}}
\put(65.00,107.00){\makebox(0,0)[cc]{1}}
\put(48.00,118.00){\makebox(0,0)[cc]{$s$}}
\put(61.00,121.00){\makebox(0,0)[cc]{(0)}}
\put(92.00,90.00){\makebox(0,0)[cc]{(5)}}
\put(92.00,32.00){\makebox(0,0)[cc]{($\infty$)}}
\put(152.00,119.00){\circle{10.00}}
\put(126.00,91.00){\circle{10.00}}
\put(182.00,91.00){\circle{10.00}}
\put(152.00,60.00){\circle{10.00}}
\put(182.00,60.00){\circle{10.00}}
\put(182.00,55.00){\vector(0,-1){17.00}}
\put(182.00,55.00){\vector(-1,-1){26.00}}
\put(152.00,55.00){\vector(0,-1){23.00}}
\put(128.00,87.00){\vector(3,-4){18.67}}
\put(182.00,86.00){\vector(0,-1){21.00}}
\put(182.00,86.00){\vector(-1,-1){25.00}}
\put(131.00,91.00){\vector(1,0){16.00}}
\put(152.00,114.00){\vector(-3,-2){26.00}}
\put(152.00,114.00){\vector(3,-2){28.00}}
\bezier{236}(149.00,56.00)(119.00,58.00)(123.00,87.00)
\put(123.00,81.00){\vector(0,1){6.00}}
\put(185.00,45.00){\makebox(0,0)[cc]{1}}
\put(167.00,45.00){\makebox(0,0)[cc]{1}}
\put(146.00,43.00){\makebox(0,0)[cc]{10}}
\put(152.00,60.00){\makebox(0,0)[cc]{$r$}}
\put(182.00,60.00){\makebox(0,0)[cc]{$x$}}
\put(185.00,75.00){\makebox(0,0)[cc]{1}}
\put(198.00,60.00){\makebox(0,0)[cc]{(100)}}
\put(182.00,91.00){\makebox(0,0)[cc]{$q$}}
\put(165.00,76.00){\makebox(0,0)[cc]{1}}
\put(152.00,69.00){\makebox(0,0)[cc]{(2)}}
\put(141.00,77.00){\makebox(0,0)[cc]{1}}
\put(125.00,61.00){\makebox(0,0)[cc]{1}}
\put(126.00,91.00){\makebox(0,0)[cc]{$p$}}
\put(114.00,91.00){\makebox(0,0)[cc]{(5)}}
\put(139.00,96.00){\makebox(0,0)[cc]{5}}
\put(152.00,91.00){\makebox(0,0)[cc]{$t_{1}$}}
\put(135.00,108.00){\makebox(0,0)[cc]{1}}
\put(169.00,108.00){\makebox(0,0)[cc]{1}}
\put(152.00,119.00){\makebox(0,0)[cc]{$s$}}
\put(165.00,122.00){\makebox(0,0)[cc]{(0)}}
\put(195.00,35.00){\makebox(0,0)[cc]{($\infty$)}}
\put(48.00,7.00){\makebox(0,0)[cc]{(a) Implicit  Graph $G_{1}$}}
\put(152.00,7.00){\makebox(0,0)[cc]{(b) Implicit Graph $G_{2}$}}
\put(42.00,109.00){\line(1,0){12.00}}
\put(71.00,47.00){\line(1,0){7.00}}
\put(146.00,110.00){\line(1,0){12.00}}
\put(136.00,91.00){\line(-1,-4){2.67}}
\put(175.00,48.00){\line(1,0){7.00}}
\put(195.00,92.00){\makebox(0,0)[cc]{(5)}}
\put(44.00,22.00){\framebox(8.00,8.00)[cc]{$t_{2}$}}
\put(74.00,28.00){\framebox(8.00,8.00)[cc]{$y$}}
\put(147.00,23.00){\framebox(8.00,8.00)[cc]{$t_{2}$}}
\put(178.00,29.00){\framebox(8.00,8.00)[cc]{$y$}}
\put(43.00,86.00){\framebox(10.00,8.00)[cc]{}}
\put(147.00,86.00){\framebox(10.00,9.00)[cc]{}}
\put(152.00,48.00){\line(-1,1){9.00}}
\put(0.00,0.00){\framebox(105.00,128.00)[cc]{}}
\put(105.00,0.00){\framebox(108.00,128.00)[cc]{}}
\end{picture}
\end{figure}

The definitions in this section will be illustrated using the graphs shown in Figures \ref{fig70} and \ref{fig80}. These graphs show the arc-costs beside each arc and the heuristic values in parenthesis beside each node. The heuristic values of terminal leaf nodes are assumed to be zero. These heuristic values will be useful in later discussions on explicit graphs, where the same figures will be referred.

\special{em:linewidth 0.4pt}
\linethickness{0.4pt}

\begin{defi} \label{beta} Given any implicit graph $G$, for any MES $M(n,G)$, we denote the cost of a node $u$ in $M(n,G)$ by
$\beta(u,M(n,G))$. We define this cost function $\beta(u,M(n,G))$ as follows:

\begin{tabbing}
aaaaaaaaaaaa\=aaa\= \kill
$\beta(u,M(n,G))$ \\ 
= $0$  if $u \in  T$;  \\
=  $\infty$  if $u \in  NT$; \\
= $c(u,w) + \beta(w,M(n,G))$, if $u$ is an OR node and $w$ is the child of $u$ in $M(n,G)$  \\
= $\sum_{w \in \Gamma(u,M(n,G))} \{c(u,w) + \beta(w,M(n,G)) \}$ if $u$ is an AND node.\\

\\
$\beta(u,M(n,G))$ is undefined if $u \in Z_M \setminus \{T \bigcup NT \}$.
\end{tabbing}

\end{defi}

In Figure \ref{fig80}(a), 
$\beta(t_{1},M_{1}(s,G_{1})) = 0$, $\beta(p,M_{1}(s,G_{1})) = 5$, $\beta(r,M_{1}(s,G_{1})) = 6$, $\beta(q,M_{1}(s,G_{1})) = 7$ and
$\beta(s,M_{1}(s,G_{1})) = 14$. Similarly $\beta(t_2,M_{2}(q,G_{2})) = 0$, $\beta(y,M_{2}(q,G_{2})) = \infty$ (since $y \in NT$), $\beta(x,M_{2}(q,G_{2})) = \infty$ and $\beta(q,M_{2}(q,G_{2})) = \infty$. In $M_1(s,G_2)$, $\beta(t_1,M_{1}(s,G_{2})) = 0$. But since $\beta(r,M_{1}(s,G_{2}))$ is undefined, the $\beta$-values of $p$, $q$ and $s$ are also undefined.

\linethickness{0.4pt}
\unitlength=0.45mm
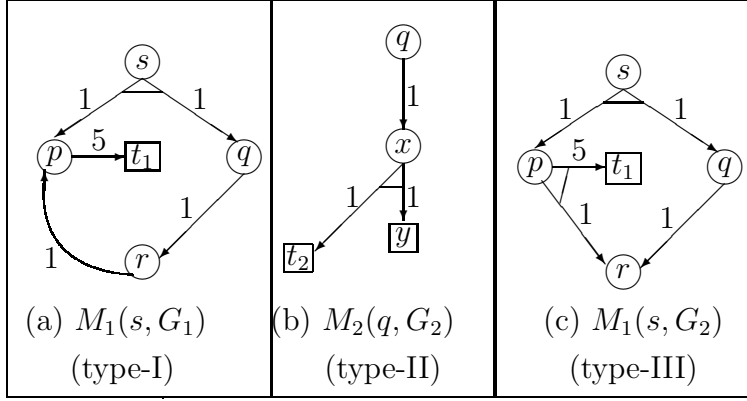
\begin{figure}
\centering
\caption{MESs from the implicit graphs of Figure \ref{fig70}} 
\label{fig80}
\begin{picture}(222.00,117.00)
\put(46.00,-0.33){\line(0,1){0.33}}
\put(40.00,99.00){\circle{10.00}}
\put(14.00,71.00){\circle{10.00}}
\put(70.00,71.00){\circle{10.00}}
\put(40.00,40.00){\circle{10.00}}
\put(70.00,66.00){\vector(-1,-1){25.00}}
\put(19.00,71.00){\vector(1,0){16.00}}
\put(40.00,94.00){\vector(-3,-2){26.00}}
\put(40.00,94.00){\vector(3,-2){28.00}}
\bezier{236}(37.00,36.00)(7.00,38.00)(11.00,67.00)
\put(11.00,61.00){\vector(0,1){6.00}}
\put(40.00,40.00){\makebox(0,0)[cc]{$r$}}
\put(70.00,71.00){\makebox(0,0)[cc]{$q$}}
\put(53.00,56.00){\makebox(0,0)[cc]{1}}
\put(13.00,41.00){\makebox(0,0)[cc]{1}}
\put(14.00,71.00){\makebox(0,0)[cc]{$p$}}
\put(27.00,76.00){\makebox(0,0)[cc]{5}}
\put(40.00,71.00){\makebox(0,0)[cc]{$t_{1}$}}
\put(23.00,88.00){\makebox(0,0)[cc]{1}}
\put(57.00,88.00){\makebox(0,0)[cc]{1}}
\put(40.00,99.00){\makebox(0,0)[cc]{$s$}}
\put(117.00,105.00){\circle{10.00}}
\put(117.00,74.00){\circle{10.00}}
\put(117.00,69.00){\vector(0,-1){17.00}}
\put(117.00,69.00){\vector(-1,-1){26.00}}
\put(117.00,100.00){\vector(0,-1){21.00}}
\put(120.00,59.00){\makebox(0,0)[cc]{1}}
\put(102.00,59.00){\makebox(0,0)[cc]{1}}
\put(117.00,74.00){\makebox(0,0)[cc]{$x$}}
\put(120.00,89.00){\makebox(0,0)[cc]{1}}
\put(117.00,105.00){\makebox(0,0)[cc]{$q$}}
\put(105.00,22.00){\makebox(0,0)[cc]{(b) $M_{2}(q,G_{2})$}} 
\put(34.00,90.00){\line(1,0){12.00}}
\put(110.00,62.00){\line(1,0){7.00}}
\put(82.00,37.00){\framebox(8.00,8.00)[cc]{$t_{2}$}}
\put(113.00,43.00){\framebox(8.00,8.00)[cc]{$y$}}
\put(35.00,67.00){\framebox(10.00,8.00)[cc]{}}
\put(32.00,22.00){\makebox(0,0)[cc]{(a) $M_{1}(s,G_{1})$}}
\put(182.00,96.00){\circle{10.00}}
\put(156.00,68.00){\circle{10.00}}
\put(212.00,68.00){\circle{10.00}}
\put(182.00,37.00){\circle{10.00}}
\put(158.00,64.00){\vector(3,-4){18.67}}
\put(212.00,63.00){\vector(-1,-1){25.00}}
\put(161.00,68.00){\vector(1,0){16.00}}
\put(182.00,91.00){\vector(-3,-2){26.00}}
\put(182.00,91.00){\vector(3,-2){28.00}}
\put(182.00,37.00){\makebox(0,0)[cc]{$r$}}
\put(212.00,68.00){\makebox(0,0)[cc]{$q$}}
\put(195.00,53.00){\makebox(0,0)[cc]{1}}
\put(171.00,54.00){\makebox(0,0)[cc]{1}}
\put(156.00,68.00){\makebox(0,0)[cc]{$p$}}
\put(169.00,73.00){\makebox(0,0)[cc]{5}}
\put(182.00,68.00){\makebox(0,0)[cc]{$t_{1}$}}
\put(165.00,85.00){\makebox(0,0)[cc]{1}}
\put(199.00,85.00){\makebox(0,0)[cc]{1}}
\put(182.00,96.00){\makebox(0,0)[cc]{$s$}}
\put(185.00,22.00){\makebox(0,0)[cc]{(c) $M_{1}(s,G_{2})$}}
\put(176.00,87.00){\line(1,0){12.00}}
\put(166.00,68.00){\line(-1,-4){2.67}}
\put(177.00,63.00){\framebox(10.00,9.00)[cc]{}}
\put(34.00,8.00){\makebox(0,0)[cc]{(type-I)}}
\put(111.00,8.00){\makebox(0,0)[cc]{(type-II)}}
\put(185.00,8.00){\makebox(0,0)[cc]{(type-III)}}
\put(0.00,0.00){\framebox(78.00,117.00)[cc]{}}
\put(78.00,0.00){\framebox(66.00,117.00)[cc]{}}
\put(144.00,0.00){\framebox(78.00,117.00)[cc]{}}
\end{picture}
\end{figure}

\begin{defi}  Given any implicit graph $G$, let $M_1(n,G)$, $M_2(n,G)$, $\ldots$,
be the all possible type-I or type-II MESs below $n$. Then $h^{*}(n)$, the cost of a minimal-cost MES below $n$ = $glb_{i \geq 1} \{\beta(n,M_i(n,G))\}$;
if no type-I or type-II MES exists below $n$, $h^{*}(n)$ is undefined.
\end{defi}

It may be easily verified that in Figure \ref{fig70}(a), 
$h^{*}(s) = 14$, and the only minimal-cost MES is $M_{1}(s,G_{1})$.
In Figure \ref{fig70}(b) $h^{*}(s)$ is undefined, while $h^{*}(q) = \infty$.

\begin{rem} 
For any node $n$ in an implicit graph $G$,
\end{rem}
\begin{description}
\item[({\em i})] if $n$ is of type-I, then $h^{*}(n) < \infty$;
\item[({\em ii})] if $n$ is of type-II, then $h^{*}(n) = \infty$;
\item[({\em iii})] if $n$ is of type-III, then $h^{*}(n)$ is undefined.
\end{description}

\begin{lem} \label{lem02.5}
Let $M(n)$ be a minimal-cost MES below a type-I or type-II node $n$ in $G$. Then for every node $x \in M(n)$, the sub-MES $\xi(x,M(n))$ is also a minimal-cost MES below $x$.
\end{lem}

Proof. Clear. \qed.

\subsection{MESs in an Explicit AND/OR Graph}

The concepts of MES and sub-MES, and the costs and properties of an MES, have been discussed in detail for an implicit AND/OR graph. Now in the context of an explicit AND/OR graph, we present similar concepts. 

In an explicit graph $G^{\prime}$, the child\_set is defined identically as in Definition \ref{def00}. $Z_{G^{\prime}}$ represents the set of all tip nodes in $G^{\prime}$, i.e. 
$Z_{G^{\prime}} = \{x \mid x \in G^{\prime} \bigwedge (\Gamma(x, G^{\prime}) = \phi) \}$.

\begin{defi} An MES $M(n,G^{\prime})$ in an explicit graph $G^{\prime}$ is defined
similarly as
in the implicit graph $G$, with $G$ replaced by $G^{\prime}$ throughout.
\end{defi}

As in the case for implicit graph, multiple MESs below a node $n \in G^{\prime}$
are named as $M_1(n, G^{\prime})$, $M_2(n, G^{\prime})$ \ldots.

\begin{defi} A sub-MES $\xi(p,M(n,G^{\prime}))$ below a node $p$ in an MES $M(n,G^{\prime})$ is defined similarly as a sub-MES below a node $p$ in an MES $M(n,G)$.
\end{defi}

\begin{rem}
\end{rem}

As explained in Remark \ref{submes} using Figure \ref{fig28} for implicit graphs, in case of explicit graphs as well a sub-MES may not itself be an MES. In this context, we may consider the entire implicit graph in Figure \ref{fig28}(a) as an explicit graph. Then the sub-MES $\xi(p,M_1(s,G^{\prime}))$, obtained from $M_1(s,G^{\prime})$ will not be an MES in $G^{\prime}$.

\subsubsection{Classification of MESs in Explicit Graphs}

In case of an implicit graph, the MESs could be distinguished into type-I, type-II and
type-III, as they are fully extended up to the leaf nodes of $G$, or immediately prior to the formation of a cycle. In an explicit graph, those MESs which have encountered cycles are labelled as type-III. On the other hand, those MESs which are yet to encounter cycles cannot be labelled as type-I or type-II. This is because, such MESs may not be fully extended up to the leaf nodes of $G$ yet. All such MESs are collectively labelled as non-type-III.
A non-type-III MES is identical to a potential solution graph (psg) defined in the context of an acyclic AND/OR graph. In this paper, we shall use these two terms interchangeably.

\begin{defi}
In an explicit graph $G^{\prime}$, 
an
MES $M(n,G^{\prime})$ is said to be a:
\begin{list}{\roman{defctr6})}{\usecounter{defctr6}}
\item {\bf non-type-III MES} (or a {\bf potential solution graph} or {\bf psg}), if $Z_{M(n,G^{\prime})} \subseteq
Z_{G^{\prime}}$;
\item {\bf type-III MES}, if $Z_{M(n,G^{\prime})} \not\subseteq Z_{G^{\prime}}$.
\end{list}
\end{defi}

For example, consider the explicit graph $G_{1}^{\prime}$ in Figure \ref{fig40}(a),
obtained by expanding nodes $s$, $n$ and $p$ from $G$ in Figure \ref{fig30}(a). We show a psg,
$M_1(n,G_1^{\prime})$ below $n$ in Figure \ref{fig40}(b).
When the explicit graph is augmented by expanding node $x$ to form $G_2^{\prime}$, the corresponding MES $M_1(n,G_2^{\prime})$, however, becomes type-III (shown in Figure \ref{fig40}(d)). Note that $M_1(n,G_1^{\prime})$ and $M_1(n,G_2^{\prime})$ look quite similar although they are actually different. In $M_1(n,G_1^{\prime})$ $x$ is a tip node, while in $M_1(n,G_2^{\prime})$ $x$ is an expanded node.

\unitlength=0.45mm
\special{em:linewidth 0.4pt}
\linethickness{0.4pt}
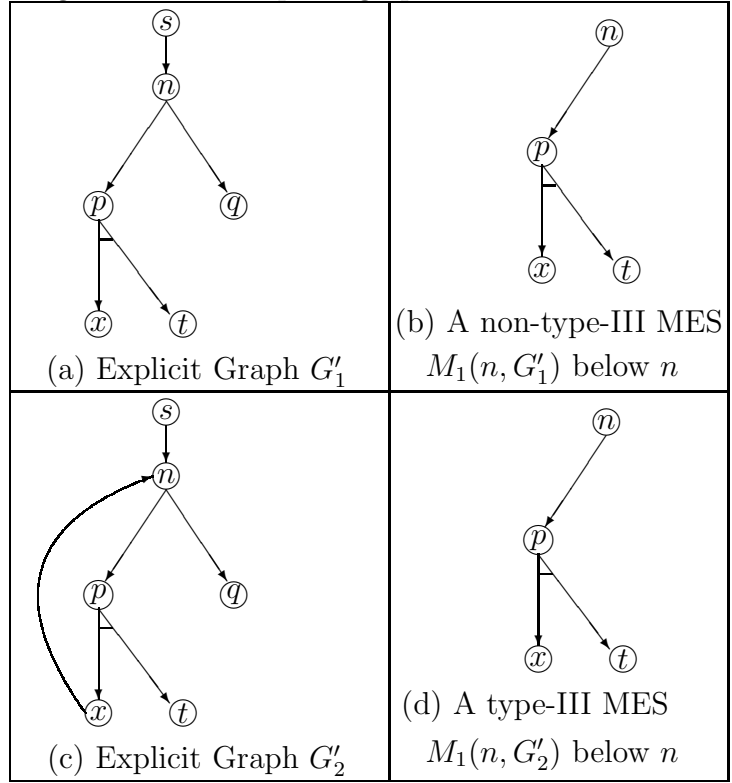
\begin{figure}
\centering
\caption{Two explicit graphs and their MESs} 
\label{fig40}
\begin{picture}(200.00,230.00)
\put(46.00,205.00){\circle{8.00}}
\put(26.00,170.00){\circle{8.25}}
\put(66.00,170.00){\circle{8.00}}
\put(26.00,135.00){\circle{8.00}}
\put(51.00,135.00){\circle{8.00}}
\put(46.00,201.00){\vector(-2,-3){18.00}}
\put(46.00,201.00){\vector(2,-3){18.00}}
\put(26.00,166.00){\vector(0,-1){27.00}}
\put(26.00,166.00){\vector(3,-4){21.00}}
\put(26.00,160.00){\line(1,0){4.00}}
\put(46.00,205.00){\makebox(0,0)[cc]{$n$}}
\put(66.00,170.00){\makebox(0,0)[cc]{$q$}}
\put(26.00,170.00){\makebox(0,0)[cc]{$p$}}
\put(26.00,135.00){\makebox(0,0)[cc]{$x$}}
\put(51.00,135.00){\makebox(0,0)[cc]{$t$}}
\put(55.00,121.00){\makebox(0,0)[cc]{(a) Explicit Graph $G_1^{\prime}$}}
\put(155.00,22.00){\makebox(0,0)[cc]{(d) A type-III MES}}
\put(160.00,8.00){\makebox(0,0)[cc]{$M_1(n,G_2^{\prime})$ below $n$}}
\put(46.00,224.00){\circle{8.00}}
\put(46.00,220.00){\vector(0,-1){11.00}}
\put(46.00,224.00){\makebox(0,0)[cc]{$s$}}
\put(177.00,221.00){\circle{8.00}}
\put(157.00,186.00){\circle{8.25}}
\put(157.00,151.00){\circle{8.00}}
\put(182.00,151.00){\circle{8.00}}
\put(177.00,217.00){\vector(-2,-3){18.00}}
\put(157.00,182.00){\vector(0,-1){27.00}}
\put(157.00,182.00){\vector(3,-4){21.00}}
\put(157.00,176.00){\line(1,0){4.00}}
\put(177.00,221.00){\makebox(0,0)[cc]{$n$}}
\put(157.00,186.00){\makebox(0,0)[cc]{$p$}}
\put(157.00,151.00){\makebox(0,0)[cc]{$x$}}
\put(182.00,151.00){\makebox(0,0)[cc]{$t$}}
\put(162.00,135.00){\makebox(0,0)[cc]{(b) A non-type-III MES}}
\put(46.00,90.00){\circle{8.00}}
\put(26.00,55.00){\circle{8.25}}
\put(66.00,55.00){\circle{8.00}}
\put(26.00,20.00){\circle{8.00}}
\put(51.00,20.00){\circle{8.00}}
\put(46.00,86.00){\vector(-2,-3){18.00}}
\put(46.00,86.00){\vector(2,-3){18.00}}
\put(26.00,51.00){\vector(0,-1){27.00}}
\put(26.00,51.00){\vector(3,-4){21.00}}
\put(26.00,45.00){\line(1,0){4.00}}
\put(46.00,90.00){\makebox(0,0)[cc]{$n$}}
\put(66.00,55.00){\makebox(0,0)[cc]{$q$}}
\put(26.00,55.00){\makebox(0,0)[cc]{$p$}}
\put(26.00,20.00){\makebox(0,0)[cc]{$x$}}
\put(51.00,20.00){\makebox(0,0)[cc]{$t$}}
\put(55.00,6.00){\makebox(0,0)[cc]{(c) Explicit Graph $G_2^{\prime}$}}
\put(46.00,109.00){\circle{8.00}}
\put(46.00,105.00){\vector(0,-1){11.00}}
\put(46.00,109.00){\makebox(0,0)[cc]{$s$}}
\bezier{484}(22.00,20.00)(-14.00,70.00)(42.00,90.00)
\put(39.00,89.00){\vector(3,1){3.00}}
\put(176.00,106.00){\circle{8.00}}
\put(156.00,71.00){\circle{8.25}}
\put(156.00,36.00){\circle{8.00}}
\put(181.00,36.00){\circle{8.00}}
\put(176.00,102.00){\vector(-2,-3){18.00}}
\put(156.00,67.00){\vector(0,-1){27.00}}
\put(156.00,67.00){\vector(3,-4){21.00}}
\put(156.00,61.00){\line(1,0){4.00}}
\put(176.00,106.00){\makebox(0,0)[cc]{$n$}}
\put(156.00,71.00){\makebox(0,0)[cc]{$p$}}
\put(156.00,36.00){\makebox(0,0)[cc]{$x$}}
\put(181.00,36.00){\makebox(0,0)[cc]{$t$}}
\put(160.00,122.00){\makebox(0,0)[cc]{$M_1(n,G_1^{\prime})$ below $n$}}
\put(112.00,0.00){\framebox(100.00,115.00)[cc]{}}
\put(0.00,0.00){\framebox(112.00,115.00)[cc]{}}
\put(112.00,115.00){\framebox(100.00,115.00)[cc]{}}
\put(0.00,115.00){\framebox(112.00,115.00)[cc]{}}
\end{picture}
\end{figure}

\subsubsection{Classification of Nodes in Explicit Graphs}

The nodes in an explicit graph are classified into two types depending on the type of MESs below them.

\begin{defi}
In $G^{\prime}$, a node is called non-type-III if it has a 
non-type-III MES (or psg) below it, otherwise it is called type-III.
\end{defi}

For example, in Figures \ref{fig40}(a) and \ref{fig40}(c), both $p$ and $n$ remain non-type-III before and after expansion of $x$. In both $G_1^{\prime}$ and $G_2^{\prime}$, $n$ has a non-type-III MES through $q$, while $p$ has a non-type-III MES through $\{x,t\}$ in $G_1^{\prime}$ and through $\{x,t,n,p\}$ in $G_2^{\prime}$.

\subsubsection {Properties of MESs in Explicit Graphs}

The MESs in an explicit graph follow the similar properties as do MESs in an implicit graph.

\begin{lem} \label{lem03} If $M(n,G^{\prime})$ is of non-type-III, then for every $p \in M(n,G^{\prime})$, the sub-MES $\xi(p,M(n,G^{\prime}))$ is also an MES and it is of non-type-III.
\end{lem}

{\bf Proof.} Clear. \qed

\begin{lem} \label{lem04} Let $M$ be any MES below a node
$n \in G^{\prime}$. Now if $M$ is of non-type-III, then every $p \in M$ is of non-type-III.
\end{lem}

{\bf Proof.} Clear from Lemma \ref{lem03}. \qed

\begin{lem}  \label{lem05} Let $\chi_1$ and $\chi_2$ be two acyclic AND/OR graphs 
with node and edge sets $V(\chi_1)$, $E(\chi_1)$ and $V(\chi_2)$, $E(\chi_2)$ respectively. Let $V(\chi_1) \cap V(\chi_2) = \phi$. Now, if edges \{$e_1$, $e_2$, \ldots , $e_n$\} are added from nodes in
$V(\chi_1)$ to nodes in $V(\chi_2)$, then the resulting graph with
$V = V(\chi_1) \cup V(\chi_2)$, $E = E(\chi_1) \cup E(\chi_2)
\cup \{e_1, e_2, \ldots ,e_n\}$ is acyclic. \\
\end{lem}

Proof. Clear. \qed\\

\begin{th} \label{thm25} {\bf Sub-problem Composition Theorem for Explicit Graphs.} Let $G^{\prime}$ be an explicit
AND/OR graph and n be any internal node in $G^{\prime}$. Now, \\
\newcounter{rom_ctr13}
\begin{list}{\roman{rom_ctr13})}{\usecounter{rom_ctr13}}
   \item If $n$ is an OR node, then:
   
   (a) $n$ is of non-type-III iff at least one child of $n$ is of non-type-III;
   
   (b) $n$ is of type-III iff no child of $n$ is of non-type-III.
   
   \item If $n$ is an AND node, then:
   
   (a)$n$ is of non-type-III iff every child of $n$ is of non-type-III
   
   (b) $n$ is of type-III iff at least one child of $n$ is of type-III.
\end{list}
\end{th}

{\bf Proof.} 
\newcounter{rom_ctr71}
\begin{list}{\roman{rom_ctr71})}{\usecounter{rom_ctr71}}
  
  \item ({\it a}) 
  
  $\Longrightarrow$\\
   Let $p$ be a child of $n$ in $G^{\prime}$. We assume $p$ is of non-type-III, and
   $M(p)$ is a non-type-III MES below $p$. We will show that $n$ is of non-type-III.\\
   \underline{Case I: $n \notin M(p)$:} Let $M(n)$ be the MES, created by joining the arc $(n,p)$ to $M(p)$. 
   Then $Z_{M(n)} = Z_{M(p)}  \subseteq Z_{G^{\prime}}$. 
   Then $n$ is also of non-type-III. \\      
   \underline{Case II: $n \in M(p)$:} Since $M(p)$ is of non-type-III,  by Lemma \ref{lem04} $n$ must be of non-type-III.
   \\
  $\Longleftarrow$\\
  Since it is given that $n$ is of non-type-III, there must be a non-type-III
  MES $M(n)$ below $n$. Let $p$ be the child of $n$ which belongs to $M(n)$.
  Clearly $p$ is of non-type-III (by Lemma \ref{lem04}).

({\it b}) Clear from ({\it a}).

\item ({\it a})

$\Longrightarrow$\\

Every child of $n$ is of non-type-III. We have to show that $n$ is of non-type-III.

Let $p_1, p_2, ... , p_k, k \ge 2$, be the children of $n$. Let 
$M(p_j)$ be a non-type-III MES below $p_j$. Clearly, $n \notin M(p_j) \forall j$, otherwise $M(p_j)$ would contain a cycle (since $n$ is an AND node and must include its child $p_j$ in any MES), thereby violating the acyclicity property of an MES.\\

Now, for proving that $n$ is of non-type-III, it is sufficient to construct a 
non-type-III MES $M(n)$ from the given non-type-III MESs $M(p_j)$ and the arcs ($n$,$p_j$), $1 \leq j \leq k$.\\

We construct $M(n)$ by first taking the nodes and arcs of $M(p_j)$s as follows. First, the MES $M(p_1)$ is taken in its entirety. We call this $Q_1$. Next, we augment $Q_1$ with nodes and arcs from $M(p_2)$. The selection of nodes and arcs from $M(p_2)$ is a recursive process that starts at $p_2$. If $p_2 \in Q_1$, we stop there and there is nothing to select from $M(p_2)$. Otherwise, we traverse the MES $M(p_2)$ in a depth-first manner, selecting nodes and arcs on the way, until we arrive either at a tip node of $M(p_2)$ or at a node $q$ in $Q_1$. In the former case, the selection process stops by selecting the tip node of $M(p_2)$; in the latter case, it stops at the node $q$ in $Q_1$. The nodes and arcs that are selected from $M(p_2)$ are added with $Q_1$, to form an augmented graph $Q_2$. Next, we take $Q_2$ and add nodes and arcs from $M(p_3)$ in a similar manner, to form $Q_3$. The process goes on like this, until we have considered each of the $M(p_j)$s and come up with a graph $Q_k$. We now present the procedure for the said construction of $Q_k$. In this construction, $V$ denotes the set of all nodes, and $E$ the set of all
directed arcs of an AND/OR graph.

\begin{tabbing}

tex\=mor\=stilltex\= \kill  

{\bf Procedure Construct\_Qk} \\

1.  Set $Q_1 = M(p_1)$ (i.e. $V(Q_1) = V(M(p_1))$ and $E(Q_1) = E(M(p_1))$); \\
        
2.  For $j$ = 2 to $k$, construct $Q_j$ as follows:\\
\>Initialize: $V(Q_j) = V(Q_{j-1})$ and $E(Q_j) = E(Q_{j-1})$;\\
\>Call Add($p_j$);\\ \\

{\bf Procedure Add($x$)} \\
 
      If $x \notin V(Q_j)$ \\
\>     $V(Q_j) = V(Q_j) \cup \{x\}$; \\
\>      If $x$ has no children in $M(p_j)$ \\
\>\> Return; \\
\>      Else  \\
\>         For every child $y$ of $x$ in $M(p_j)$ \\
\>         Begin \\
\>\>            $E(Q_j) = E(Q_j) \cup (x,y)$; \\
\>\>            Call Add($y$); \\
\>         End \\
      Return; \\
      
\end{tabbing}

Finally, we add the node $n$, and each of the arcs ($n$,$p_j$), $1 \leq j \leq k$, to $Q_k$, which gives us a structure called $M(n)$. We claim that $M(n)$ is a non-type-III MES below $n$. This is proved through the following three propositions.

\vspace{10pt}

\underline{{\bf Proposition 1.}} $Q_j$ is acyclic, $1 \leq j \leq k$.

The proof is by induction on $j$. For $j=1$, we have $Q_1 = M(p_1)$ which is an MES and must be acyclic.

Let us assume that all the graphs up to $Q_{j-1}$ are acyclic. Now $Q_j$ is constructed from $Q_{j-1}$ and $M(p_j)$ which are both acyclic; moreover we do not include any arc from a node in $V(Q_{j-1})$ to a node in $V(M(p_j)) \setminus V(Q_{j-1})$. Thus it follows from Lemma \ref{lem05} that $Q_j$ is acyclic.

Hence $Q_k$ is acyclic.

\vspace{10pt}

\underline{{\bf Proposition 2.}} For every node $m$ in $Q_k$, $Q_k$ contains a non-type-III MES below $m$.

The proof is by double induction, first on $j$ and then on the nodes of $Q_j$.

For $j=1$, the proposition holds for all nodes of $Q_1$, by Lemma \ref{lem03}.

Next we assume that the proposition holds up to $Q_{j-1}$. We show that it also holds for $Q_j$.

Let $V^{\prime} = V(Q_j) \setminus V(Q_{j-1})$. Let $m_1, m_2, \ldots, m_l=p_j$ be the nodes of $V^{\prime}$, sorted topologically.

\underline{$m_1$ is a tip node:} Then the MES below $m_1$ is $m_1$ itself and is clearly of non-type-III.

\underline{$m_1$ is not a tip node:} The children of $m_1$ in $M(p_j)$ (1 child if $m_1$ is OR node, all children if $m_1$ is AND node) must belong to $Q_{j-1}$. Since by hypothesis, every node in $Q_{j-1}$ has a non-type-III MES below it, and $Q_j$ is acyclic, it is clear that $m_1$ will have a non-type-III MES below it via its children in $Q_{j-1}$.

Thus the proposition holds for $m_1$. Let us assume that it holds for $m_i$, $i \geq 1$. We would show that it holds for $m_{i+1}$.

\underline{$m_{i+1}$ is a tip node:} Clear.

\underline{$m_{i+1}$ is not a tip node:} Since by hypothesis, every child of $m_{i+1}$ in $Q_j$ has a non-type-III MES below it, and $Q_j$ is acyclic, it is clear that $m_{i+1}$ will have a non-type-III MES below it via its children in $Q_j$. 

Thus the proposition holds for every node in $Q_j$.

\vspace{10pt}

\underline{{\bf Proposition 3.}} $M(n)$ is a non-type-III MES.

Observe that, $M(n)$ has been obtained by using the last $Q_j$, i.e. $Q_k$ and adding the node $n$ to it via the arcs ($n$,$p_j$), $1 \leq j \leq k$. Clearly, as $n \notin M(P_j) \forall j$, this resulting structure is acyclic. Thus $n$ will have a non-type-III MES through its children $p_1, p_2, \ldots, p_k$.

This proves that $n$ is a non-type-III node.\\ 

$\Longleftarrow$\\
       $n$ is a non-type-III node. Let $M(n)$ be a non-type-III MES below it, and $p_1, p_2, \ldots, p_k$ be the children of $n$. Then by Lemma \ref{lem03}, the sub-MES below every $p_j$ is a non-type-III MES, and every $p_j$ must be of non-type-III.
       
({\it b}) Clear from ({\it a}).         \qed\\

\end{list}

The importance of the Sub-problem Composition Theorems lies in the fact that they express 
the general notion of obtaining the solution graph at a node by using the solution graphs
of its children; the fact that this can be done at all, even in the presence of cycles as proved in the theorems above, will be used in our correctness proofs of the algorithms S1 and S2. Because of their fundamental importance, we illustrate the construction process used in the proof (routine Construct\_Qk) in Figure \ref{fig60} below.

\unitlength=0.45mm
\special{em:linewidth 0.4pt}
\linethickness{0.4pt}
\begin{figure}
\centering
\caption{Illustration of the Sub-problem Composition Theorem} 
\label{fig60}
\begin{picture}(330.00,133.00)
\put(33.00,124.00){\circle{10.00}}
\put(33.00,124.00){\makebox(0,0)[cc]{$n$}}
\put(7.00,95.00){\circle{10.20}}
\put(57.00,95.00){\circle{10.00}}
\put(57.00,95.00){\makebox(0,0)[cc]{$p_2$}}
\put(7.00,95.00){\makebox(0,0)[cc]{$p_1$}}
\put(7.00,60.00){\circle{10.00}}
\put(57.00,60.00){\circle{10.00}}
\put(32.00,67.00){\circle{10.00}}
\put(33.00,20.00){\circle{10.00}}
\put(7.00,90.00){\vector(0,-1){25.00}}
\put(33.00,119.00){\vector(-4,-3){26.00}}
\put(33.00,119.00){\vector(1,-1){21.00}}
\put(28.00,115.00){\line(1,0){9.00}}
\put(57.00,90.00){\vector(0,-1){25.00}}
\put(57.00,90.00){\vector(-2,-1){49.00}}
\put(7.00,90.00){\vector(2,-1){50.00}}
\bezier{208}(53.00,56.00)(34.00,41.00)(11.00,56.00)
\put(7.00,55.00){\vector(3,-4){23.33}}
\put(57.00,55.00){\vector(-2,-3){20.67}}
\put(7.00,60.00){\makebox(0,0)[cc]{$x$}}
\put(32.00,67.00){\makebox(0,0)[cc]{$y$}}
\put(57.00,60.00){\makebox(0,0)[cc]{$z$}}
\put(33.00,20.00){\makebox(0,0)[cc]{$t$}}
\bezier{76}(12.00,60.00)(17.00,67.00)(27.00,67.00)
\put(25.00,67.00){\vector(1,0){2.00}}
\put(12.00,55.00){\vector(-1,1){1.00}}
\bezier{68}(37.00,67.00)(46.00,67.00)(52.00,61.00)
\put(50.00,63.00){\vector(2,-1){2.00}}
\put(32.00,4.00){\makebox(0,0)[cc]{(a) Part of $G^{\prime}$}}
\put(72.00,95.00){\circle{10.20}}
\put(72.00,95.00){\makebox(0,0)[cc]{$p_1$}}
\put(72.00,60.00){\circle{10.00}}
\put(122.00,60.00){\circle{10.00}}
\put(98.00,20.00){\circle{10.00}}
\put(72.00,90.00){\vector(2,-1){50.00}}
\bezier{208}(118.00,56.00)(99.00,41.00)(76.00,56.00)
\put(72.00,55.00){\vector(3,-4){23.33}}
\put(72.00,60.00){\makebox(0,0)[cc]{$x$}}
\put(122.00,60.00){\makebox(0,0)[cc]{$z$}}
\put(98.00,20.00){\makebox(0,0)[cc]{$t$}}
\put(77.00,55.00){\vector(-1,1){1.00}}
\put(95.00,4.00){\makebox(0,0)[cc]{(b) $M(p_1)$}}
\put(188.00,95.00){\circle{10.00}}
\put(188.00,95.00){\makebox(0,0)[cc]{$p_2$}}
\put(138.00,60.00){\circle{10.00}}
\put(188.00,60.00){\circle{10.00}}
\put(163.00,67.00){\circle{10.00}}
\put(164.00,20.00){\circle{10.00}}
\put(188.00,90.00){\vector(-2,-1){49.00}}
\put(188.00,55.00){\vector(-2,-3){20.67}}
\put(138.00,60.00){\makebox(0,0)[cc]{$x$}}
\put(163.00,67.00){\makebox(0,0)[cc]{$y$}}
\put(188.00,60.00){\makebox(0,0)[cc]{$z$}}
\put(164.00,20.00){\makebox(0,0)[cc]{$t$}}
\bezier{76}(143.00,60.00)(148.00,67.00)(158.00,67.00)
\put(156.00,67.00){\vector(1,0){2.00}}
\bezier{68}(168.00,67.00)(177.00,67.00)(183.00,61.00)
\put(181.00,63.00){\vector(2,-1){2.00}}
\put(165.00,4.00){\makebox(0,0)[cc]{(c) $M(p_2)$}}
\put(208.00,94.00){\circle{10.20}}
\put(258.00,94.00){\circle{10.00}}
\put(258.00,94.00){\makebox(0,0)[cc]{$p_2$}}
\put(208.00,94.00){\makebox(0,0)[cc]{$p_1$}}
\put(208.00,59.00){\circle{10.00}}
\put(258.00,59.00){\circle{10.00}}
\put(234.00,19.00){\circle{10.00}}
\put(258.00,89.00){\vector(-2,-1){49.00}}
\put(208.00,89.00){\vector(2,-1){50.00}}
\bezier{208}(254.00,55.00)(235.00,40.00)(212.00,55.00)
\put(208.00,54.00){\vector(3,-4){23.33}}
\put(208.00,59.00){\makebox(0,0)[cc]{$x$}}
\put(258.00,59.00){\makebox(0,0)[cc]{$z$}}
\put(234.00,19.00){\makebox(0,0)[cc]{$t$}}
\put(213.00,54.00){\vector(-1,1){1.00}}
\put(230.00,4.00){\makebox(0,0)[cc]{(d) $Q_2$}}
\put(0.00,0.00){\framebox(66.00,133.00)[cc]{}}
\put(66.00,0.00){\framebox(64.00,133.00)[cc]{}}
\put(130.00,0.00){\framebox(70.00,133.00)[cc]{}}
\put(200.00,0.00){\framebox(65.00,133.00)[cc]{}}
\put(299.00,123.00){\circle{10.00}}
\put(299.00,123.00){\makebox(0,0)[cc]{$n$}}
\put(273.00,94.00){\circle{10.20}}
\put(323.00,94.00){\circle{10.00}}
\put(323.00,94.00){\makebox(0,0)[cc]{$p_2$}}
\put(273.00,94.00){\makebox(0,0)[cc]{$p_1$}}
\put(273.00,59.00){\circle{10.00}}
\put(323.00,59.00){\circle{10.00}}
\put(299.00,19.00){\circle{10.00}}
\put(299.00,118.00){\vector(-4,-3){26.00}}
\put(299.00,118.00){\vector(1,-1){21.00}}
\put(294.00,114.00){\line(1,0){9.00}}
\put(323.00,89.00){\vector(-2,-1){49.00}}
\put(273.00,89.00){\vector(2,-1){50.00}}
\bezier{208}(319.00,55.00)(300.00,40.00)(277.00,55.00)
\put(273.00,54.00){\vector(3,-4){23.33}}
\put(273.00,59.00){\makebox(0,0)[cc]{$x$}}
\put(323.00,59.00){\makebox(0,0)[cc]{$z$}}
\put(299.00,19.00){\makebox(0,0)[cc]{$t$}}
\put(278.00,54.00){\vector(-1,1){1.00}}
\put(295.00,4.00){\makebox(0,0)[cc]{(e) $M(n)$}}
\put(265.00,0.00){\framebox(65.00,133.00)[cc]{}}
\end{picture}
\end{figure}
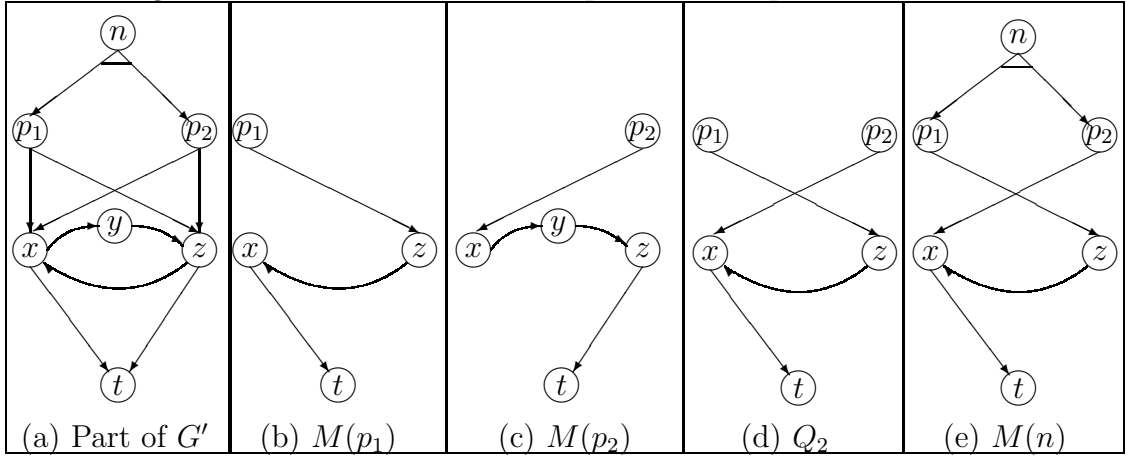

Consider the graph shown in Figure \ref{fig60}(a). Assume that it is a part of an explicit graph $G^{\prime}$ (not shown fully) below a node $n$ in it. Now two MESs below $p_1$ and $p_2$ are shown in Figure \ref{fig60}(b) and \ref{fig60}(c) respectively. The result of combining these two MESs to form an MES below $n$ using the 
Construct\_Qk routine is shown in Figure \ref{fig60}(e). Note that:

(\textit{i}) Although $M(p_1)$ and $M(p_2)$ taken together create a cycle, the construction of $M(n)$ is such that when $p_2$ is attempted to be included in it, the routine finds that $p_2$'s child $x$ is already there from $p_1$ in the previous construction. Hence, it retains the MES below $x$ as the MES below $p_2$ also. This strategy helps in avoiding the cycle between the MES of $p_1$ and the MES of $p_2$.

(\textit{ii}) While constructing $M(n)$ from the MESs of its children $p_i$, not all nodes from an MES below a child $p_i$ need be retained in the MES below $n$. For instance, in Figure \ref{fig60}(c) $y$ is a node in MES $M(p_2)$, but $y$ does not appear in the combined MES $M(n)$. However, as we've seen in the proof of the theorem, this does not introduce any error in the construction. 

\subsubsection{Computation of Costs for Explicit-graph MESs}

The costs of an MES in an explicit graph are defined in the same way as they are defined in the implicit graph.

\begin{defi}  \label{betaexpl} Given any explicit graph $G^{\prime}$, for any MES 
$M(n,G^{\prime})$ we denote the cost of a node $u$ in $M(n,G^{\prime})$ by 
$\beta(u, M(n,G^{\prime}))$. We define this cost function $\beta(u, M(n,G^{\prime}))$  as follows:

\begin{tabbing}
aaaaaaaaaaaa\=aaa\= \kill
$\beta(u,M(n,G^{\prime}))$ \\ 
= $0$  if $u \in  T$;  \\
=  $\infty$  if $u \in  NT$; \\
= $\hat{h}(u)$ if $u \in  Z_{M(n,G^{\prime})} \bigwedge u \not\in (T \bigcup NT)$, $\hat{h}(u)$ being the heuristic estimate at node $u$;\\
= $c(u,w) + \beta(w,M(n,G^{\prime}))$, if $u$ is an OR node and $w$ is the child of $u$ in $M(n,G^{\prime})$;  \\
= $\sum_{w \in \Gamma(u,M(n,G^{\prime}))} \{c(u,w) + \beta(w,M(n,G^{\prime})) \}$ if $u$ is an AND node.\\

\\

$\beta(u,M(n,G^{\prime}))$ is undefined if $u \in Z_{M(n,G^{\prime})} \setminus Z_{G^{\prime}}$.

\end{tabbing}
\end{defi}

\begin{defi} Let $M_{1}(n,G^{\prime})$, $M_{2}(n,G^{\prime})$, $\ldots$,
$M_{k}(n,G^{\prime})$ be the psgs below a node $n$ in an explicit graph
$G^{\prime}$. Then $h^{\prime}(n)$, the cost of a minimal-cost psg below $n$ =
$min_{1 \le i \le k} \{\beta(n,M_{i}(n,G^{\prime}))\}$; if there are no psgs below $n$ in $G^{\prime}$, $h^{\prime}(n)$ is undefined.
\end{defi}

\linethickness{0.4pt}
\begin{figure}
\centering
\caption{Explicit graphs from $G_{2}$ (Figure \ref{fig70}(b))} 
\label{fig90}
\begin{picture}(204.00,116.00)
\put(46.00,-0.33){\line(0,1){0.33}}
\put(54.00,92.00){\circle{10.00}}
\put(28.00,64.00){\circle{10.00}}
\put(84.00,64.00){\circle{10.00}}
\put(54.00,33.00){\circle{10.00}}
\put(30.00,60.00){\vector(3,-4){18.67}}
\put(33.00,64.00){\vector(1,0){16.00}}
\put(54.00,87.00){\vector(-3,-2){26.00}}
\put(54.00,87.00){\vector(3,-2){28.00}}
\put(54.00,33.00){\makebox(0,0)[cc]{$r$}}
\put(84.00,64.00){\makebox(0,0)[cc]{$q$}}
\put(54.00,44.00){\makebox(0,0)[cc]{(2)}}
\put(43.00,50.00){\makebox(0,0)[cc]{1}}
\put(28.00,64.00){\makebox(0,0)[cc]{$p$}}
\put(16.00,64.00){\makebox(0,0)[cc]{(5)}}
\put(41.00,69.00){\makebox(0,0)[cc]{5}}
\put(54.00,64.00){\makebox(0,0)[cc]{$t_{1}$}}
\put(37.00,81.00){\makebox(0,0)[cc]{1}}
\put(71.00,81.00){\makebox(0,0)[cc]{1}}
\put(54.00,92.00){\makebox(0,0)[cc]{$s$}}
\put(67.00,95.00){\makebox(0,0)[cc]{(0)}}
\put(98.00,64.00){\makebox(0,0)[cc]{(5)}}
\put(155.00,108.00){\circle{10.00}}
\put(129.00,80.00){\circle{10.00}}
\put(185.00,80.00){\circle{10.00}}
\put(155.00,49.00){\circle{10.00}}
\put(155.00,44.00){\vector(0,-1){23.00}}
\put(131.00,76.00){\vector(3,-4){18.67}}
\put(134.00,80.00){\vector(1,0){16.00}}
\put(155.00,103.00){\vector(-3,-2){26.00}}
\put(155.00,103.00){\vector(3,-2){28.00}}
\bezier{236}(152.00,45.00)(122.00,47.00)(126.00,76.00)
\put(126.00,70.00){\vector(0,1){6.00}}
\put(150.00,32.00){\makebox(0,0)[cc]{10}}
\put(155.00,49.00){\makebox(0,0)[cc]{$r$}}
\put(185.00,80.00){\makebox(0,0)[cc]{$q$}}
\put(155.00,58.00){\makebox(0,0)[cc]{(2)}}
\put(144.00,66.00){\makebox(0,0)[cc]{1}}
\put(128.00,50.00){\makebox(0,0)[cc]{1}}
\put(129.00,80.00){\makebox(0,0)[cc]{$p$}}
\put(117.00,80.00){\makebox(0,0)[cc]{(5)}}
\put(142.00,85.00){\makebox(0,0)[cc]{5}}
\put(155.00,80.00){\makebox(0,0)[cc]{$t_{1}$}}
\put(138.00,97.00){\makebox(0,0)[cc]{1}}
\put(172.00,97.00){\makebox(0,0)[cc]{1}}
\put(155.00,108.00){\makebox(0,0)[cc]{$s$}}
\put(168.00,108.00){\makebox(0,0)[cc]{(0)}}
\put(54.00,5.00){\makebox(0,0)[cc]{(a) $G_{1}^{\prime}$}}
\put(150.00,5.00){\makebox(0,0)[cc]{(b) $G_{2}^{\prime}$}}
\put(48.00,83.00){\line(1,0){12.00}}
\put(149.00,99.00){\line(1,0){12.00}}
\put(139.00,80.00){\line(-1,-4){2.67}}
\put(196.00,84.00){\makebox(0,0)[ct]{(5)}}
\put(150.00,12.00){\framebox(8.00,8.00)[cc]{$t_{2}$}}
\put(155.00,37.00){\line(-5,6){7.00}}
\put(151.00,76.00){\framebox(8.00,8.00)[cc]{}}
\put(49.00,60.00){\framebox(9.00,8.00)[cc]{}}
\put(41.00,64.00){\line(-2,-5){4.67}}
\put(108.00,0.00){\framebox(96.00,116.00)[cc]{}}
\put(0.00,0.00){\framebox(108.00,116.00)[cc]{}}
\end{picture}
\end{figure}
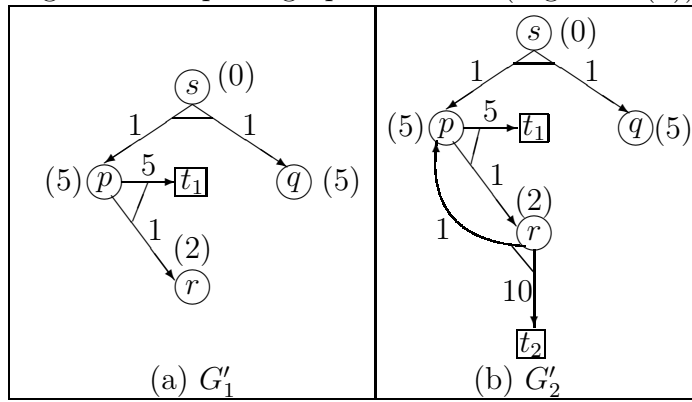

Thus in Figure \ref{fig90}(a) $h^{\prime}(r) = 2$, $h^{\prime}(t_1) = 0$, $h^{\prime}(q) = 5$, $h^{\prime}(p) = 8$ and $h^{\prime}(s) = 15$, while in Figure \ref{fig90}(b), $h^{\prime}(r)$, $h^{\prime}(p)$ and $h^{\prime}(s)$
are undefined as there are no psgs below them in $G_{2}^{\prime}$.

\section{\bf Algorithm S1}

We now present a best-first bottom-up algorithm, S1, that operates on the full 
implicit graph $G$. S1 assumes that $G$ contains
only finitely many nodes and arcs. The set of terminal and nonterminal leaves of $G$,
as well as the problem composition and decomposition rules, are needed as inputs to
S1. It also
maintains the $h$-value of a node to represent the currently known minimum cost of
solving the node. S1 uses two lists OPEN and CLOSED. S1 starts by putting all the leaf nodes of $G$ in OPEN with an $h$-value of $0$ or $\infty$
according as the leaf node is terminal or non-terminal. S1 then proceeds by removing a node
from OPEN
that has the minimum $h$-value. Whenever a node is removed from OPEN, it is put into CLOSED, and
all its parents are obtained by applying inverse operators. These parents are checked for possible
inclusion in OPEN and their $h$-values are updated, if
necessary, 
in an additive manner.

The point that needs
special mention here is the treatment of AND nodes. An AND node may be removed
from OPEN only if all of its children have already entered CLOSED. This is ensured
by using a label "eligible" to identify a subset of nodes in OPEN. The nodes
that are eligible are either OR nodes, or AND nodes with all their children in
CLOSED.

S1 continues in this manner until the start node is
removed from
OPEN or it is evident that the start node is of type-III. If the start node is not of type-III then $h(s)$ equals the cost of a minimal-cost solution graph and is outputted by S1. Note that, by appropriately maintaining pointers, S1 can trace the solution graph whose cost is outputted by it. For
simplicity, those
details have been left out in this paper.

\subsection{\bf Algorithm S1}

\begin{list}%
{{\bf S1.\arabic{S1ct1}}}{\usecounter{S1ct1}}
             
    \item Create a list, OPEN, and set $OPEN = Z_{G}$ (leaf nodes of $G$). 
    For each $n$ in OPEN, label $n$ as "eligible". Now, 
    if $n$ is a terminal node, set $h(n) = 0$; else set $h(n) = \infty$.
    
    \item Create a list, CLOSED, that is initially empty.
    \item While (OPEN contains an eligible node) do

        {  \begin{list}%
           {{\bf S1.\arabic{S1ct1}.\arabic{S1ct2}}}{\usecounter{S1ct2}}
                
           \item Find an eligible node $n$ from OPEN which has the minimum $h$-value. 
           (Resolve ties arbitrarily, but always in favour of the start node $s$.)
           Put $n$ in CLOSED.
           \item If $n = s$ then
              if $h(s) = \infty$, terminate with FAILURE;
                else output $h(s)$ and terminate with SUCCESS.
           \item Let $p_{1},p_{2},\ldots,p_{k}$ be the immediate predecessors of $n$ in $G$.
For each $p_{i}$, $1 \le i \le k$, do the following:
                 
                 \underline{{\bf Case I:} $p_{i}$ is an OR node} 
                 
                 If $p_{i}$ is not already in OPEN or CLOSED, set  $h(p_{i}) = h(n) +
c(p_{i},n)$. 
                 Put $p_{i}$ in OPEN and label $p_{i}$ as "eligible";                                 
                 elseif $p_{i}$ is already in OPEN with $h(p_{i}) > h(n) + c(p_{i},n)$ then set
                    $h(p_{i}) = h(n) + c(p_{i},n)$.

                 \underline{{\bf Case II:} $p_{i}$ is an AND node}
                  
                  If $p_{i}$ is not already present in OPEN, put it in OPEN 
                  and set $h(p_{i}) = c(p_{i},n) + h(n)$;
                  else set $h(p_{i}) = h(p_{i}) + c(p_{i},n) + h(n)$.
                  
                  If all children of $p_{i}$ are in CLOSED, label $p_{i}$ as "eligible".
                  
        \end{list}
        }
        \item Terminate with FAILURE.   $\Box$
\end{list}

\subsection{\bf Working of S1}

We now illustrate the working of algorithm S1. S1 is an uninformed search algorithm. Its working is shown on the implicit graph $G$
of Figure \ref{fig70}(a). In the figure, the arcs are labelled with their costs and the heuristic values of nodes are 
shown in parenthesis. However, these heuristic values are not for use by S1. They are to be used by algorithm S2 which is presented later. 
The iteration-by-iteration
working of S1 is presented in Table 1. Nodes in CLOSED are shown inside a square box. Among the nodes in OPEN, the non-underlined nodes are those which are marked eligible, and underlined nodes are those which are yet to become eligible. The $h$-value of each node is superscripted. The node
$n$ shown in the second column is the node selected by S1 from OPEN
in each iteration.

In step S1.1, OPEN is created with nodes $t_1$, $t_2$ and $y$ having $h$-values $0$, $0$ and $\infty$ respectively. All these nodes are marked "eligible" in OPEN. Then
in the first iteration $t_{1}$ is selected from OPEN and put in CLOSED with $h(t_1) = 0$, and
its parent $p$ is inserted into OPEN with $h(p) = 5$. The snapshots of OPEN and CLOSED at the end of iteration 1 are shown in the
first row with $t_{1}$ inside a square box.
In subsequent iterations $t_2$, $p$, $r$, $q$ and $s$ are selected from OPEN, one in each iteration, and put into CLOSED.
Finally, S1
terminates by finding a minimal-cost solution
of cost 14 of the solution graph $s, p, q, r, t_{1}$.

\renewcommand{\arraystretch}{1.6}
\setlength\fboxsep{0.05cm}

\begin{table}
\begin{centering}
\begin{tabular}{||l|l|l||}    \hline
Itn. & $n$ &  Nodes in OPEN and CLOSED                                    \\ 
No.  &     &  format $n^{h(n)}$                                           \\ \hline
 
1.   & $t_{1}$ & \fbox{$t_{1}^{0}$}, $t_{2}^{0}$, $y^{\infty}$, $p^{5}$                      \\ \hline
2.   & $t_{2}$ & \fbox{$t_{1}^{0}$}, \fbox{$t_{2}^{0}$}, $p^{5}$, $y^{\infty}$, $r^{10}$, \underline{$x^{1}$} \\ \hline
3.   & $p$     & \fbox{$t_{1}^{0}$}, \fbox{$t_{2}^{0}$}, \fbox{$p^{5}$}, $r^{6}$,
$\underline{x^{1}}$, $y^{\infty}$, \underline{$s^{6}$}     \\ \hline
4.   & $r$     & \fbox{$t_{1}^{0}$}, \fbox{$t_{2}^{0}$}, \fbox{$p^{5}$}, \fbox{$r^{6}$},
$y^{\infty}$, $\underline{x^{1}}$, $\underline{s^{6}}$, $q^{7}$     \\ \hline 
5.   & $q$     & \fbox{$t_{1}^{0}$}, \fbox{$t_{2}^{0}$}, \fbox{$p^{5}$}, \fbox{$r^{6}$},
\fbox{$q^{7}$}, $y^{\infty}$, $\underline{x^{1}}$, $s^{14}$      \\ \hline  
6.   & $s$     & \fbox{$t_{1}^{0}$}, \fbox{$t_{2}^{0}$}, \fbox{$p^{5}$}, \fbox{$r^{6}$},
\fbox{$q^{7}$}, \fbox{$s^{14}$}, $y^{\infty}$, $\underline{x^{1}}$        \\ \hline
\end{tabular}
\caption{Working of S1 on the graph of Figure \ref{fig70}(a)}
\end{centering}
\end{table}

\subsection{Analysis of S1}
The following definitions will be used in proving some of the properties of S1.

\begin{defi}  By an {\bf iteration} of S1, we mean one complete execution
of step S1.3, i.e.\ of substeps S1.3.1, S1.3.2 and S1.3.3 (unless S1 terminates at
S1.3.2, in which case the iteration consists of steps S1.3.1 and S1.3.2 only).
\end{defi}

\begin{defi}  By an {\bf instant}, we mean the time point when 
step S1.3 is about to be executed. Thus at instant $j$, S1.3 is executed for the
$j^{th}$ time.
\end{defi}

\begin{defi}  Let $p$ be a node in $G$, 
and $M(p)$ be a type-I or type-II MES below $p$. At any instant during the 
execution of S1, a node $n \in M(p)$ is a {\bf leading node} of $M(p)$ if 
({\em i}) $n$ is in OPEN \underline{\bf and} ({\em ii}) all its successors in $M(p)$, if any, are
in CLOSED. 
\end{defi}

\begin{lem} \label{lem08} If $M(p)$ is a type-I or
type-II MES below a  
node $p \in G$, then at any instant during the execution of S1, the following must
hold:
  
    (a) If $q$ is a leading node in $M(p)$, then $q$ must be eligible in OPEN;

    (b) If $p \not\in CLOSED$ then there must exist
    at least one leading node of $M(p)$.

\end{lem}

{\bf Proof.} 
({\em a}) Clearly, $q$ must be in OPEN, with all its successors from $M(p)$ in CLOSED
(by the definition of a leading node). This makes it eligible irrespective of whether
it is an AND node or an OR node.

({\em b}) We sort the nodes of $M(p)$ in topological
order, based on their height values $H$ in $M(p)$ 
(Such a topological sorting of nodes of $M(p)$ is possible, since it is
an MES which is acyclic by definition).
For any node $y \in M(p)$, its height
$H(y)$ in $M(p)$ is defined as follows:
\begin{tabbing}
aaaa\=aaa\= \kill
$H(y)$ \>=  0 if $y$ is a leaf node; \\
      \>=  $\max_{1 \le i \le k} \{1 + H(y_{i}) \}$, where $y_{1}, y_{2}, ..., y_{k}$    
           are children of $y$ in $M(p)$.
\end{tabbing}

Let the sorted list, in descending order of $H$-values,
be called $L$, and let a
sublist of $L$ be $L^{\prime} = L \setminus CLOSED$,
and let $x$ be the rightmost node
in $L^{\prime}$.

If $x$ is a leaf node, it is clearly a leading node of $M(p)$.

If $x$ is an internal node in $M(p)$, all its successors  are in CLOSED
(otherwise
$x$ could not be the rightmost node in $L^{\prime}$).
Also, $x$ must be in OPEN, as
({\em i}) $x \not\in CLOSED$, and
({\em ii}) all successors of $x$ in $M(p)$ are in CLOSED.
Hence $x$ must be a leading node of $M(p)$.  \qed\\

\begin{figure}
\centering
\caption{Illustration of Leading node}
\label{fig100}
\begin{picture}(168.00,112.00)
\put(53.00,104.00){\circle{8.00}}
\put(13.00,58.00){\circle{8.00}}
\put(46.00,58.00){\circle{8.00}}
\put(32.00,35.00){\circle{8.00}}
\put(30.00,31.00){\vector(-3,-2){14.00}}
\put(27.00,22.00){\makebox(0,0)[cc]{10}}
\put(13.00,54.00){\vector(0,-1){31.00}}
\put(32.00,35.00){\makebox(0,0)[cc]{$x$}}
\put(19.00,38.00){\makebox(0,0)[cc]{10}}
\put(13.00,58.00){\makebox(0,0)[cc]{$r$}}
\put(31.00,88.00){\circle{8.00}}
\put(31.00,84.00){\vector(-2,-3){15.33}}
\put(31.00,84.00){\vector(1,-2){11.00}}
\put(21.00,73.00){\makebox(0,0)[cc]{1}}
\put(31.00,88.00){\makebox(0,0)[cc]{$p$}}
\put(39.00,73.00){\makebox(0,0)[cc]{1}}
\put(46.00,58.00){\makebox(0,0)[cc]{$q$}}
\put(46.00,54.00){\vector(-2,-3){10.67}}
\put(46.00,54.00){\vector(2,-3){10.67}}
\put(55.00,46.00){\makebox(0,0)[cc]{1}}
\put(38.00,46.00){\makebox(0,0)[cc]{1}}
\put(53.00,104.00){\makebox(0,0)[cc]{$s$}}
\put(53.00,100.00){\vector(-2,-1){19.00}}
\bezier{44}(27.00,78.00)(31.00,74.00)(34.00,78.00)
\put(45.00,4.00){\makebox(0,0)[cc]{(a) Implicit Graph $G$}}
\put(41.00,98.00){\makebox(0,0)[cc]{1}}
\put(126.00,59.00){\circle{8.00}}
\put(159.00,59.00){\circle{8.00}}
\put(145.00,36.00){\circle{8.00}}
\put(143.00,32.00){\vector(-3,-2){14.00}}
\put(140.00,23.00){\makebox(0,0)[cc]{10}}
\put(126.00,55.00){\vector(0,-1){31.00}}
\put(145.00,36.00){\makebox(0,0)[cc]{$x$}}
\put(132.00,39.00){\makebox(0,0)[cc]{10}}
\put(126.00,59.00){\makebox(0,0)[cc]{$r$}}
\put(144.00,89.00){\circle{8.00}}
\put(144.00,85.00){\vector(-2,-3){15.33}}
\put(144.00,85.00){\vector(1,-2){11.00}}
\put(134.00,74.00){\makebox(0,0)[cc]{1}}
\put(144.00,89.00){\makebox(0,0)[cc]{$p$}}
\put(152.00,74.00){\makebox(0,0)[cc]{1}}
\put(159.00,59.00){\makebox(0,0)[cc]{$q$}}
\put(159.00,55.00){\vector(-2,-3){10.67}}
\put(151.00,47.00){\makebox(0,0)[cc]{1}}
\bezier{44}(140.00,79.00)(144.00,75.00)(147.00,79.00)
\put(137.00,5.00){\makebox(0,0)[cc]{(b) MES $M(p)$}}
\put(7.00,14.00){\framebox(8.00,8.00)[cc]{$t_{2}$}}
\put(120.00,15.00){\framebox(8.00,8.00)[cc]{$t_{2}$}}
\put(53.00,29.00){\framebox(8.00,8.00)[cc]{$t_{1}$}}
\put(100.00,0.00){\framebox(68.00,112.00)[cc]{}}
\put(0.00,0.00){\framebox(100.00,112.00)[cc]{}}
\end{picture}
\end{figure}
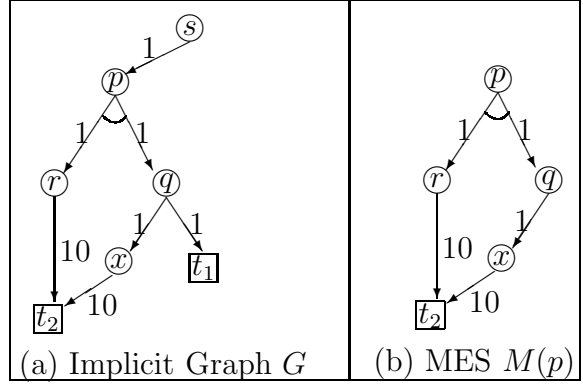

\begin{rem}
\end{rem}
If $q \in M(p)$ is eligible in OPEN then $q$ need not be a leading node of $M(p)$.
This can clearly be illustrated from the following example.
For the implicit graph $G$ shown
in Figure \ref{fig100}(a), we consider the MES shown in Figure \ref{fig100}(b).
After the first two instants of S1, $t_{2}$ and $t_{1}$
have travelled to CLOSED. However,
although $q \in M(p)$ is eligible, it is not a leading node
in $M(p)$. Here $x$ and $r$
are leading nodes in $M(p)$.

\linethickness{0.4pt}

\begin{lem} \label{lem10} When S1 runs on a finite AND/OR graph $G$, 
at any instant $i$, no type-III node ever enters CLOSED. \end{lem}

{\bf Proof.} This may be easily seen from the following statements.

(a) Only eligible nodes from OPEN can enter CLOSED.

(b) No eligible node can ever be of type-III.

Statement (a) is clear from Step S1.3.1 of algorithm S1 ;
the proof of (b) follows. The proof is by induction on the instant $i$ of S1.

\underline {For $i = 1$:} The only eligible nodes
are leaf nodes, which are of type-I or type-II.

\underline  {Up to $i = k$:} We assume that no node that has
become eligible is of type-III.

\underline {$i = k+1$ :}
Let $n$ be the node selected from OPEN and sent to CLOSED at
instant $k$ and let $p$ be a parent of $n$.
Now if $p$ becomes eligible in OPEN at instant $k+1$,
we show that $p$ can not be of type-III.

\underline {Case I:} $p$ is an OR node. Since, by assumption, $n$ is of type-I or type-II,
$p$ would also be of type-I or type-II (by the Sub-problem Composition Theorem for Implicit Graphs).

\underline {Case II:} $p$ is an AND node.
Since $p$ is now becoming eligible (by assumption), all the children
$p_1,p_2,\ldots,p_l$ of $p$ must now be in
CLOSED, i.e.\ they must have
been eligible at some instant prior to instant $k+1$.
By hypothesis, none of these children of $p$ can be of
type-III. Therefore, by the Sub-problem Composition Theorem for Implicit Graphs, $p$ cannot be of type-III, either.   \qed\\

\begin{lem} \label{lem20} When S1 sends a node n to CLOSED, $h(n) = h^{*}(n)$. 
\end{lem}

{\bf Proof:} 
Let $n_{i}$ be the node travelling to CLOSED at instant $i$, $i = 1,2,\ldots$. 
We prove the lemma by induction on $i$.

This lemma is applicable when a node travels to CLOSED. At instant 1, $n_{1}$ is
clearly a leaf node with its $h$-value set
to $h^{*}$ (0 or $\infty$ according as $n_{i}$ is terminal or nonterminal).
We assume that
 algorithm S1 has put nodes $n_{1}, n_{2},\ldots ,n_{k}$ in
CLOSED with $h(n_{i}) = h^{*}(n_{i}), 1 \leq i \leq k$. We need to show that $n_{k+1}$ goes to CLOSED with
$h(n_{k+1}) = h^*(n_{k+1})$. Clearly, by Lemma \ref{lem10}, $n_{k+1}$ must be
either of type-I or type-II.\\

\underline {Case I :} $n_{k+1}$ is a leaf node. Trivially true. \\

\underline {Case II :} $n_{k+1}$ is an AND node. Clearly, since $n_{k+1}$ is now eligible,
all its children must have been
previously put into CLOSED with their $h$-values = $h^{*}$ values (by hypothesis).
Now, since S1 computes the $h$-value of an AND node by successively
adding the $h$-values of its children when each of them is selected from OPEN,
then $h(n_{k+1}) = h^{*}(n_{k+1})$ for the AND node $n_{k+1}$. \\

\underline {Case III:} $n_{k+1}$ is an OR node. Let $p$ be the child through which $n_{k+1}$ had
last received its $h$-value prior to getting selected from OPEN, i.e. $h(n_{k+1}) = c(n_{k+1},p) + h(p)$. 
If $h(n_{k+1}) \not= h^{*}(n_{k+1})$, let $q$ be the child of $n_{k+1}$ in a minimal-cost MES $M$ below $n_{k+1}$, i.e. $h^{*}(n_{k+1}) = c(n_{k+1},q) + h^{*}(q)$. And, it must be the case
that $q$ has not travelled to CLOSED yet. We shall show that this leads to a contradiction. 

Let $\xi(q,M(n_{k+1}))$ be the sub-MES below $q$ in $M(n_{k+1})$.
By Lemma \ref{lem01}, $\xi(q,M(n_{k+1}))$ must be of type-I or type-II as $M$ is of  type-I or type-II.
Moreover, by Lemma \ref{lem08}, as $q$ has not entered CLOSED,
$\xi(q,M(n_{k+1}))$ must have leading node(s) at instant $k+1$.

Let $q_{1}, q_{2},\ldots ,q_{m} (m \geq 1)$ be the leading nodes of $\xi(q,M(n_{k+1}))$ at instant
$k+1$.

Now, from the definition of a leading node, each $q_{j}$
must have all its children from $\xi(q,M(n_{k+1}))$ in CLOSED.
Clearly, by the induction hypothesis, each of these children must
have its $h = h^{*}$.

Further, $\xi(q,M(n_{k+1}))$ is a minimal-cost MES below $q$, and all
the children
of $q_j$ from $\xi(q,M(n_{k+1}))$ have already travelled to CLOSED
and updated the $h(q_j)$ value, if needed. Thus $h(q_j) = h^{*}(q_j)$, $1 \leq j \leq m$.
  
Therefore, when $n_{k+1}$ is selected from OPEN at instant $k+1$, every $q_{j},
1 \leq j \leq m,$  has,

\begin{tabbing}
aaaaa\=aaaa\= \kill
$h(q_{j})$ \> =  $h^{*}(q_{j})$ \\
           \> $ < h^{*}(n_{k+1})$ \\ 
           \> $< c(n_{k+1},p) + h^{*}(p)$, as $p$ is $n_{k+1}$'s child \\
           \> = $c(n_{k+1},p) + h(p)$, by induction hypothesis \\
           \> = $h(n_{k+1})$,  as assumed previously
\end{tabbing}
 Therefore, when node $n_{k+1}$ was selected from OPEN, 
   $q_{j}$, being a leading node, was also eligible in OPEN with $h(q_{j}) < h(n_{k+1})$.
   This is clearly in contradiction to the best-first node selection criterion,
   on the basis of minimum $h$, used by S1. \qed\\

\begin{th}\label{thm28}
S1, while running on a finite AND/OR graph $G$:
\begin{enumerate}
  \item terminates with SUCCESS, outputting $h(s) = h^*(s)$ if $G$ contains at least
  one solution graph;
  \item terminates with FAILURE, otherwise.
\end{enumerate}
\end{th}

{\bf Proof.} 
\begin{enumerate}
\item \underline{($G$ contains a solution graph.)} As $G$ contains only finitely many nodes, no
node returns to OPEN from 
CLOSED, and in each iteration one eligible node is removed from OPEN and put into
CLOSED, S1 can continue for finitely many iterations only.

Let $M$ be a minimal-cost type-I MES below $s$ in $G$. At any instant before $s$
goes to CLOSED, there will always be at least one leading node $n$ from $M$, by Lemma \ref{lem08}. $n$, being a leading node of a minimal-cost type-I MES, must have $h(n) = h^{*}(n) \leq h^{*}(s)$. Note that $n$ is also
eligible in OPEN. Now $n$ cannot be left in OPEN indefinitely, since S1 runs for only finitely many iterations and selects eligible nodes from OPEN on the basis of minimum $h$-value. Thus eventually $n$ is bound to be selected from OPEN and
put into CLOSED. When $n = s$, the algorithm will terminate with SUCCESS, outputting
$h(s) = h^*(s)$.

\item \underline{($G$ does not contain a solution graph.)} If $s$ is of type-II,
consider the argument in (1) above. Here also the argument follows surrounding the
key concepts of a minimal-cost type-II MES $M$ below $s$, and the leading nodes of $M$.
Ultimately $s$ will enter CLOSED with $h(s) = h^*(s) = \infty$, and S1 terminates
with FAILURE.

If $s$ is of type-III, since no type-III node enters CLOSED, the algorithm will
continue as long as there are type-I and type-II nodes in $G$. Since there are only
finitely many such nodes, ultimately OPEN will become empty of eligible nodes, and
S1 will terminate with FAILURE. \qed\\

\end{enumerate}

\begin{th}\label{thm30} Let $G$ be an AND/OR graph with finitely many nodes and arcs.
When
S1 runs on $G$, it makes
\begin{enumerate}
  \item exactly $N_1$ iterations, where 

  $N_1 = 1 + \mid \{n \mid n$ is a type-I node with $h^*(n) < h^*(s)\}\mid$, if $G$
  contains a solution graph;

  \item at most $N_1$ iterations, where

  $N_1 = \mid \{n \mid n$ is a type-I or a type-II node $\} \mid$, if $G$ contains
  no solution graph.
\end{enumerate}
\end{th}

{\bf Proof.} At each iteration of S1 before it terminates, one (new) eligible node is selected
from OPEN and put into CLOSED.
\begin{enumerate}
\item If $G$ contains a solution graph, i.e.\ $s$ is of type-I, let $M$ be a minimal-cost
type-I MES below $s$. Then, at every iteration before $s$ is selected, there will
be at least one leading node $n$ of $M$, such that $n$ is eligible in OPEN with
$h(n) = h^*(n) < h^*(s)$. Thus, at every iteration $i$, before $s$ is sent to CLOSED,
if the node $p$ is selected from OPEN at iteration $i$, $h(p) \le h(n) = h^*(n) < h^*(s) < \infty$
(since $G$ has a solution graph). Besides these iterations, one more iteration will be
there with $p = n = s$ and $h(p) = h^*(p) = h^*(s)$.

\item Since no type-III node enters CLOSED, the only nodes entering CLOSED are of
type-I or type-II, and the lemma follows easily.\qed\\
\end{enumerate}

\begin{defi} By a {\bf node evaluation} we mean a single computation of
$h$-values at step S1.3.3. Similarly by a {\bf node selection}, we mean selecting an
eligible node from OPEN at step S1.3.1.
\end{defi} \

\begin{th}\label{thm40} Given any finite AND/OR graph $G$, the following are true
about the execution
of S1: 

\begin{enumerate} 
  \item S1 makes $O(N_1)$ node selections  
  \item S1 makes $O(N_1K_1)$ node evaluations, where $N_1$ is as defined in Theorem \ref{thm30},
  and $K_1$ is the total number of nodes in $G$.
\end{enumerate}           
\end{th}
{\bf Proof.} \begin{enumerate}
\item Clear, since there are $O(N_1)$ iterations and in  each iteration there is
exactly one node selection.

\item There are $O(N_1)$ iterations of step S1.3, in each of which one node is selected
and $O(K_1)$ parents of a node may get evaluated.     \qed\\
\end{enumerate}

\section{\bf ALGORITHM S2}
S2 is an improved version of the uninformed search algorithm S1. It resembles $\mbox{AO}^{*}$ and does heuristically-guided search in a top-down fashion. S2 works on an implicit
AND/OR graph $G$, which is either finite, or infinite containing at least one solution graph. Thus S2 can work effectively on graphs having paths of infinite length, which is not possible by S1 due to its entirely bottom-up nature beginning from the leaf nodes.

S2 contains a procedure Bottom\_Up that works in a manner identical to S1 on explicit graphs.
Additionally, S2 maintains a variable, $front$, with every node. The purpose of 
the variable $front$ with any node $n$ is to identify one of its successors $q$ which is 
preferably an unsolved tip node of a least-costly psg below $n$. Thus in any iteration
prior to the termination of S2, the $front$ of $s$ is an unsolved tip node of a
least-costly psg below $s$, and is the candidate node to be expanded next.

\subsection{\bf Algorithm S2}

\begin{list}
{{\bf S2.\arabic{S2ctr1}}}{\usecounter{S2ctr1}}  

\item Create an explicit graph $G^{\prime}$ consisting solely of the start node $s$. 
Set $front(s) = s$. If $s$ is a terminal leaf set $h(s) = 0$; else if $s$ is 
a nonterminal leaf set $h(s) = \infty$. 
\item While ( ($front(s)$ is not a terminal leaf) and ($h(s) \not = \infty$) ) do:
        \begin{list}
        {{\bf S2.\arabic{S2ctr1}.\arabic{S2ctr2}}}{\usecounter{S2ctr2}}     
        
        \item Let $n = front(s)$. Expand $n$, generating all its children $n_{1},
n_{2},\ldots,n_{k}$. 
        Install each $n_{i}$ in $G^{\prime}$ as child of $n$, by setting the arc ($n,n_{i}$). For
        each newly occurring node $n_{i}$ in $G^{\prime}$ set $front(n_{i}) = n_{i}$. 
        If $n_{i}$ is a terminal leaf set $h(n_{i}) = 0$; else if $n_{i}$ is a 
        nonterminal leaf set $h(n_{i}) = \infty$; else set $h(n_{i}) = \hat{h}(n_{i})$.

        \item Set $OPEN = Z_{G^{\prime}}$. Label all the nodes in OPEN as {\bf eligible}, and
{\bf initial}.
        \item Call Bottom\_Up(OPEN). 
        \end{list}
\item If $front(s)$ is a terminal leaf node, output $h(s)$ and terminate with SUCCESS; else
terminate with
FAILURE.                                                                                                       
\end{list}

PROCEDURE Bottom\_Up (List OPEN)
\begin{list}
{{\bf B\arabic{S2ctr3}}}{\usecounter{S2ctr3}}

\item Initialize a list, CLOSED, to nil.

\item While (OPEN contains an eligible node and $s \not\in CLOSED$) do:
        \begin{list}{{\bf B\arabic{S2ctr3}.\arabic{S2ctr4}}}{\usecounter{S2ctr4}}
        \item Select an eligible node $q$ from OPEN that has minimum $h$-value. (Resolve
        ties arbitrarily, but always in favour of $s$).
        \item If $q$ is not an initial node, then do the following:
              
              Let $q_{1}, q_{2}, \ldots, q_{r}$ be the children of $q$ in $G^{\prime}$ 
              which are in CLOSED.    

              \underline{{\bf Case I:}} $q$ is an OR node.
                
                Let $\tau = min_{1 \leq i \leq r} \{c(q,q_{i}) + h(q_{i})\}$  occur for $i=j$ 
                (resolve ties arbitrarily, but in favour of a node whose front 
                is a terminal leaf). Set $front(q) = front(q_{j})$.
              
              \underline{{\bf Case II:}} $q$ is an AND node.

                Let $q_{j}$ be the leftmost child of $q$ whose front is not a terminal leaf.
                If no such $q_{j}$ exists (i.e. every  child of $q$ has a terminal leaf as its 
                front) set $front(q) = front(q_{1})$; else set $front(q) = front(q_{j})$.

        \item Put $q$ in CLOSED. Let $p_{1}, p_{2},\ldots, p_{k}$ be the parents of $q$ in
$G^{\prime}$. 
              For each $p_{i}$ do:  
   
                \underline{{\bf Case I:}} $p_{i}$ is an OR node. 
       
       If $p_{i}$ is not already present in OPEN or CLOSED, set $h(p_{i}) = h(q) +
c(p_{i},q)$. Put $p_{i}$ in OPEN and
       mark it eligible; elseif $p_{i}$ is already present in OPEN with $h(p_{i}) > h(q) +
c(p_{i},q)$, set $h(p_{i}) =
       h(q) + c(p_{i},q)$.
                
                \underline{{\bf Case II:}} $p_{i}$ is an AND node.  
       
       If $p_{i}$ is not already present in OPEN, put it in OPEN and set $h(p_{i}) = h(q) +
c(p_{i},q)$; else set $h(p_{i})
       = h(p_{i}) + c(p_{i},q) + h(q)$. If all children of $p_{i}$ are in CLOSED, mark $p_{i}$
as eligible.            
        
        \end{list}
\item Remove any remaining nodes from OPEN.
                                                                                                         
\item If $s \not \in CLOSED$ set $h(s) = \infty$.  $\Box$   
\end{list}

\subsection{\bf Working of S2}

In Tables 2 and 3, we present the working of S2 on the graphs of Figures \ref{fig70}(a) and \ref{fig70}(b).
Each iteration of S2 is quite similar to the working of S1 as presented in Table 1. Here, $n$ is the node which is expanded in each iteration. With each node, an additional variable "front"' is added. Tip nodes have themselves as their $front$s; other (internal)
nodes initially have their $front$s "carried over" from the previous iteration,
and later the $front$s are decided when these internal nodes enter CLOSED.
In each iteration, the first row of column three shows the tip nodes of the explicit graph, that are initially put in OPEN and labeled as "eligible" and "initial". Nodes that are not eligible are underlined. CLOSED nodes are put in rectangular boxes. The $h$ and $front$ values of a node are superscripted above it.
S2 makes use of heuristic values and runs in the top-down fashion.
For the graph $G_1$ in Figure \ref{fig70}(a), S2 outputs 
$h^{*}(s) = 14$, i.e.\ the cost of a minimal-cost solution graph. For the graph $G_2$ in Figure \ref{fig70}(b), S2 terminates with FAILURE as there is no solution graph below $s$.

\renewcommand{\arraystretch}{1.6}

\setlength\fboxsep{0.05cm}

\begin{table}
\begin{centering}
\begin{tabular}{||l|l|l|c||}    \hline
$Itn$ & Exp.  &  Nodes in OPEN and CLOSED & Explicit graph after each iteration \\ 
    &  node  & in Bottom\_Up computation & \\
    &   $n$   &  format $m^{h(m),front(m)}$  & \\ \hline
 
1.    & $s$ &  $p^{5,p}$, $q^{5,q}$ (Initial OPEN) & \\ \cline{3-3}
    &  &  \fbox{$p^{5,p}$}, $q^{5,q}$, $\underline{s}^{6,s}$ & \\ \cline{3-3}
       &     &  \fbox{$p^{5,p}$}, \fbox{$q^{5,q}$}, $s^{12,s}$     & \\ \cline{3-3}

       &     &  \fbox{$p^{5,p}$}, \fbox{$q^{5,q}$}, \fbox{$s^{12,p}$}
  &  
\raisebox{0pt}[0pt]{
\linethickness{0.4pt}
\begin{picture}(84.00,36.00)
\put(46.00,-0.33){\line(0,1){0.33}}
\put(40.00,31.00){\circle{10.00}}
\put(14.00,3.00){\circle{10.00}}
\put(70.00,3.00){\circle{10.00}}
\put(40.00,26.00){\vector(-3,-2){26.00}}
\put(40.00,26.00){\vector(3,-2){28.00}}
\put(70.00,3.00){\makebox(0,0)[cc]{$q$}}
\put(14.00,3.00){\makebox(0,0)[cc]{$p$}}
\put(2.00,3.00){\makebox(0,0)[cc]{5,$p$}}
\put(23.00,20.00){\makebox(0,0)[cc]{1}}
\put(57.00,20.00){\makebox(0,0)[cc]{1}}
\put(40.00,31.00){\makebox(0,0)[cc]{$s$}}
\put(53.00,30.00){\makebox(0,0)[cc]{12,$p$}}
\put(84.00,3.00){\makebox(0,0)[cc]{5,$q$}}
\put(34.00,22.00){\line(1,0){12.00}}
\end{picture}
}

\\ \cline{1-4}

 2.    & $p$     &  $t_{1}^{0,t_{1}}$, $r^{2,r}$, $q^{5,q}$ (Initial OPEN) & \\  \cline{3-3}
    &      &  \fbox{$t_{1}^{0,t_{1}}$}, $r^{2,r}$, $q^{5,q}$, $p^{5,p}$ & \\  \cline{3-3} 
     &      &  \fbox{$t_{1}^{0,t_{1}}$}, \fbox{$r^{2,r}$}, $q^{5,q}$, $p^{3,p}$  & \\  \cline{3-3}
     &      &  \fbox{$t_{1}^{0,t_{1}}$}, \fbox{$r^{2,r}$}, $q^{5,q}$, \fbox{$p^{3,r}$}, $\underline{s}^{4,p}$  & \\  \cline{3-3}
     &      &  \fbox{$t_{1}^{0,t_{1}}$}, \fbox{$r^{2,r}$}, \fbox{$q^{5,q}$}, \fbox{$p^{3,r}$}, $s^{10,p}$  & \\ \cline{3-3}
     &      &  \fbox{$t_{1}^{0,t_{1}}$}, \fbox{$r^{2,r}$}, \fbox{$q^{5,q}$}, \fbox{$p^{3,r}$}, \fbox{$s^{10,r}$} & 
\raisebox{0pt}[0pt]{     
\linethickness{0.4pt}
\begin{picture}(84.00,68.00)
\put(46.00,-0.33){\line(0,1){0.33}}
\put(40.00,63.00){\circle{10.00}}
\put(14.00,35.00){\circle{10.00}}
\put(70.00,35.00){\circle{10.00}}
\put(40.00,4.00){\circle{10.00}}
\put(16.00,31.00){\vector(3,-4){18.67}}
\put(40.00,58.00){\vector(-3,-2){26.00}}
\put(40.00,58.00){\vector(3,-2){28.00}}
\put(40.00,4.00){\makebox(0,0)[cc]{$r$}}
\put(70.00,35.00){\makebox(0,0)[cc]{$q$}}
\put(40.00,13.00){\makebox(0,0)[cc]{2,$r$}}
\put(29.00,21.00){\makebox(0,0)[cc]{1}}
\put(14.00,35.00){\makebox(0,0)[cc]{$p$}}
\put(2.00,35.00){\makebox(0,0)[cc]{3,$r$}}
\put(24.00,40.00){\makebox(0,0)[cc]{5}}
\put(36.00,35.00){\makebox(0,0)[cc]{$t_{1}$}}
\put(23.00,52.00){\makebox(0,0)[cc]{1}}
\put(57.00,52.00){\makebox(0,0)[cc]{1}}
\put(40.00,63.00){\makebox(0,0)[cc]{$s$}}
\put(53.00,62.00){\makebox(0,0)[cc]{10,$r$}}
\put(84.00,35.00){\makebox(0,0)[cc]{5,$q$}}
\put(34.00,54.00){\line(1,0){12.00}}
\put(31.00,31.00){\framebox(10.00,8.00)[cc]{}}
\put(43.00,35.00){\makebox(0,0)[lc]{0,$t_1$}}
\put(19.00,35.00){\vector(1,0){12.00}}
\end{picture}

}     
     \\ \cline{1-4}

 3.    & $r$     &  $t_{1}^{0,t_{1}}$, $t_{2}^{0,t_{2}}$, $q^{5,q}$ (Initial OPEN)  &  \\ \cline{3-3}
    &    &  \fbox{$t_{1}^{0,t_{1}}$}, $t_{2}^{0,t_{2}}$, $q^{5,q}$, $p^{5,r}$  &  \\ \cline{3-3}
     &      &  \fbox{$t_{1}^{0,t_{1}}$}, \fbox{$t_{2}^{0,t_{2}}$}, $q^{5,q}$, $p^{5,r}$, $r^{10,r}$  &  \\ \cline{3-3}
     &      &  \fbox{$t_{1}^{0,t_{1}}$}, \fbox{$t_{2}^{0,t_{2}}$}, \fbox{$q^{5,q}$}, $p^{5,r}$, $r^{10,r}$, $\underline{s}^{6,r}$ &  \\ \cline{3-3}
     &      &  \fbox{$t_{1}^{0,t_{1}}$}, \fbox{$t_{2}^{0,t_{2}}$}, \fbox{$q^{5,q}$}, \fbox{$p^{5,t_{1}}$}, $r^{6,r}$, $s^{12,r}$   & \\ \cline{3-3}
     &      &  \fbox{$t_{1}^{0,t_{1}}$}, \fbox{$t_{2}^{0,t_{2}}$}, \fbox{$q^{5,q}$}, \fbox{$p^{5,t_{1}}$}, \fbox{$r^{6,t_{1}}$}, $s^{12,r}$  & \\ \cline{3-3}
     &      &  \fbox{$t_{1}^{0,t_{1}}$}, \fbox{$t_{2}^{0,t_{2}}$}, \fbox{$q^{5,q}$}, \fbox{$p^{5,t_{1}}$}, \fbox{$r^{6,t_{1}}$}, \fbox{$s^{12,q}$} &  \\
     &      &                                                 &

\raisebox{0pt}[0pt]{    
\linethickness{0.4pt}
\begin{picture}(84.00,95.00)
\put(46.00,-0.33){\line(0,1){0.33}}
\put(40.00,90.00){\circle{10.00}}
\put(14.00,62.00){\circle{10.00}}
\put(70.00,62.00){\circle{10.00}}
\put(40.00,31.00){\circle{10.00}}
\put(16.00,58.00){\vector(3,-4){18.67}}
\put(40.00,85.00){\vector(-3,-2){26.00}}
\put(40.00,85.00){\vector(3,-2){28.00}}
\bezier{236}(37.00,27.00)(7.00,29.00)(11.00,58.00)
\put(11.00,52.00){\vector(0,1){6.00}}
\put(34.00,19.00){\makebox(0,0)[cc]{10}}
\put(40.00,31.00){\makebox(0,0)[cc]{$r$}}
\put(70.00,62.00){\makebox(0,0)[cc]{$q$}}
\put(40.00,40.00){\makebox(0,0)[cc]{6,$t_1$}}
\put(29.00,48.00){\makebox(0,0)[cc]{1}}
\put(13.00,32.00){\makebox(0,0)[cc]{1}}
\put(14.00,62.00){\makebox(0,0)[cc]{$p$}}
\put(2.00,62.00){\makebox(0,0)[cc]{5,$t_1$}}
\put(24.00,67.00){\makebox(0,0)[cc]{5}}
\put(36.00,62.00){\makebox(0,0)[cc]{$t_{1}$}}
\put(23.00,79.00){\makebox(0,0)[cc]{1}}
\put(57.00,79.00){\makebox(0,0)[cc]{1}}
\put(40.00,90.00){\makebox(0,0)[cc]{$s$}}
\put(53.00,89.00){\makebox(0,0)[cc]{12,$q$}}
\put(84.00,62.00){\makebox(0,0)[cc]{5,$q$}}
\put(47.00,6.00){\makebox(0,0)[lc]{0,$t_2$}}
\put(34.00,81.00){\line(1,0){12.00}}
\put(36.00,2.00){\framebox(8.00,8.00)[cc]{$t_{2}$}}
\put(31.00,58.00){\framebox(10.00,8.00)[cc]{}}
\put(43.00,62.00){\makebox(0,0)[lc]{0,$t_1$}}
\put(19.00,62.00){\vector(1,0){12.00}}
\put(40.00,26.00){\vector(0,-1){16.00}}
\end{picture}
}     
     \\ \cline{1-4}

 4.  & $q$     &  $t_{1}^{0,t_{1}}$, $t_{2}^{0,t_{2}}$, $x^{100,x}$ (Initial OPEN) &  \\ \cline{3-3}
  &     &  \fbox{$t_{1}^{0,t_{1}}$}, $t_{2}^{0,t_{2}}$, $x^{100,x}$, $p^{5,t_{1}}$  &  \\ \cline{3-3}
     &      &  \fbox{$t_{1}^{0,t_{1}}$}, \fbox{$t_{2}^{0,t_{2}}$}, $x^{100,x}$, $p^{5,t_{1}}$, $r^{10,t_{1}}$ & \\ \cline{3-3}
     &      &  \fbox{$t_{1}^{0,t_{1}}$}, \fbox{$t_{2}^{0,t_{2}}$}, $x^{100,x}$, \fbox{$p^{5,t_{1}}$}, $r^{6,t_{1}}$, $\underline{s}^{6,q}$  & \\ \cline{3-3}
     &      &  \fbox{$t_{1}^{0,t_{1}}$}, \fbox{$t_{2}^{0,t_{2}}$}, $x^{100,x}$, \fbox{$p^{5,t_{1}}$}, \fbox{$r^{6,t_{1}}$}, \underline{$s^{6,q}$}, $q^{7,q}$ & \\ \cline{3-3}
     &      &  \fbox{$t_{1}^{0,t_{1}}$}, \fbox{$t_{2}^{0,t_{2}}$}, $x^{100,x}$, \fbox{$p^{5,t_{1}}$}, \fbox{$r^{6,t_{1}}$}, $s^{14,q}$, \fbox{$q^{7,t_1}$}  & \\ \cline{3-3}
     &      &  \fbox{$t_{1}^{0,t_{1}}$}, \fbox{$t_{2}^{0,t_{2}}$}, $x^{100,x}$, \fbox{$p^{5,t_{1}}$}, \fbox{$r^{6,t_{1}}$}, \fbox{$s^{14,t_{1}}$}, \fbox{$q^{7,t_1}$}  & \\ 
     &      &      &

\raisebox{0pt}[0pt]{
\linethickness{0.4pt}
\begin{picture}(86.00,95.00)
\put(46.00,-0.33){\line(0,1){0.33}}
\put(40.00,90.00){\circle{10.00}}
\put(14.00,62.00){\circle{10.00}}
\put(70.00,62.00){\circle{10.00}}
\put(40.00,31.00){\circle{10.00}}
\put(70.00,31.00){\circle{10.00}}
\put(16.00,58.00){\vector(3,-4){18.67}}
\put(70.00,57.00){\vector(0,-1){21.00}}
\put(70.00,57.00){\vector(-1,-1){25.00}}
\put(40.00,85.00){\vector(-3,-2){26.00}}
\put(40.00,85.00){\vector(3,-2){28.00}}
\bezier{236}(37.00,27.00)(7.00,29.00)(11.00,58.00)
\put(11.00,52.00){\vector(0,1){6.00}}
\put(34.00,19.00){\makebox(0,0)[cc]{10}}
\put(40.00,31.00){\makebox(0,0)[cc]{$r$}}
\put(70.00,31.00){\makebox(0,0)[cc]{$x$}}
\put(73.00,46.00){\makebox(0,0)[cc]{1}}
\put(86.00,34.00){\makebox(0,0)[cc]{100,$x$}}
\put(70.00,62.00){\makebox(0,0)[cc]{$q$}}
\put(53.00,47.00){\makebox(0,0)[cc]{1}}
\put(40.00,40.00){\makebox(0,0)[cc]{6,$t_1$}}
\put(29.00,48.00){\makebox(0,0)[cc]{1}}
\put(13.00,32.00){\makebox(0,0)[cc]{1}}
\put(14.00,62.00){\makebox(0,0)[cc]{$p$}}
\put(2.00,62.00){\makebox(0,0)[cc]{5,$t_1$}}
\put(24.00,67.00){\makebox(0,0)[cc]{5}}
\put(36.00,62.00){\makebox(0,0)[cc]{$t_{1}$}}
\put(23.00,79.00){\makebox(0,0)[cc]{1}}
\put(57.00,79.00){\makebox(0,0)[cc]{1}}
\put(40.00,90.00){\makebox(0,0)[cc]{$s$}}
\put(53.00,89.00){\makebox(0,0)[cc]{14,$t_1$}}
\put(84.00,62.00){\makebox(0,0)[cc]{7,$t_1$}}
\put(47.00,6.00){\makebox(0,0)[lc]{0,$t_2$}}
\put(34.00,81.00){\line(1,0){12.00}}
\put(36.00,2.00){\framebox(8.00,8.00)[cc]{$t_{2}$}}
\put(31.00,58.00){\framebox(10.00,8.00)[cc]{}}
\put(43.00,62.00){\makebox(0,0)[lc]{0,$t_1$}}
\put(19.00,62.00){\vector(1,0){12.00}}
\put(40.00,26.00){\vector(0,-1){16.00}}
\end{picture}

} 
     \\ \cline{1-4}
\end{tabular}
\caption{Working of S2 on the graph of Figure \ref{fig70}(a)}
\end{centering}
\end{table}

\renewcommand{\arraystretch}{1.6}

\setlength\fboxsep{0.05cm}

\begin{table}
\begin{centering}
\begin{tabular}{||l|l|l|l||}    \hline
$Itn$ & Exp.  &  Nodes in OPEN and CLOSED & Explicit graph after each iteration \\ 
    &  node  & in Bottom\_Up computation & \\
    &   $n$   &  format $m^{h(m),front(m)}$  & \\ \hline
 
 1.    & $s$     &  $p^{5,p}$, $q^{5,q}$ (Initial OPEN) &    \\  \cline{3-3}
     &     &  \fbox{$p^{5,p}$}, $q^{5,q}$, $\underline{s}^{6,s}$  &    \\  \cline{3-3} 
     &      &  \fbox{$p^{5,p}$}, \fbox{$q^{5,q}$}, $s^{12,s}$  & \\ \cline{3-3}
     &      &  \fbox{$p^{5,p}$}, \fbox{$q^{5,q}$}, \fbox{$s^{12,p}$} & 

\raisebox{0pt}[0pt]{
\linethickness{0.4pt}
\begin{picture}(84.00,36.00)
\put(46.00,-0.33){\line(0,1){0.33}}
\put(40.00,31.00){\circle{10.00}}
\put(14.00,3.00){\circle{10.00}}
\put(70.00,3.00){\circle{10.00}}
\put(40.00,26.00){\vector(-3,-2){26.00}}
\put(40.00,26.00){\vector(3,-2){28.00}}
\put(70.00,3.00){\makebox(0,0)[cc]{$q$}}
\put(14.00,3.00){\makebox(0,0)[cc]{$p$}}
\put(2.00,3.00){\makebox(0,0)[cc]{5,$p$}}
\put(23.00,20.00){\makebox(0,0)[cc]{1}}
\put(57.00,20.00){\makebox(0,0)[cc]{1}}
\put(40.00,31.00){\makebox(0,0)[cc]{$s$}}
\put(53.00,30.00){\makebox(0,0)[cc]{12,$p$}}
\put(84.00,3.00){\makebox(0,0)[cc]{5,$q$}}
\put(34.00,22.00){\line(1,0){12.00}}
\end{picture}     
}
     \\ \cline{1-4}

 2.    & $p$     &  $t_{1}^{0,t_{1}}$, $r^{2,r}$, $q^{5,q}$ (Initial OPEN) & \\  \cline{3-3}
    &     &  \fbox{$t_{1}^{0,t_{1}}$}, $r^{2,r}$, $q^{5,q}$, $\underline{p}^{5,p}$ & \\  \cline{3-3}
     &      &  \fbox{$t_{1}^{0,t_{1}}$}, \fbox{$r^{2,r}$}, $q^{5,q}$, $p^{8,p}$  & \\  \cline{3-3}
     &      &  \fbox{$t_{1}^{0,t_{1}}$}, \fbox{$r^{2,r}$}, \fbox{$q^{5,q}$}, $p^{8,p}$, $\underline{s}^{6,p}$  & \\  \cline{3-3}
     &      &  \fbox{$t_{1}^{0,t_{1}}$}, \fbox{$r^{2,r}$}, \fbox{$q^{5,q}$}, \fbox{$p^{8,r}$}, $s^{15,p}$  & \\ \cline{3-3}
     &      &  \fbox{$t_{1}^{0,t_{1}}$}, \fbox{$r^{2,r}$}, \fbox{$q^{5,q}$}, \fbox{$p^{8,r}$}, \fbox{$s^{15,r}$} & 

\raisebox{0pt}[0pt]{
\linethickness{0.4pt}
\begin{picture}(84.00,68.00)
\put(46.00,-0.33){\line(0,1){0.33}}
\put(40.00,63.00){\circle{10.00}}
\put(14.00,35.00){\circle{10.00}}
\put(70.00,35.00){\circle{10.00}}
\put(40.00,4.00){\circle{10.00}}
\put(16.00,31.00){\vector(3,-4){18.67}}
\put(40.00,58.00){\vector(-3,-2){26.00}}
\put(40.00,58.00){\vector(3,-2){28.00}}
\put(40.00,4.00){\makebox(0,0)[cc]{$r$}}
\put(70.00,35.00){\makebox(0,0)[cc]{$q$}}
\put(40.00,13.00){\makebox(0,0)[cc]{2,$r$}}
\put(29.00,21.00){\makebox(0,0)[cc]{1}}
\put(14.00,35.00){\makebox(0,0)[cc]{$p$}}
\put(2.00,35.00){\makebox(0,0)[cc]{8,$r$}}
\put(24.00,40.00){\makebox(0,0)[cc]{5}}
\put(36.00,35.00){\makebox(0,0)[cc]{$t_{1}$}}
\put(23.00,52.00){\makebox(0,0)[cc]{1}}
\put(57.00,52.00){\makebox(0,0)[cc]{1}}
\put(40.00,63.00){\makebox(0,0)[cc]{$s$}}
\put(53.00,62.00){\makebox(0,0)[cc]{15,$r$}}
\put(84.00,35.00){\makebox(0,0)[cc]{5,$q$}}
\put(34.00,54.00){\line(1,0){12.00}}
\put(31.00,31.00){\framebox(10.00,8.00)[cc]{}}
\put(43.00,35.00){\makebox(0,0)[lc]{0,$t_1$}}
\put(19.00,35.00){\vector(1,0){12.00}}
\put(23.00,35.00){\line(-1,-2){4.00}}
\end{picture}
}     
     \\ \cline{1-4}

 3.    & $r$     &  $t_{1}^{0,t_{1}}$, $t_{2}^{0,t_{2}}$, $q^{5,q}$ (Initial OPEN)  &  \\ \cline{3-3}
    &     &  \fbox{$t_{1}^{0,t_{1}}$}, $t_{2}^{0,t_{2}}$, $q^{5,q}$, $\underline{p}^{5,r}$  &  \\ \cline{3-3} 
     &      &  \fbox{$t_{1}^{0,t_{1}}$}, \fbox{$t_{2}^{0,t_{2}}$}, $q^{5,q}$, $\underline{p}^{5,r}$, $\underline{r}^{10,r}$  &  \\ \cline{3-3}
     &      &  \fbox{$t_{1}^{0,t_{1}}$}, \fbox{$t_{2}^{0,t_{2}}$}, \fbox{$q^{5,q}$}, $\underline{p}^{5,r}$, $\underline{r}^{10,r}$, $\underline{s}^{6,r}$ &  \\ \cline{3-3}
     &      &  \fbox{$t_{1}^{0,t_{1}}$}, \fbox{$t_{2}^{0,t_{2}}$}, \fbox{$q^{5,q}$} (all nodes which are not eligible are & \\
     &      &  removed from OPEN and $h(s)$ is set to $\infty$ ) & \\ 
     &      &                                                    &
\raisebox{0pt}[0pt]{
\linethickness{0.4pt}
\begin{picture}(84.00,84.00)
\put(46.00,-0.33){\line(0,1){0.33}}
\put(40.00,79.00){\circle{10.00}}
\put(14.00,56.00){\circle{10.00}}
\put(70.00,56.00){\circle{10.00}}
\put(40.00,25.00){\circle{10.00}}
\put(16.00,52.00){\vector(3,-4){18.67}}
\bezier{236}(37.00,21.00)(7.00,23.00)(11.00,52.00)
\put(11.00,46.00){\vector(0,1){6.00}}
\put(40.00,13.00){\makebox(0,0)[lc]{10}}
\put(40.00,25.00){\makebox(0,0)[cc]{$r$}}
\put(70.00,56.00){\makebox(0,0)[cc]{$q$}}
\put(40.00,34.00){\makebox(0,0)[cc]{10,$r$}}
\put(29.00,42.00){\makebox(0,0)[cc]{1}}
\put(13.00,26.00){\makebox(0,0)[cc]{1}}
\put(14.00,56.00){\makebox(0,0)[cc]{$p$}}
\put(2.00,56.00){\makebox(0,0)[cc]{5,$r$}}
\put(24.00,61.00){\makebox(0,0)[cc]{5}}
\put(36.00,56.00){\makebox(0,0)[cc]{$t_{1}$}}
\put(20.00,68.00){\makebox(0,0)[cc]{1}}
\put(60.00,68.00){\makebox(0,0)[cc]{1}}
\put(40.00,79.00){\makebox(0,0)[cc]{$s$}}
\put(53.00,78.00){\makebox(0,0)[cc]{$\infty$,$r$}}
\put(84.00,56.00){\makebox(0,0)[cc]{5,$q$}}
\put(47.00,2.00){\makebox(0,0)[lc]{0,$t_2$}}
\put(36.00,-2.00){\framebox(8.00,8.00)[cc]{$t_{2}$}}
\put(31.00,52.00){\framebox(10.00,8.00)[cc]{}}
\put(43.00,56.00){\makebox(0,0)[lc]{0,$t_1$}}
\put(19.00,56.00){\vector(1,0){12.00}}
\put(23.00,56.00){\line(-1,-2){4.00}}
\put(31.00,22.00){\line(6,-5){9.00}}
\put(40.00,74.00){\vector(-2,-1){25.00}}
\put(40.00,74.00){\vector(2,-1){27.00}}
\put(32.00,70.00){\line(1,0){16.00}}
\put(40.00,20.00){\vector(0,-1){14.00}}
\end{picture}

}     
     \\  \cline{1-4}
\end{tabular}
\caption{Working of S2 on the graph of Figure \ref{fig70}(b)}
\end{centering}
\end{table}

\subsection{\bf Analysis of S2}
The results on the correctness and complexity of S2 are presented below. In this discussion, 
by a "Bottom-Up computation" we shall mean a call to the Bottom-Up procedure during an iteration of S2.

\begin{defi} \label{admheur} A heuristic function $\hat{h} \geq 0$ defined on the nodes of $G$ is said to
be
{\bf admissible} if $\hat{h}(n) \leq h^{*}(n)$ for all type-I or type-II nodes in $G$.
\end{defi} 

\begin{defi} An execution of S2.2 (i.e.\ substeps S2.2.1, S2.2.2 and S2.2.3) is called
an {\bf iteration} of the algorithm S2.
\end{defi}

\begin{defi} Given any iteration of S2, by an {\bf instant} within it we refer to the time point when substep B2 of step S2.2.3 is about to be executed.
\end{defi}

Thus at instant $j$ of an iteration, the substep B2 is executed for the $j$th time.

\begin{rem} 
\end{rem}
An iteration of S1 contains exactly one instant, while an iteration of S2 will have many instants within it.

\begin{defi}
Let $p$ be a node in $G^{\prime}$, and $M(p)$ be a psg below $p$ in $G^{\prime}$. During any iteration of S2, a node $n \in M(p)$ is a leading node of $M(p)$ at instant $j$ if (i) $n$ is in OPEN and (ii) all its successors in $M(p)$, if any, are in CLOSED at that instant $j$.
\end{defi}

\begin{lem} \label{lem30}
During an iteration of S2, let $M(p)$ be a psg below a  
node $p \in G^{\prime}$. Then at any instant of that iteration, the following must
hold:
  
    (a) If $q$ be a leading node in $M(p)$, then $q$ is eligible in OPEN;
    (b) Conversely, if $p \not\in CLOSED$, there will exist at least one leading node of $M(p)$.

\end{lem}

{\bf Proof.} 
Similar to the proof of Lemma \ref{lem08}.
 \qed\\

\begin{lem} \label{lem40} 
In any iteration of S2, when a node $n$ is sent to CLOSED during
the bottom-up computation, the followings hold:
\begin{enumerate}
  \item $h(n) = h^{\prime}(n)$.
  \item $front(n)$ = a tip node of a minimal-cost psg below $n$.
\end{enumerate} 
\end{lem}

{\bf Proof.}
\begin{enumerate}
\item  By double induction. First on iteration $i$, and then on the nodes 
of the explicit graph which are sent to CLOSED in that iteration.

\underline{\bf Induction Basis.} At iteration $i = 1$, $s$ is expanded. Let $n_1, n_2, \ldots,
n_k$, be those children of $s$
which are tip nodes and are put in OPEN.

Now for each $n_j$, $1 \le j \le k$, $h(n_j)$ equals $0, \infty$ or $\hat{h}(n_j)$
according as $n_j$ is a terminal leaf node, a nonterminal leaf node or an internal
node of $G$. Since each $n_j$ is a tip node, whichever psg they may belong to,
$h(n_j) = h^{\prime}(n_j)$, $1 \le j \le k$ (from the definition of $h^{\prime}$). Whenever any of these nodes travels to CLOSED, it will have $h = h^{\prime}$.

Now we show that the result holds for $s$.

\underline{If $s$ is an AND node}

\underline{Case I: $s$ did not enter CLOSED} Clearly, by step B2 of the algorithm,
$s$ was not eligible. Then, among the children of $s$, at least one did not
enter CLOSED. Since this is the first iteration and $s$ is the only expanded node, this case is possible only if there is a self-loop at $s$,  implying that
$s$ is of type-III. Thus $s$ did not enter CLOSED, and the lemma holds trivially.

\underline{Case II: $s$ entered CLOSED}. Since $s$ is
an AND node, all its children must have been sent to CLOSED prior to $s$ becoming eligible. Now, S2 sends each child of $s$, $n_j$, to CLOSED with $h(n_j) = h^{\prime}(n_j)$ (as
$n_j$ is a tip node). Hence
from the definition of $h^{\prime}$, clearly, $h(s) = h^{\prime}(s)$ when $s$ is sent to CLOSED.

\underline{If $s$ is an OR node}

\underline{Case I: $s$ did not enter CLOSED.} Then the lemma is vacuously true.
(Note that, since $s$ is an OR node, if any of its children had entered OPEN, eventually $s$
would
also have entered OPEN, and then, CLOSED. Then it must be that no child of $s$ even
entered OPEN. This is possible only if $s$ is a type-III node and there is just
a self-loop from $s$.)

\underline{Case II: $s$ entered CLOSED.} Then some child(ren) of $s$, prior to
$s$ itself, must have entered OPEN, and then CLOSED. Let $M$ be a minimal-cost psg
below $s$ in the explicit graph of iteration 1. Let $Z_M = \{n_p\}$. Let us assume
that $s$ entered CLOSED with its $h$-value defined by some child $n_q \not= n_p$
and $h(s) > h^{\prime}(s)$. Then $h(s) > h^{\prime}(s) = c(s,n_p) + h^{\prime}(n_p) >
h^{\prime}(n_p) = h(n_p)$
(since $n_p \in Z_M$). Then clearly, $n_p$ remained in OPEN when $s$ got selected
and sent to CLOSED. This is in contradiction to the criterion of node
selection from OPEN on the basis of minimum $h$-value.

Thus the lemma holds for all nodes entering CLOSED in iteration 1.

\underline{\bf Induction Hypothesis.} Let the lemma be true up to iteration $i = l$.

\underline{\bf Induction Step:} $i = l + 1$. Let the nodes that enter CLOSED be $n_1, n_2,
\ldots $.
Clearly, the first node, $n_1$, that enters CLOSED from OPEN must be a tip node
of the explicit graph at instant $l + 1$, for which $h(n_1) = 0, \infty,$ or $\hat{h}(n_1)$
according as $n_1$ is a terminal leaf, a nonterminal leaf, or an internal node of $G$.
Thus $h(n_1) = h^{\prime}(n_1)$ for the tip node $n_1$. 

Let us assume that the lemma holds up to the $k$th node at instant $l+1$, i.e.\ nodes
$n_1, \ldots n_k$ enter CLOSED with $h(n_j) = h^{\prime}(n_j), 1 \le j \le k$.

We need to show that $n_{k+1}$ goes to CLOSED with
$h(n_{k+1}) = h^{\prime}(n_{k+1})$. 

\underline {Case I:} $n_{k+1}$ is a tip node. Trivially true.

\underline{Case II:} $n_{k+1}$ is an internal node. There can be two subcases within this.

\underline {Case II(a):} $n_{k+1}$ is an AND node. Clearly, $n_{k+1}$ must be eligible and all the children
of
$n_{k+1}$ must have been
previously put into CLOSED, otherwise $n_{k+1}$ could not have become eligible. Hence
$n_{k+1}$'s
children must occur among $n_{1},\ldots ,n_{k}$ and have $h$-values = $h^{\prime}$,
according
to
the induction hypothesis. Now, since S2 computes the $h$-value of an AND node by
successively
adding the $h$-values of its children when each of them is selected from OPEN, it is clear  
that $h(n_{k+1}) = h^{\prime}(n_{k+1})$ for the AND node $n_{k+1}$. 

\underline {Case II(b):} $n_{k+1}$ is an OR node. Let $p$ be the child through which $n_{k+1}$ had
last received its $h$-value prior to its getting selected from OPEN, i.e. $h(n_{k+1}) = c(n_{k+1},p) + h(p)$. 
If $h(n_{k+1}) \not= h^{\prime}(n_{k+1})$, let $q$ be the child of $n_{k+1}$ in a minimal-cost psg $M$ below $n_{k+1}$, 
i.e. $h^{\prime}(n_{k+1}) = c(n_{k+1},q) + h^{\prime}(q)$. 
We shall show that this leads to a contradiction. 
   
Since $q$ has not yet entered CLOSED, the MES $M(n_{k+1},G^{\prime})$ that contains $q$ must have leading nodes, by Lemma \ref{lem30}. Let the leading nodes of $M(n_{k+1},G^{\prime})$ be $q_1, q_2, \ldots, q_m$ ($m \geq 1$). Since $M(n_{k+1},G^{\prime})$ is a minimal-cost MES, we have $h(q_j) = h^{\prime}(q_j)$, $1 \leq j \leq m$.

Thus every $q_{j}, 1 \leq j \leq m,$  has, when $n_{k+1}$ is selected from OPEN, 
\begin{tabbing}
aaaaa\=aaaa\= \kill
$h(q_{j})$ \> = $h^{\prime}(q_{j})$ \\ 
           \> $< h^{\prime}(n_{k+1})$ \\ 
           \> $< c(n_{k+1},p) + h^{\prime}(p)$, as p is $n_{k+1}$'s child \\
           \> = $c(n_{k+1},p) + h(p)$, by induction hypothesis \\
           \> = $h(n_{k+1})$,  as assumed previously
\end{tabbing}
 Therefore, when node $n_{k+1}$ was selected from OPEN, 
   $q_{j}$, being a leading node, was also eligible in OPEN with $h(q_{j}) < h(n_{k+1})$.
   This is clearly in contradiction to the best-first node selection criterion,
   on the basis of minimum $h$, used by S2. Hence the result.

\item As in (1), the proof is by double induction, first on the iteration and then on the nodes of $G^{\prime}$ entering CLOSED in that iteration. We simply give an outline of the proof below.

If a tip node $n$ enters CLOSED, $n$ has itself as its front, and the lemma
is trivially true, as $n$ is the only node in a minimal-cost psg below it.

When an internal node $n$ goes to CLOSED, $h(n)$ is set to $h^{\prime}(n)$,
(the cost of a minimal-cost psg below $n$), as proved in part (1) above. Now, when $n$ enters CLOSED, its $front$ is decided in step B2.2. If $n$ is an OR node, $front(n)$ is set to the $front$ of one of its children which, in turn, defines $h^{\prime}(n)$. Note that $h^{\prime}(n)$ is the cost of a minimal-cost psg below $n$. Thus $front(n)$ becomes a tip node of a minimal-cost psg below $n$. If $n$ is an AND node, its $h^{\prime}(n)$-value is computed by adding the $h^{\prime}$-values of all the children and the costs of the arcs connecting them with $n$. Now, $front(n)$ is set to the $front$ of one of the children of $n$, which in turn, is a tip node of a minimal-cost psg below it. \qed\\
\end{enumerate}

\begin{lem} \label{lem50} If $s$ is not of type-III, then at the end of every Bottom-up
computation,
it must enter CLOSED.
\end{lem}

{\bf Proof.} In any iteration, given the explicit graph, if $s$ is not of type-III, there will exist psgs below $s$ in $G^{\prime}$.
Let $M_1, M_2, \ldots, M_k$ be the all possible psgs rooted at $s$. Let
$M_j, 1 \le j \le k$, be a minimal-cost psg below $s$ with cost $h^{\prime}(s)$.
Now it is easy to show that eventually $s$ must enter CLOSED.
At each instant, (i.e. execution of step B2 of Bottom-Up computation), a distinct node is selected from OPEN and put into CLOSED. Once a node enters CLOSED, it never returns to OPEN. Since in any iteration, there are only finitely many nodes in $G^{\prime}$, it is clear that the Bottom-Up computation cannot continue indefinitely. On the other hand, the Bottom-Up computation cannot get stuck since there must exist a leading node from $M_j$ which is eligible (Lemma \ref{lem30}), prior to sending $s$ to CLOSED. Thus it is clear that,  
after finitely many instants
of a Bottom-Up computation, $s$ will become the leading node of $M_j$ and will
eventually be sent from OPEN to CLOSED.\qed\\

\begin{lem} \label{lem60} Under admissible heuristics, at the end of Bottom-Up
computation of every iteration of S2,
we have $h^{\prime}(n) \le h^{*}(n)$, where $n$ is a type-I or type-II node in $G$ and is currently included in $G^{\prime}$. 
\end{lem}

{\bf Proof.} Let $M$ be a minimal-cost MES below $n$ in $G$, and let 
$M^{\prime}$ be the portion of $M$ contained in $G^{\prime}$. Thus, $M^{\prime}$ is a psg below $n$ in $G^{\prime}$.
Now clearly the following observations will hold:

({\em a}) From the definition of $h^{\prime}$, $h^{\prime}(n) \le \beta(n,M^{\prime})$, 
since $M^{\prime}$ is just one of the psgs below $n$ and $M^{\prime}$ need not
define $h^{\prime}(n)$, {\em and} 
({\em b}) $\beta(n,M^{\prime}) \le \beta(n,M) = h^{*}(n)$, by the definition of
$\beta$ (Definition \ref{betaexpl}) and the admissibility of the heuristic function (Definition \ref{admheur}).

Now combining ({\em a}) and ({\em b}), the lemma follows. \qed\\

\begin{lem} \label{lem70} In any iteration of S2, during the Bottom\_Up computation, no type-III
node in $G^{\prime}$ ever enters CLOSED.
\end{lem}

{\bf Proof.} The proof is similar to that of Lemma \ref{lem10} for S1. Note that the Sub-problem Composition Theorem for Explicit Graphs (i.e. Theorem \ref{thm25}) is used in place of the Sub-problem Composition Theorem for Implicit Graphs (Theorem \ref{thm20}).  \qed\\

\begin{th}\label{thm50} If the implicit graph $G$ has at least one solution graph, then S2 running with admissible heuristics terminates with SUCCESS by outputting $h(s) = h^{*}(s)$.
\end{th}

{\bf Proof.} Let $G$ be any AND/OR graph containing at least one solution graph.
Since node branching factor is finite, there are only finitely many psgs $M$ below $s$
which are subgraphs of $G$, having costs $\beta(s,M) \le h^*(s)$. Now, combining Lemmas
\ref{lem40}
and \ref{lem60}, $h(s) = h^{\prime}(s) \le h^*(s)$ ($s$ being a type-I node, as it has a
solution graph below it) at the end of each iteration of S2. Now by Lemma \ref{lem40}, at the
end of each iteration, $front(s)$ is set to a tip node of a minimal-cost psg below
$s$, and this $front(s)$ is expanded in the next iteration. Since every $front(s)$ is unique and arc-costs are positive, after finitely many iterations there will be no more psgs  having costs $\leq h^*(s)$. Thus unless S2 terminates, it has to continue
the search with psgs having costs $> h^*(s)$, which is in contradiction to the
Lemma \ref{lem60}.
Thus S2 must terminate after finitely many iterations.

Again, as $s$ is of type-I, $h^*(s) < \infty$, implying that at the end of each iteration
$h(s) < \infty$. Hence S2 cannot terminate with FAILURE.

Hence the alternative termination condition namely, $front(s)$ is a terminal leaf,
must hold. Let $M$ be the psg below $s$ that sets $front(s)$ to be a terminal leaf.
Then $Z_M$ cannot contain a non-leaf tip node (otherwise $front(s)$ could not be a
terminal leaf, from the step B2.2 of S2). Now $Z_M$ cannot contain any nonterminal
leaf either, as it would otherwise violate the fact that $s$ is a type-I node and $front(s)$
is a tip node of a minimal-cost psg below $s$.

Therefore, $M$ must be a solution graph, and $h(s) = \beta(s,M) \ge h^*(s)$.

Combining this with $h(s) = h^{\prime}(s) \leq h^{*}(s)$ at the end of every iteration (Lemma \ref{lem40} and Lemma \ref{lem60}), we have $h(s) = h^*(s)$ at termination of S2. \qed\\

\begin{th}\label{thm60} S2 terminates with FAILURE on a finite AND/OR graph $G$
that does not contain a solution graph.
\end{th}

{\bf Proof.} Let $G$ be a finite AND/OR graph that does not contain a solution graph. While S2 runs on $G$, in its every iteration a node called $front(s)$ is expanded. But in every iteration, $front(s)$ is a distinct node, i.e. a tip node of a minimal-cost psg below $s$. Since $G$ is finite, S2 can run for finitely many iterations. 

The termination of S2 can happen either when $front(s)$ is a terminal leaf or when $h(s) = \infty$. Now, if $front(s)$ is a terminal leaf, S2 must have obtained a solution graph below $s$. This contradicts the fact that $G$ does not contain a solution graph. Hence, the other condition for termination namely, $h(s) = \infty$ must hold, implying that S2 terminates with FAILURE. (This FAILURE termination can happen in either of two ways. $s$ could be a type-II node in $G$, in which case $h(s)$ is set to $\infty$ in step B2.3. Alternatively, $s$ could be a type-III node in $G$, in which case it does not travel to CLOSED at the end of a Bottom-Up computation and has its $h$-value set to $\infty$ at step B4. In either case S2 terminates with FAILURE.) \qed\\

\subsection{Complexity Analysis of S2}

In this section, we present the complexity analysis of S2. 

\begin{defi} \label{defcompl}

(i) Let $G$ be an AND/OR graph. We define a set of nodes $V$ as follows:
\begin{enumerate} 

\item If $G$ has at least one solution graph, then:

\begin{enumerate}
\item $s$ is in $V$ if $s$ is not a terminal or nonterminal leaf node;
\item a node $n$ is in $V$ if $n$ is not a terminal leaf node,
\underline{and} if there exists a psg $M$ below $s$
in some explicit graph $G^{\prime}$ for $G$ such that $n$ is a tip node
in $M$ and $\beta(s,M) \leq h^{*}(s)$.
\end{enumerate}

\item If $G$ has no solution graph, but it is finite, then every non-leaf node $n \in G$ will belong to $V$.
\end{enumerate}

(ii) Let $N_2 = \mid V \mid$.

\end{defi}

\begin{th}\label{thm70} Let $G$ be an implicitly defined AND/OR graph, such that: (a) $G$
contains at least one solution graph or (b) $G$ is finite. Now 
when S2 runs on $G$ with an admissible heuristic, the followings are true:
\begin{enumerate}
    \item S2 requires $O(K_2) = O(N_2)$ storage;
    \item S2 makes at most $N_2+1$ iterations;
    \item S2 runs in  $O(N_2K_2^{2}) = O(N_2^{3})$ time;  
\end{enumerate}
where $N_2$ is as in Definition \ref{defcompl}, and $K_2$ is the total number of nodes in the explicit graph $G^{\prime}$ in the last iteration of
S2 ($K_2 \le b*N_2$, where $b$ is the maximum branching factor of a node, which is
finite).
\end{th}

{\bf Proof.} 
\begin{enumerate}
\item S2 stores the entire explicit graph having $K_2$ nodes, with $O(b)$
information at each node. As $b$ is finite, the result follows.

\item 

(a) If $s$ is a terminal leaf node, S2 will terminate in the first iteration. 
Otherwise, S2 continues the search by expanding $front(s)$ in every iteration. Thus, as long as S2 does not terminate, it must be the case 
that $front(s)$ is neither a terminal node, nor is it a nonterminal node. Moreover, $front(s)$ will be assigned to a distinct node in every iteration of S2.

Thus at each iteration prior to termination, S2 expands a distinct node, say $n$.
Now, from Lemma \ref{lem40}(2), $n$ is a tip node of a minimal-cost psg  $M$ below $s$.
Such a psg will have cost $\beta(s,M) = h^{\prime}(s) \le h^{*}(s)$ (Lemma \ref{lem60}).
Thus the expanded node $n$ (= $front(s)$) will be one of the nodes of $V$. Thus prior to termination, S2 makes at most $N_2$ node expansions in as many iterations, plus one more iteration for termination.

(b) If $s$ is a nonterminal leaf node, S2 will terminate in the first iteration. Otherwise, it continues the search by expanding a distinct node as assigned to $front(s)$ in every iteration. Since there are $N_2$ internal nodes in $G$, S2 can make at most $N_2$ expansions in $N_2$ iterations before terminating in the ($N_2 + 1$)th iteration.

\item S2 makes $N_2+1$ iterations, from (2) above. In each iteration (prior to the terminating one),
it does the
followings:
 \begin{enumerate}
 \item (Step S2.2) \underline{Checks} for termination, in $O(1)$ time;
 \item (Step S2.2.1) \underline{Expands} a node and generates its children, in $O(b)$ time
(where
 $b$ is the maximum branching factor of a node, which is finite);
 \item (Step S2.2.2) \underline{Creates} OPEN, in $O(K_2)$ time, since there are $K_2$ 
 nodes in $G^{\prime}$;
 \item (Step S2.2.3) \underline{Calls} Bottom-Up; in each call, it:
     \begin{enumerate}
     \item (Step B1) \underline{Creates} CLOSED, in $O(1)$ time;
     \item (Step B2) \underline{Checks} conditions, in $O(1)$ time, and makes $O(K_2)$
     iterations of the following steps:
        \begin{enumerate}
          \item (Step B2.1) \underline{Selects} an eligible node from OPEN, in $O(K_2)$ time;
          \item (Step B2.2) \underline{Decides} front, in $O(b)$ time;
          \item (Step B2.3) \underline{Evaluates} the parents of the selected node, in $O(K_2)$ time;
        \end{enumerate}
     \item (Step B3) \underline{Clears} OPEN, in $O(1)$ time;
     \item (Step B4) \underline{Checks} whether $s \in CLOSED$, in $O(1)$ time.
     \end{enumerate}
  \end{enumerate}

Thus the overall time complexity of S2 is 

$O(N_2(1+b+K_2+(1+1+K_2(K_2+b+K_2)+1+1)))$ = $O(N_2K_2^{2}) =O(b^{2}N_2^{3}) = O(N_2^{3})$.
\end{enumerate} \qed\\

\section{Experimental Results}

The algorithms S1, $\mbox{REV}^{*}$, S2, $CFC_{REV^*}$ and $\mbox{AO}^{*}$ have been empirically compared on a DEC-Alpha workstation. All algorithms except $CFC_{REV^*}$ were programmed in C++. For $CFC_{REV^*}$, the C-code was obtained from the website mentioned in [Jim$\acute{e}$nez and Torras 2000]. The experimental results obtained are now discussed.

First the AND/OR graphs (problem instances) are generated using the following parameters: the total number of nodes in the implicit graph, the percentage of AND nodes, and whether the graph is cyclic or acyclic. For every combination of these parameters, the relevant algorithms are run over a set of one hundred randomly generated graphs. The average time of execution and the average number of nodes evaluated in each case (i.e. over 100 problems) are noted in tables 4 and 5. The time is mentioned in CPU Clock Ticks, where 1 Clock Tick = $10^{-6}$ sec. The number of node evaluations is the number of times the cost of a node is computed during bottom-up phase (one bottom-up phase for S1 and $\mbox{REV}^{*}$, multiple bottom-up phases for S2, $CFC_{REV^*}$ and $\mbox{AO}^{*}$.) The tables 4 and 5 show a snapshot of the experimental results. As it was found that variations in heuristic estimate or node branching factor do not reveal any new information, the heuristic estimate was kept at 90\% to 100\% of the solution cost for each node and the node branching factor was kept fixed at 3. 

One characteristic of the graphs we used in our experiments is that, the start node is always type-I - i.e. it is not of type-II or type-III. The reason for choosing this is that if the start node is of type-II or type-III, it is assigned a very high heuristic value by our heuristic computation program. In that case, heuristic search algorithms like S2 or $CFC_{REV^*}$ find the $h$-value of start node to be $\infty$ (in the first iteration itself) and  exit from the problem almost immediately, making a fair comparison with S1 or 
$\mbox{REV}^{*}$ impossible. For this reason, the graphs that do not have a solution graph below $s$ are discarded from our set of hundred problems in each case.

\begin{table}
\begin{centering}
\begin{tabular}{|c|c|c|c|c|c|c|}    \hline
\%  & Nodes & S1 &  $\mbox{REV}^{*}$ & S2 & $CFC_{REV^*}$ & $\mbox{AO}^{*}$   \\ \cline{3-7}
AND  & in $G$  & Time & Time & Time & Time & Time  \\ 
& & Node & Node & Node & Node & Node \\ \hline
 
30   & 1000 & 20     & 14       & 2      & 1       & 5 \\
     &      & 1418   & 3496     &1952    & 580     & 284 \\ \cline{2-7}
     & 2000 & 70     & 60       & 7      & 3       & 12  \\
     &      & 2865   & 7297     & 4689   & 1202    & 795 \\ \cline{2-7} 
     & 3000 & 148    & 139      & 14     & 6       & 20 \\ 
     &      & 4305   & 11170    & 7418   & 1771    & 1324 \\ \hline
50   & 1000 & 20     & 14       & 376    & 88      & 32 \\
     &      & 1632   & 3489     & 61711  & 6698    & 3848 \\ \cline{2-7}
     & 2000 & 66     & 60       & 1023   & 374     & 105 \\
     &      & 3108   & 7297     & 113778 & 18779   & 9023 \\ \cline{2-7}
     & 3000 & 137    & 140      & 1296   & 698     & 180 \\ 
     &      & 4496   & 11169    & 134432 & 28185   & 13804 \\ \hline     
 
\end{tabular}
\caption{Performance of Algorithms on Acyclic Graphs}
\end{centering}
\end{table}

\begin{table}
\begin{centering}
\begin{tabular}{|c|c|c|c|c|c|}    \hline
\%  & Nodes & S1 &  $\mbox{REV}^{*}$ & S2 & $CFC_{REV^*}$  \\ \cline{3-6}
AND  & in $G$  & Time & Time & Time & Time  \\ 
& & Node & Node & Node & Node \\ \hline
 
30   & 1000 & 21    & 14     & 2     & 1 \\
     &      & 1415  & 3312   & 2052  & 849 \\ \cline{2-6}
     & 2000 & 73    & 62     & 9     & 3 \\
     &      & 2919  & 6949   & 4946  & 1576  \\ \cline{2-6} 
     & 3000 & 157   & 144    & 20    & 7 \\ 
     &      & 4460  & 10732  & 8602  & 2461 \\ \hline
50   & 1000 & 17    & 9      & 12    & 7 \\
     &      & 1115  & 2110   & 5811  & 2365  \\ \cline{2-6}
     & 2000 & 54    & 35     & 25    & 17  \\
     &      & 2215  & 4110   & 10319 & 5022  \\ \cline{2-6}
     & 3000 & 112   & 82     & 42    & 35 \\ 
     &      & 3326  & 6296   & 13912 & 7497 \\ \hline     
 
\end{tabular}
\caption{Performance of Algorithms on Cyclic Graphs}
\end{centering}
\end{table}

From the tables, the following observations can be made:

\begin{enumerate}
\item the smallest execution time is taken by $\mbox{REV}^{*}$ among uninformed algorithms and by $CFC_{REV^*}$ among heuristic search algorithms;

\item S1 makes less node evaluations than $\mbox{REV}^{*}$. This is expected, as S1 is designed to work in a best-first manner. But the same effect is not observed between S2 and $CFC_{REV^*}$ - actually, S2 makes more node evaluations than $CFC_{REV^*}$. While this may appear to be contradictory, the reason is that $CFC_{REV^*}$ operates on a much smaller size of OPEN (compared to S2) in each iteration. Ultimately, this effect dominates over the best-first nature of S2. It is interesting that $\mbox{AO}^{*}$ makes even less node evaluations (than $CFC_{REV^*}$) on acyclic graphs. This is because, $\mbox{AO}^{*}$ employs the best-first principle along with a smaller set of initial nodes (in Z-list) and thereby gains in node evaluations over both S2 and $CFC_{REV^*}$. 

\item In spite of evaluating less nodes than $CFC_{REV^*}$, $\mbox{AO}^{*}$ takes more time to execute. This is because, the time taken for predecessor-checking during the bottom-up computation in $\mbox{AO}^{*}$ is quite high and dominates over the time for node evaluations;

\item S2 makes more node evaluations compared to $CFC_{REV^*}$, but takes much less time per node than $CFC_{REV^*}$. Thus S2 does less work per node compared to $CFC_{REV^*}$, which is also clear from the design of the algorithms.

\item It was observed that under identical tie-resolution strategy, S2, $\mbox{AO}^{*}$ and $CFC_{REV^*}$ expand the same number of nodes while running on acyclic graphs. The same is true about S2 and $CFC_{REV^*}$ on cyclic graphs. The data has been omitted from this presentation.

\end{enumerate}

\section{Summary and Future Work}

Best-first search in cyclic AND/OR graphs had been a long-unresolved problem
of artificial intelligence. Over the last few years, a number of studies have been reported on this topic. However, all of these studies lacked an unified theoretical framework for both cyclic and acyclic AND/OR graphs, and this led to a lacuna in the theoretical proofs of those algorithms. In this paper, we have taken a fresh look at the problem. First, a new and comprehensive framework for cyclic AND/OR graphs has been
presented, which should be of use to future researchers as well. Then two best-first algorithms, S1 and S2, have been developed for searching AND/OR graphs in the presence of cycles. The new theoretical framework has been useful in establishing the correctness and complexity results of S1 and S2 in detail.

S1 and S2 have been implemented on a DEC-Alpha
Workstation, and a large number of experiments on
randomly-generated graphs have yielded correct results in all cases. 
However, computational times show that the execution time of S2 is not favourable compared to that of $CFC_{REV^*}$ (on the same set of random graphs and heuristic distribution). Again, the node evaluations of $CFC_{REV^*}$ is not favourable compared to that of $\mbox{AO}^{*}$ (on acyclic graphs). Clearly, the design of an algorithm that has the best-first nature of S2, the time performance of $CFC_{REV^*}$ and the node performance of $\mbox{AO}^{*}$ remains a research challenge for the future.

\section[*]{Acknowledgements}

The authors would like to express their acknowledgements to a number of individuals. P. P. Chakrabarti has been very helpful in discussing his work on the topic while on a trip to the authors' Institute. D. Hvalica has corresponded with the authors and has sent his earlier work in this topic, which was a great help. P. Jim$\acute{e}$nez and C. Torras have provided the code of their $CFC_{REV^*}$ algorithm and also explained the method of running it, which was very useful in the experimental part of the work.

\newpage

\section{Appendix}

The problem of cyclic AND/OR graph search has attracted considerable research attention
in recent times [Chakrabarti 1994; Hvalica 1996; Jim$\acute{e}$nez and Torras 2000]. However all these attempts
have overlooked certain important issues, which underline the fundamental nature of the problem.
We briefly review these recent work in this section.

\subsection{Algorithm $\mbox{REV}^{*}$ }

[Chakrabarti 1994] has suggested a definitional framework and two algorithms,
Iterative\_Revise and $\mbox{REV}^{*}$. 

\begin{center}
\underline{\bf Scenario 1. Basic structure collapses}
\end{center}

Nilsson's statement 
{\em "This recursive definition is satisfactory because we are assuming acyclic
graphs"} [Nilsson 1980, pp. 102]  was a warning in disguise that for cyclic AND/OR graphs, it may be difficult
to have a formalism based on recursive structures. [Chakrabarti 1994] defined  the basic structure $opt(A,n)$ (pp. 331)
which enters into an infinite recursion.
[Chakrabarti 1994] used the following notation:

$A$: Implicit AND/OR graph

$opt(A,n)$: Cost of a minimal-cost solution graph below node $n$ in $A$

$t(n)$: Non-negative cost at terminal node $n$

$D(n)$: Solution graph below $n$

We now reproduce the definition of $opt(A,n)$ from [Chakrabarti 1994]:

For every AND/OR graph $A$, the quantity $opt(A,n)$ is defined recursively as
follows:

\begin{tabbing}
aaaaaaaaaa\=aaa\= \kill
$opt(A,n)$ \> = $t(n)$, if $n$ is a terminal node in $A$; \\
           \> = $\infty$, if there does not exist any  solution graph $D(n)$ in $A$; \\
           \> = $min_{1 \le i \le k} \{opt(A,n_i) + c(n,n_i)\}$,  for OR node $n$ with immediate \\
           \> \> successors $n_i$, $1 \le i \le k$; \\
           \> = $\sum_{1 \le i \le k} \{opt(A,n_i) + c(n,n_i)\}$, for AND node $n$ with immediate \\
           \> \> successors $n_i$, $1 \le i \le k$. 
\end{tabbing}

\vspace{5pt}

We now illustrate the severe drawback in this definition on the implicit graphs
$A$ of Figure \ref{fig110}.

\linethickness{0.4pt}
\unitlength=0.5mm
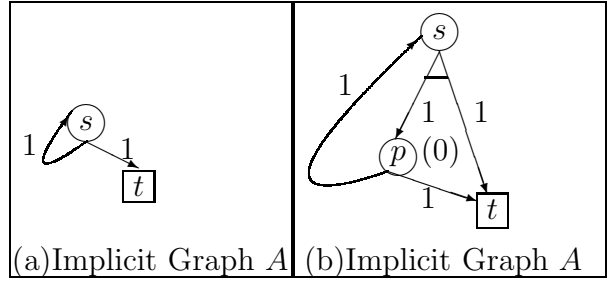
\begin{figure}
\centering
\caption{$opt(A,s)$ collapses}
\label{fig110}
\linethickness{0.4pt}
\begin{picture}(158.00,73.00)
\put(20.00,41.00){\makebox(0,0)[cc]{$s$}}
\put(20.00,41.00){\circle{10.00}}
\bezier{216}(20.00,36.00)(-1.00,21.00)(16.00,44.00)
\put(12.00,38.00){\vector(2,3){3.33}}
\put(37.00,4.00){\makebox(0,0)[cc]{(a)Implicit Graph $A$}}
\put(20.00,36.00){\vector(2,-1){14.00}}
\put(114.00,65.00){\makebox(0,0)[cc]{$s$}}
\put(114.00,65.00){\circle{10.00}}
\put(128.00,18.00){\makebox(0,0)[cc]{$t$}}
\put(34.00,24.00){\makebox(0,0)[cc]{$t$}}
\put(29.00,34.00){\makebox(0,0)[lc]{1}}
\put(7.00,35.00){\makebox(0,0)[rc]{1}}
\put(123.00,44.00){\makebox(0,0)[lc]{1}}
\put(103.00,32.00){\circle{10.00}}
\put(30.00,20.00){\framebox(8.00,8.00)[cc]{}}
\put(124.00,13.00){\framebox(8.00,9.00)[cc]{}}
\put(103.00,32.00){\makebox(0,0)[cc]{$p$}}
\put(109.00,31.00){\makebox(0,0)[lb]{(0)}}
\put(109.00,44.00){\makebox(0,0)[lc]{1}}
\put(100.00,28.00){\vector(3,-1){24.00}}
\put(91.00,51.00){\makebox(0,0)[rc]{1}}
\bezier{496}(100.00,28.00)(54.00,12.00)(109.00,64.00)
\put(114.00,60.00){\vector(-1,-2){11.67}}
\put(114.00,60.00){\vector(1,-3){12.67}}
\put(106.00,61.00){\vector(1,1){2.00}}
\put(115.00,4.00){\makebox(0,0)[cc]{(b)Implicit Graph $A$}}
\put(111.00,21.00){\makebox(0,0)[cc]{1}}
\put(110.00,53.00){\line(1,0){6.00}}
\put(75.00,0.00){\framebox(83.00,73.00)[cc]{}}
\put(0.00,0.00){\framebox(75.00,73.00)[cc]{}}
\end{picture}
\end{figure}

On the OR graph of Figure \ref{fig110}(a), $opt(A,s) = min \{opt(A,s) + c(s,s), opt(A,t) + c(s,t)\}$,
which leads to an infinite recursion. Similarly on the
AND/OR graph of Figure \ref{fig110}(b), $opt(A,s) = [\{opt(A,p) + c(s,p)\} + \{opt(A,t) + c(s,t)\}]
= [\{min \{opt(A,s) + c(p,s), opt(A,t) + c(p,t)\} + c(s,p)\} +
\{opt(A,t) + c(s,t)\}]$, which again leads to an infinite recursion.

\begin{center}
\underline{\bf Scenario 2. Best-first principle violated}
\end{center}

We now turn to the algorithms presented in [Chakrabarti 1994].
Of the two algorithms, $\mbox{REV}^{*}$ is an improved version of
the depth first algorithm Iterative\_Revise. $\mbox{REV}^{*}$
performs bottom-up search by treating the implicit graph in an explicit fashion.
Starting the search from the leaf nodes, $\mbox{REV}^{*}$ applies inverse
operators, using a list OPEN on the way.

However, when $\mbox{REV}^{*}$ selects a node $n$ from OPEN, it immediately tries to select and 
evaluate its predecessor nodes, without letting these predecessors to enter OPEN. {\em This
causes serious violations of the best-first criterion} and as a result it ends up 
selecting nodes with higher (or $\infty$) costs than the minimum cost of a node
in OPEN. This paradoxical phenomenon is clearly portrayed in the two examples
given in Figure \ref{fig120}.

\unitlength=0.45mm
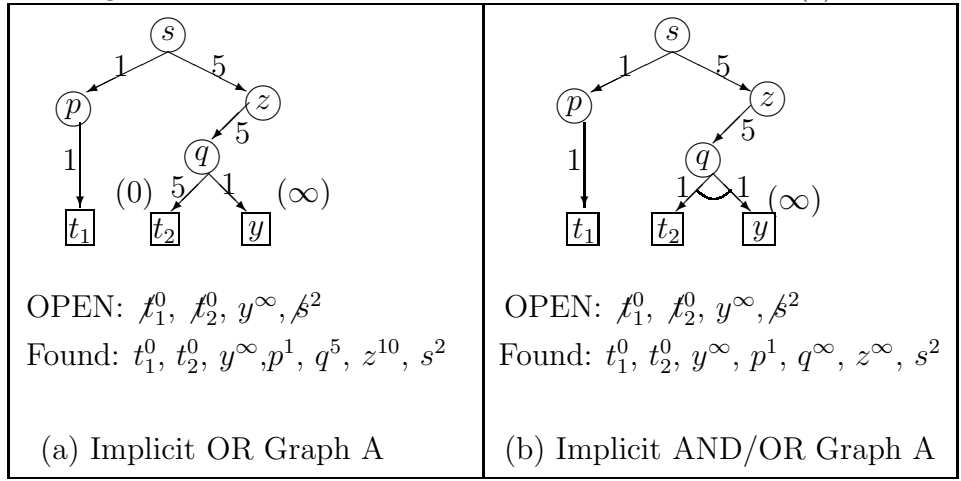
\begin{figure}
\centering
\caption{$\mbox{REV}^{*}$ selects nodes with $\infty$ cost, while $h^{*}(s) = 2$}
\label{fig120}
\linethickness{0.4pt}
\begin{picture}(280.00,140.00)
\put(19.00,109.00){\makebox(0,0)[cc]{$p$}}
\put(18.00,94.00){\makebox(0,0)[cc]{1}}
\put(21.00,73.00){\makebox(0,0)[cc]{$t_{1}$}}
\put(46.00,73.00){\makebox(0,0)[cc]{$t_{2}$}}
\put(50.00,86.00){\makebox(0,0)[cc]{5}}
\put(37.00,84.00){\makebox(0,0)[cc]{(0)}}
\put(73.00,73.00){\makebox(0,0)[cc]{$y$}}
\put(87.00,84.00){\makebox(0,0)[cc]{($\infty$)}}
\put(65.00,87.00){\makebox(0,0)[cc]{1}}
\put(57.00,95.00){\makebox(0,0)[cc]{$q$}}
\put(69.00,102.00){\makebox(0,0)[cc]{5}}
\put(75.00,111.00){\makebox(0,0)[cc]{$z$}}
\put(47.00,126.00){\vector(2,-1){23.00}}
\put(62.00,122.00){\makebox(0,0)[cc]{5}}
\put(33.00,122.00){\makebox(0,0)[cc]{1}}
\put(47.00,131.50){\makebox(0,0)[cc]{$s$}}
\put(21.00,105.00){\vector(0,-1){25.00}}
\put(59.00,90.00){\vector(-1,-1){11.00}}
\put(71.00,111.00){\vector(-1,-1){11.00}}
\put(47.00,126.00){\vector(-2,-1){24.00}}
\put(59.00,90.00){\vector(1,-1){11.00}}
\put(167.00,110.00){\makebox(0,0)[cc]{$p$}}
\put(167.00,94.00){\makebox(0,0)[cc]{1}}
\put(170.00,73.00){\makebox(0,0)[cc]{$t_{1}$}}
\put(195.00,73.00){\makebox(0,0)[cc]{$t_{2}$}}
\put(199.00,86.00){\makebox(0,0)[cc]{1}}
\put(222.00,73.00){\makebox(0,0)[cc]{$y$}}
\put(232.00,82.00){\makebox(0,0)[cc]{($\infty$)}}
\put(217.00,86.00){\makebox(0,0)[cc]{1}}
\put(196.00,126.00){\vector(2,-1){23.00}}
\put(211.00,122.00){\makebox(0,0)[cc]{5}}
\put(182.00,122.00){\makebox(0,0)[cc]{1}}
\put(196.00,131.50){\makebox(0,0)[cc]{$s$}}
\put(170.00,105.00){\vector(0,-1){25.00}}
\put(208.00,90.00){\vector(-1,-1){11.00}}
\put(219.00,110.00){\vector(-1,-1){11.00}}
\put(196.00,126.00){\vector(-2,-1){24.00}}
\put(208.00,90.00){\vector(1,-1){11.00}}
\bezier{56}(203.00,85.00)(208.00,80.00)(213.00,85.00)
\put(47.00,131.00){\circle{10.00}}
\put(75.00,111.00){\circle{10.00}}
\put(57.00,95.00){\circle{10.00}}
\put(19.00,109.00){\circle{10.00}}
\put(196.00,131.00){\circle{10.00}}
\put(224.00,112.00){\circle{10.00}}
\put(205.00,94.00){\circle{10.00}}
\put(167.00,110.00){\circle{10.00}}
\put(224.00,112.00){\makebox(0,0)[cc]{$z$}}
\put(216.00,103.00){\makebox(0,0)[lc]{5}}
\put(205.00,94.00){\makebox(0,0)[cc]{$q$}}
\put(5.00,50.00){\makebox(0,0)[lc]{OPEN: $\not t_{1}^{0}$, $\not t_{2}^{0}$, $y^{\infty}$,$\not s^{2}$}}
\put(17.00,69.00){\framebox(8.00,10.00)[cc]{}}
\put(42.00,69.00){\framebox(8.00,9.00)[cc]{}}
\put(69.00,69.00){\framebox(8.00,9.00)[cc]{}}
\put(165.00,69.00){\framebox(9.00,9.00)[cc]{}}
\put(190.00,69.00){\framebox(9.00,9.00)[cc]{}}
\put(217.00,69.00){\framebox(9.00,9.00)[cc]{}}
\put(190.00,50.00){\makebox(0,0)[cc]{OPEN: $\not t_{1}^{0}$, $\not t_{2}^{0}$, $y^{\infty}$,$\not s^{2}$}}
\put(5.00,36.00){\makebox(0,0)[lc]{Found: $t_{1}^{0}$, $t_{2}^{0}$, $y^{\infty}$,$p^{1}$, $q^{5}$, $z^{10}$, $s^{2}$}}
\put(210.00,36.00){\makebox(0,0)[cc]{Found: $t_{1}^{0}$, $t_{2}^{0}$, $y^{\infty}$, $p^{1}$, $q^{\infty}$, $z^{\infty}$, $s^2$}}
\put(60.00,8.00){\makebox(0,0)[cc]{(a) Implicit OR Graph A}}
\put(210.00,8.00){\makebox(0,0)[cc]{(b) Implicit AND/OR Graph A}}
\put(140.00,0.00){\framebox(140.00,140.00)[cc]{}}
\put(0.00,0.00){\framebox(140.00,140.00)[cc]{}}
\end{picture}
\end{figure}

Figure \ref{fig120}(a) presents a simple OR graph, for which a best-first search algorithm like
Dijkstra's will never visit nodes $q$ with cost 5, and $z$ with cost 10. But $\mbox{REV}^{*}$
works as follows: initially it puts in OPEN the "found" nodes $t_{1}$, $t_{2}$ and $y$ with
costs 0,0 and
$\infty$ respectively.
Then, after selecting $t_{1}$, it continues its upward computation through the "found" node $p$
up to $s$, and
inserts $s$ with cost 2 (i.e. the cost of a minimal-cost solution graph) in OPEN. Next the node
$t_{2}$ is selected. Now, although $s$
awaits in OPEN with cost 2 for selection, $\mbox{REV}^{*}$ proceeds upwards from
$t_{2}$, selecting "found" nodes $q$ and $z$, with costs 5 and 10, ignoring the legitimate 
superior  candidacy of $s$ in OPEN.  

Figure \ref{fig120}(b) depicts a similar situation in presence of an AND node $q$. Here,
nodes $q$ and $z$, with $\infty$ costs, will get preference over $s$ with cost 2
in OPEN.

\begin{center}
\underline{\bf Scenario 3. Basic theorem fails}
\end{center}

The most severe fallout of $\mbox{REV}^{*}$'s violation of the best-first
principle is that, the {\bf Theorem 5.3(iii) of [Chakrabarti 1994] fails},
where it was claimed that "{\em algorithm $\mbox{REV}^{*}$ examines all those nodes in OPEN
for which $opt(n) < opt(s)$.}" This can be readily verified from the graph of Figure \ref{fig130}.
On this graph, $\mbox{REV}^{*}$ starts by putting nodes $t_1$, $x$, $t_2$ and
$t_3$, with $UB$-values 0, $\infty$, 0 and 0 respectively, into OPEN. Then after
removing $t_1$ and inserting $n$ with $UB(n) = 10$ into OPEN, $\mbox{REV}^{*}$
removes $t_2$ and inserts $p$ with $UB(p) = 1$ in OPEN. Next, it selects $t_3$ from
OPEN, but cannot declare $s$ "found" as $p$ is not yet "found". Finally, it selects
$p$ from OPEN, declares $p$ as "found" and then, declares $s$ as "found", too, with
$UB(s) =  102$. Then $\mbox{REV}^{*}$ terminates, {\em without selecting $n$ from OPEN
which has $opt(n) = 10 < opt(s) = 102.$} This is a clear contradiction to the
Theorem 5.3(iii) of [Chakrabarti 1994], according to which $\mbox{REV}^{*}$ should
also have selected $n$ from OPEN, before termination.

\unitlength=0.45mm
\linethickness{0.4pt}
\begin{figure}
\centering
\caption{$\mbox{REV}^{*}$ theorem fails}
\label{fig130}
\begin{picture}(105.00,150.00)
\put(6.00,25.00){\framebox(8.00,8.00)[cc]{$t_1$}}
\put(39.00,25.00){\framebox(8.00,8.00)[cc]{$x$}}
\put(50.00,29.00){\makebox(0,0)[lc]{($\infty$)}}
\put(36.00,52.00){\circle{10.00}}
\put(20.00,82.00){\circle{10.00}}
\put(43.00,107.00){\circle{10.00}}
\put(20.00,77.00){\vector(-1,-4){11.00}}
\put(20.00,77.00){\vector(2,-3){14.00}}
\put(38.00,47.00){\vector(1,-2){7.00}}
\put(64.00,78.00){\framebox(8.00,8.00)[cc]{$t_2$}}
\put(66.00,132.00){\circle{10.00}}
\put(82.00,103.00){\framebox(8.00,8.00)[cc]{$t_3$}}
\put(66.00,127.00){\vector(-4,-3){20.00}}
\put(66.00,127.00){\vector(4,-3){19.00}}
\put(43.00,102.00){\vector(-4,-3){20.00}}
\put(43.00,102.00){\vector(3,-2){23.00}}
\put(60.00,122.00){\line(1,0){12.00}}
\put(66.00,132.00){\makebox(0,0)[cc]{$s$}}
\put(77.00,121.00){\makebox(0,0)[lc]{100}}
\put(55.00,122.00){\makebox(0,0)[rc]{1}}
\put(43.00,107.00){\makebox(0,0)[cc]{$p$}}
\put(55.00,97.00){\makebox(0,0)[lc]{1}}
\put(33.00,97.00){\makebox(0,0)[rc]{1}}
\put(20.00,82.00){\makebox(0,0)[cc]{$n$}}
\put(36.00,52.00){\makebox(0,0)[cc]{$r$}}
\put(30.00,66.00){\makebox(0,0)[lc]{1}}
\put(12.00,54.00){\makebox(0,0)[rc]{10}}
\put(44.00,41.00){\makebox(0,0)[lc]{1}}
\put(42.00,10.00){\makebox(0,0)[cc]{Implicit Graph $A$}}
\put(0.00,0.00){\framebox(105.00,150.00)[cc]{}}
\end{picture}
\end{figure}
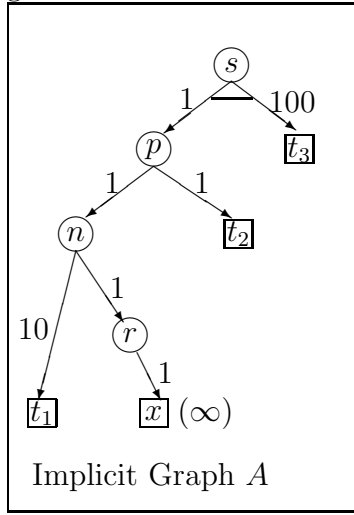

\subsection{Hvalica's Method}

In a recent paper [Hvalica 1996] has attempted to solve cyclic AND/OR graphs by attaching an arc to a new dummy node $x_{f}$, with a high heuristic value, from the node currently being expanded. This method is based on the
premise that, even if expansion of the current node creates a cycle, the algorithm can come out of the cycle by looping through it a sufficient number of times (when the cost of the expanded node, computed through its children, exceeds the high cost attached to the dummy child $x_{f}$.) [pp 108, Hvalica 1996]. In the example of Figure \ref{fig140}, this method will make a large number (H) of unnecessary evaluations of node $p$, where H represents a very high value. This method clearly violates the best-first search principle. Even if a node $n$ has a solution graph below it (and a self-loop of unit cost), and a choice of $H = h^{*}(n)$ is used when expanding $n$, there will exist cases where the looping at $n$ will violate the best-first principle globally, with respect to some other less-costly node of the graph. Similarly we can construct cases where the looping, even though not violating the best-first principle, results in unnecessary computations.

\unitlength=0.45mm
\linethickness{0.4pt}
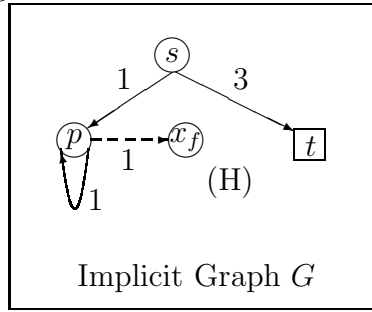
\begin{figure}
\centering
\caption{Hvalica's method loops}
\label{fig140}
\begin{picture}(110.00,90.00)
\put(48.00,75.00){\circle{10.00}}
\put(19.00,50.00){\circle{10.00}}
\put(52.00,50.00){\circle{10.00}}
\put(48.00,70.00){\vector(-3,-2){25.00}}
\put(49.00,70.00){\vector(2,-1){35.00}}
\bezier{280}(23.00,47.00)(19.00,12.00)(15.00,47.00)
\put(24.00,50.00){\line(1,0){3.00}}
\put(29.00,50.00){\line(1,0){3.00}}
\put(34.00,50.00){\line(1,0){3.00}}
\put(39.00,50.00){\line(1,0){3.00}}
\put(44.00,50.00){\vector(1,0){3.00}}
\put(48.00,75.00){\makebox(0,0)[cc]{$s$}}
\put(19.00,50.00){\makebox(0,0)[cc]{$p$}}
\put(52.00,50.00){\makebox(0,0)[cc]{$x_f$}}
\put(89.00,48.00){\makebox(0,0)[cc]{$t$}}
\put(66.00,65.00){\makebox(0,0)[lb]{3}}
\put(36.00,65.00){\makebox(0,0)[rb]{1}}
\put(58.00,42.00){\makebox(0,0)[lt]{(H)}}
\put(23.00,32.00){\makebox(0,0)[lc]{1}}
\put(20.00,9.00){\makebox(0,0)[lc]{Implicit Graph $G$}}
\put(84.00,44.00){\framebox(9.00,9.00)[cc]{}}
\put(16.00,42.00){\vector(-1,4){1.00}}
\put(35.00,47.00){\makebox(0,0)[ct]{1}}
\put(0.00,0.00){\framebox(110.00,90.00)[cc]{}}
\end{picture}
\end{figure}

\subsection{Algorithm $CFC_{REV^*}$}

As mentioned by the authors [Jim$\acute{e}$nez and Torras 2000], the algorithm $CFC_{REV^*}$ has been designed primarily keeping the efficiency in mind. While that objective seems to have been achieved (as observed in the Experimental Results), the algorithm lacks any clear theoretical framework. The paper has used the "standard notation and definitions stated in" [Mahanti and Bagchi 1985] which, however, was written for acyclic AND/OR graphs. The problems that arise in this situation (i.e. the infinite recursion in cyclical definitions) has been discussed in detail under $\mbox{REV}^{*}$, so we refrain from repeating that here. We only observe that, given the absence of a correct theoretical framework, the correctness proofs of algorithm $CFC_{REV^*}$ stand on a weak base. Again the best-first nature of the algorithm, which was shown to be violated in the case of $\mbox{REV}^{*}$, is easily violated in the case of $CFC_{REV^*}$ as well. This is precisely the reason why $CFC_{REV^*}$ evaluates many more nodes than $\mbox{AO}^{*}$ on acyclic graphs, which is based on best-first search.
 
\end{document}